\documentclass[onehalf,11pt]{beavtex}

\title{Integrated Novelty Detection Systems: From Sensor Networks to Multi-Novelty Computer Vision}
\author{Iren\'e Tematelewo}
\degree{Doctor of Philosophy}
\doctype{Dissertation}
\department{Electrical Engineering and Computer Science}
\depttype{School}
\depthead{Head}
\major{Computer Science}
\advisor{Thomas G. Dietterich}
\submitdate{September 18, 2024}
\commencementyear{2025}
\abstract{Novelty detection is crucial in various technological and scientific domains. Its importance spans from ensuring the reliability of sensor networks to enhancing the adaptability of computer vision systems. As AI systems become increasingly sophisticated and complex, effective anomaly detection is essential for maintaining reliability and accuracy. This work addresses novelty detection in weather sensor networks and computer vision. In weather sensor networks, particularly for monitoring rainfall, detecting blocked rain gauges can significantly impact data reliability, food security, and disaster management. This thesis evaluates four existing quality control methods for detecting blocked rain gauges and demonstrates their limitations, particularly in detecting clusters of faulty sensors. By comparing methods such as the Conditional Mixture Model (CMM), PSWQ (Fz-filter), Kernel Regression, and Gaussian Method, we identify their shortcomings and propose three novel quality control techniques—PCSD-SAT, PCSD-JIGM, and PCSD-JIKR—that integrate probabilistic fault detection and consistency-based diagnosis framework to improve timely detection of blocked sensors, while reducing false alarms.
	
	In the realm of computer vision, current systems often fail to handle multiple types of novelty due to their focus on single-novelty scenarios. We highlight the need for integrated systems capable of handling multiple types of novelty. To support this, we introduce the Novel Snapshot Serengeti (NOVEL-SS) dataset, derived from the Snapshot Serengeti camera trap data, which includes various forms of novelty. We also present the DCA (Detect, Characterize, and Accommodate) baseline system, designed to handle multiple novelty by detecting, characterizing (identifying the type of novelty), and accommodating them (applying the right strategy for the novelty). This thesis aims to advance the robustness and adaptability of novelty detection systems across both weather sensor networks and computer vision applications.}

\acknowledgements{
	I would like to start by thanking God for the grace, motivation, and strength that have guided me throughout this research journey. \\
	
	\noindent
	I extend my heartfelt gratitude to my advisor, Dr. Thomas Dietterich. His unwavering belief in me and mentorship has been invaluable. His ability to simplify complex problems and encourage a first-principles approach has profoundly shaped my research skills. I am also grateful for the opportunities he provided to attend conferences, which greatly enhanced my academic growth. \\
	
	\noindent
	I would like to express my appreciation to my PhD committee members --- Dr. Kiri Wagstaff, Dr. Alan Fern, Dr. Weng-Keen Wong, Dr. Ciira Maina, and my graduate council representative, Dr. Mathew Campbell --- for their time and effort in evaluating this thesis. \\
	
	\noindent
	My sincere thanks go to Dr. John Selker and the TAHMO organization for providing the dataset used in part of this thesis. I also wish to acknowledge the entire Rain Quality Control (RainQC) team at TAHMO, for their contribution to understanding rainfall data and analysis the alarms generated by the quality control system deployed in the TAHMO network. \\
	
	\noindent
	Finally, I want to express my gratitude to my parents, brother, and sisters for their unwavering support, encouragement, and prayers throughout this journey. I also extend my gratitude to my friends and colleagues --- Michael Slater, Alexander Guyer, Amin Ullah, Tadesse Zemicheal, Risheek Garrepelli, Si Liu, Femi Omilani and all those from the graduate student offices at the Kelley Engineering Center at OSU --- for their camaraderie and support along the way. \\
	
	\noindent
	This research was partly funded by ARO grant W911NF-22-2-0149. The views and conclusions contained in this document are those of the author and should not be interpreted as necessarily representing the official policies, either expressed or implied, of ARO or the United States Government.
	
	\noindent
	This material is based upon work supported by the National Science Foundation under Grant No. 1521687.}

\usepackage{algorithm}
\usepackage{algcompatible}
\usepackage{float}

\algnewcommand\algorithmicreturn{\textbf{return}}
\algnewcommand\RETURN{\State \algorithmicreturn}

\usepackage[table]{xcolor} 
\usepackage{graphicx}
\usepackage{tabularx, ragged2e}
\usepackage{caption}
\usepackage{subcaption} 
\usepackage{amsmath, bm}
\usepackage{amssymb}
\usepackage[table, dvipsnames]{xcolor}
\usepackage[numbers]{natbib}
\usepackage{url}          
\usepackage{booktabs}

\usepackage{eqparbox}
\algrenewcommand{\algorithmiccomment}[1]{\hspace{1em}$\hfill \#$ #1}

\usepackage{amsthm}
\usepackage{mdframed}
\theoremstyle{definition}

\newtheorem{definition}{Definition}[section]
\newtheorem{example}{Example}[section]

\DeclareMathOperator*{\argmax}{arg\,max}

\usepackage{amsfonts}      
\usepackage{nicefrac}      
\usepackage{microtype}

\usepackage{multirow}

\newcolumntype{C}{>{\centering\arraybackslash}X}
\usepackage{diagbox}
\usepackage{soul} 

\usepackage{multicol}

\def\etal{\emph{et al}.}

\algnewcommand\AND{\textbf{ and }}

\begin{document}
	\maketitle
	
	\mainmatter
	
	\clearpage
	\begin{center}
		\thispagestyle{empty}
		\vspace*{\fill}
		This thesis is dedicated to my parents, Telewo Jean-Paul and Nganmini P. Marceline. Your boundless love, unwavering support, and heartfelt prayers have consistently guided me, especially throughout this journey. Despite having only a primary school education, your sacrifices and encouragement have given me the courage to take bold risks, and embrace challenges without complaint. I am profoundly grateful for everything you have done to make this accomplishment possible. Thank you.
		\vspace*{\fill}
	\end{center}
	\clearpage

	\chapter{Introduction}
	
	\newpage
	\section{Introduction}
	Novelty detection—the identification of patterns or occurrences that deviate from the norm—has emerged as a critical capability in a broad spectrum of technological and scientific fields. Its significance encompasses such diverse tasks as safeguarding the reliability of sensor networks to enhancing the accuracy and adaptability of computer vision systems. As the complexity of systems increases, the ability to effectively detect and address anomalies becomes essential for maintaining operational excellence and data integrity.
	
	In sensor networks, novelty detection is instrumental in diagnosing unexpected failures or malfunctions. These networks are integral to numerous applications, including environmental monitoring, industrial automation, and infrastructure management. Reliable sensor data is crucial for informed decision-making and operational efficiency. Novelty detection helps to identify issues such as sensor drift, degradation, or interference, thereby ensuring that data remains accurate and the system operates smoothly.
	
	In the realm of computer vision, novelty detection also plays a pivotal role. Computer vision systems are increasingly employed in a variety of contexts, from autonomous vehicles to surveillance and quality control in manufacturing. The ability to detect anomalies---such as unexpected objects or changes in the scene---is vital for the effective functioning of these systems. Building computer vision systems that can handle multiple types of novelty is essential for their robustness and adaptability.
	
	Current novelty-aware computer vision systems often face limitations due to their design, which frequently focuses on handling a single type of novelty. This can constrain their effectiveness in real-world scenarios where diverse and unpredictable anomalies are encountered. For instance, a system trained exclusively to detect unusual objects may struggle to handle changes in lighting conditions, varying viewing angles, or different types of occlusions simultaneously. This limitation underscores the need for advanced systems capable of managing multiple novelties to ensure comprehensive and reliable visual analysis.

	In the specific context of environmental monitoring, quality control is of utmost importance. For instance, accurate rainfall prediction using weather sensor networks is critical for regions where rainfed agriculture is vital, such as in many countries in Africa, Latin America, and South Asia \citep{rockstrom2015agriculture, wani2009rainfed}. Reliable rainfall data directly impacts agriculture by guiding irrigation, crop management, and water resource allocation. However, rain gauges, which measure rainfall, can be obstructed by debris or ice, leading to erroneous readings. Implementing fault detection techniques to identify these blockages ensures more accurate data, supports effective agricultural decision-making, and enhances disaster prevention capabilities.
	
	This thesis addresses the challenge of multiple novelty detection in weather sensor networks and computer vision. In weather sensor networks, we specifically focus on the timely detection of blocked rain gauges when auxiliary data, such as satellite data, is unavailable. We examine blocked rain gauges under two scenarios: independent sensor faults (Spatially Random Faults, SRF) and simultaneous sensor faults (Spatially Clustered Faults, SCF). Spatially Clustered Faults are further categorized into SCF-IN, where faulty sensors are surrounded by working ones, and SCF-OUT, where the faulty sensors are at the network's periphery. These faults could result from localized phenomena like dust or sandstorms. Our aim is to develop quality control (QC) methods that detect these faults promptly while minimizing false alarms. We evaluate four methods using rainfall data from the Oklahoma Mesonet \cite{mcpherson2007statewide} and the Trans-African Hydro-Meteorological Observatory (TAHMO) \citep{selker-2014-tahmo} weather networks: (i) the Conditional Mixture Model (CMM) \citep{zemicheal2020conditional}, (ii) the PSWQ (Fz-filter) \citep{de2019quality}, (iii) a kernel regression method similar to \citep{Barnes1964}, and (iv) a joint Gaussian method (GM). Our analysis reveals that despite excelling at detecting spatially random faults, these methods struggle with detecting large clusters of faulty sensors. We propose three new QC methods—PCSD-SAT, PCSD-JIGM, and PCSD-JIKR—that improve detection by applying joint reasoning across all rainfall observations. These methods integrate probabilistic fault detection into a consistency-based diagnosis framework to ensure that sensor observations are consistent with the broader network.
	
	In computer vision (CV), we highlight the limitations of current novelty detection paradigms, which handle only a single type of novelty. We advocate for the development of integrated CV systems capable of detecting and accommodating various types of novelty, as encountered in real-world environments. To support this, we introduce Novel Snapshot Serengeti (NOVEL-SS), a benchmark dataset derived from the Snapshot Serengeti camera trap dataset \cite{swanson2015snapshot}. NOVEL-SS presents multiple forms of novelty, challenging CV systems to detect and characterize these novelties to determine the appropriate response strategy. Alongside NOVEL-SS, we propose DCA (Detect, Characterize, and Accommodate), a baseline system designed for handling multiple types of novelty.
	
	\section{Contributions}
	Our first contribution is the analysis and characterization of existing methods designed to detect blocked rain gauges. We evaluate these performance under various conditions, including independent and clustered faults, as well as sparse networks. We employ task-specific metrics, such as detection delay, to assess these methods' efficacy, revealing their limitations in handling clustered faults. Detailed findings are presented in Chapter \ref{ch1:comparison-of-methods}.
	
	Our second contribution is the design of three algorithms---PCSD-SAT, PCSD-JIGM and PCSD-JIKR---to address the limitations of existing QC methods, particularly for detecting large clusters of faulty sensors. PCSD-SAT utilizes Boolean satisfiability for joint reasoning, while PCSD-JIKR and PCSD-JIGM employ kernel regression and Gaussian models, respectively, to enhance fault detection. Empirical results show that PCSD-JIGM and PCSD-JIKR outperform existing methods in detecting clustered faults, as discussed in Chapter \ref{ch2:pcsd}.
	
	Our third contribution is the introduction NOVEL-SS, a new benchmark dataset for multiple novelty aware computer vision, and DCA, a baseline system for NOVEL-SS. We propose a task structure, an evaluation protocol, and evaluation metrics, to foster the development of more integrated solutions to novelty detection and accommodation.

	\phantom{}\newpage
	\phantom{}\vfill
	\begin{center}
		\heading
		A comparison of methods for detecting blocked rain
		gauges in automated weather network operations
	\end{center}
	\vfill
	Iren\'{e} Tematelewo, Austin Kaburia, Michael Slater, Ciira wa Maina, and Thomas G. Dietterich
	\vfill\noindent
	\\
	
	\vfill
	\chapter{A comparison of methods for detecting blocked rain
		gauges in automated weather network operations} \label{ch1:comparison-of-methods}
	
	\section{Abstract}
	Precipitation is one of the most important and most difficult quantities for automated weather stations to measure. A frequent failure mode in many rain gauges is blockage, where dust and debris prevent water from entering the instrument, which causes it to falsely report zero values. Standard statistical methods fail in this case, because zero is also the most common correct value for precipitation in most locations on Earth. For network operations, it is important to detect a blocked rain gauge as soon as possible, so that a technician can visit and clean the rain gauge and prevent long periods of data loss. This paper compares four algorithms for detecting blocked rain gauges using data from the Oklahoma Mesonet and the TAHMO weather network. We measure missed detections, false detections, and detection delay on simulated clogged rain gauges.
	
	\section{Introduction} \label{sec:ch1_intro}
	Surface measurement of weather is important for a variety of reasons including climate mapping \citep{Daly2000}, climate change analysis \citep{Daly2021}, weather forecasting \citep{Szturc2022}, ecological science \citep{daly2019hja}, and severe weather warning \citep{yeung2014development}. Consequently, quality control methods for assessing the accuracy of weather station measurements have been developed to support each of these goals. Methods aimed at ecological science, climate mapping, and climate change can be applied retrospectively, weeks, months, or years after the data were collected. They can assume the availability of baseline climatology. In these applications, the focus of quality control is usually to determine which observations are trustworthy and to correct for biases resulting from variation in measurement instruments and the time that daily observations were taken \citep{Daly2021}. Often, it is also desirable to impute missing values in order to obtain a clean long-term record \citep{Bardossy2014,Teegavarapu2008}.
	
	Weather forecasting and severe weather warning operate on a much shorter time scale. Ground data for weather forecasts is typically assimilated at 6-hour intervals \citep{ecmwf2024}; readings for flash floods and severe storms need to be measured at 5-10 minute intervals. The quality control process focuses on determining which weather stations to trust. Sub-daily quality control is particularly important for measuring peak rainfall rates and for detecting over-reporting errors \citep{Villalobos-Herrera2022}. 
	
	Weather network operations fall in between these extremes. It is important to detect failed (or out-of-calibration) sensors, but data can be aggregated across multiple days to reduce the false alarm rate. Fundamentally, there is a tradeoff between detection delay---the time between sensor failure and the time it is detected---and the false alarm rate---the fraction of failure alarms that turn out to be false. For network operations, we want to detect the failure promptly, but we also want to avoid sending a technician to visit the station unnecessarily. 
	
	This paper focuses on the problem of detecting blocked rain gauges---that is, gauges where the collecting cone or input orifice has become blocked by dust or debris. The challenge with rainfall is that it exhibits a highly skewed non-Gaussian distribution \citep{Szturc2022} in which the most commonly-observed value is zero, which is also the value reported by a blocked rain gauge, so simple ``range checks'' cannot detect blockage. Precipitation also exhibits low spatial auto-correlation---it may rain at one station but be dry at a station only a few kilometers away. This means simple QC methods comparing nearby stations may not be statistically reliable.
	
	To overcome these problems, many quality control methods rely on additional information sources such as rainfall reports from high-quality well-maintained reference stations or estimates of rainfall fields from weather radars. In this paper, we are interested in methods that do not require such auxiliary data (Requirement 1). We are part of the team that is deploying and managing the Trans-African Hydro-Meteorological Observatory (TAHMO) automated weather station network \citep{selker-2014-tahmo}. This network seeks to cover all of sub-Saharan Africa. This is a region that lacks weather radars. It also lacks long-term climatological records. Furthermore, to achieve its coverage goal, the TAHMO network is continually installing new stations, so we need methods that can work on stations as soon as they are installed (Requirement 2). Finally, the TAHMO stations rely on battery-powered cellular data networks to report their observations. Data transmissions can be intermittent. Hence, we seek quality control methods that can tolerate missing readings throughout the network (Requirement 3).
	
	This paper describes and compares three QC methods that meet these requirements (abbreviated R1, R2 and R3). All of them are spatial consistency checks that evaluate the reported rainfall at a target station based on the reports from nearby stations. The simplest method is FZ-filter \citep{de2019quality}, which has apparently been in use by the US National Weather Service since at least 2006 \citep{Kondragunta2006}.\footnote{\cite{Kondragunta2006} cites \citep{Madsen1993} as the first publication of the method, but we have not been able to locate a copy of that paper.} The second method, kernel regression (KR) is a regression approach that predicts the rainfall at each target station based on a kernel regression computed from the rainfall reported by neighboring stations. The final method, GM, is based on a joint Gaussian model. A multivariate Gaussian is fit to the precipitation data from the entire network. Then, for each target station, it is easy to compute the conditional probability distribution for the amount of rain reported at that station conditioned on the values reported at all of the other stations. 
	
	These three methods are compared to a baseline automated QC method, the conditional mixture model \citep[CMM;][]{zemicheal2020conditional}, which is currently deployed on the TAHMO network. A smaller comparison---relevant to the TAHMO setting---with a manual satellite-based QC process is also provided for reference. The motivation for this study is that the CMM does not meet requirements R1, R2 and R3 -- it cannot handle missing measurement values, tends to produce too many false positives, and requires too much historical data. Training a CMM model for a target station requires at least one year of training data for the target station and for each of its neighbors. 
	
	This paper describes all four of these methods and then compares them on simulated blocked rain gauge data from Oklahoma Mesonet \cite{mcpherson2007statewide} and simulated blocked rain gauge data from TAHMO. For each method, we measure the trade-off between the false alarm rate (FAR) and the delay between the onset of the fault and the first alarm raised, for a given missed alarm rate (MAR). We visualize this trade-off as an Activity Monitoring Operating Characteristic (AMOC) curve \citep{fawcett1999activity}. We also evaluate each method's ability to detect faults in an increasingly sparse network and when faced with the failure of neighboring stations.

	The remainder of this paper is organized as follows: Section \ref{sec:ch1_methods} describes the four QC methods, while section \ref{sec:ch1_datasets} details the weather datasets used in this study. Section \ref{sec:ch1_experiments} describes the experimental setup and the evaluation metrics employed to compare these methods. The results are presented in Section \ref{sec:ch1_results}, and the paper concludes with Section \ref{sec:ch1_conclusion}.

	\section{Methods} \label{sec:ch1_methods}
\begin{figure}
		\centering
		\includegraphics[width=0.6\linewidth]{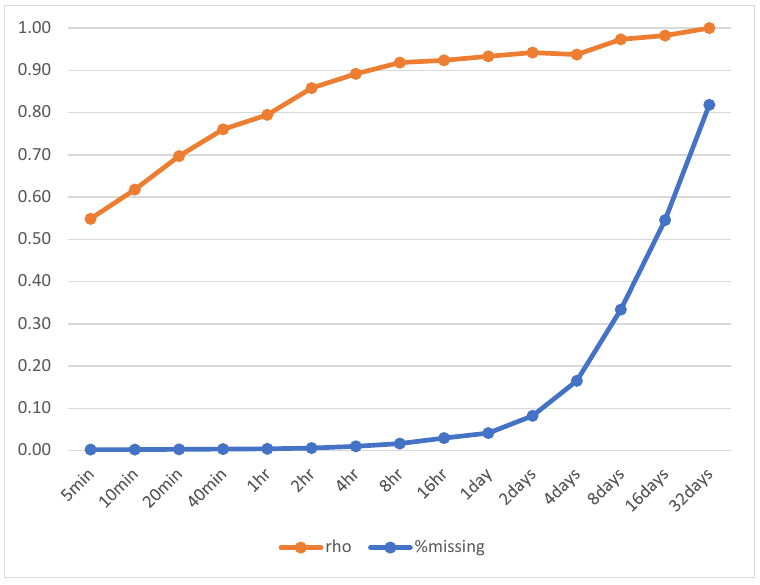}
		\caption{Correlation coefficient (rho, orange) between precipitation readings at the CHIC and NINN stations (the closest pair of stations) in the Oklahoma Mesonet as a function of the length of time over which precipitation is aggregated. The blue curve shows the fraction of missing values.}\label{fig:tradeoff}
	\end{figure}

	Let $i=1,\ldots,n$ index the $n$ rain gauges in the weather network. For TAHMO, each weather station has only one rain gauge, so $n$ is also the number of weather stations. Let $loc(i)$ be the location (i.e., longitude, latitude, and elevation) of station $i$. Let $d(i,j)$ denote the distance between stations $i$ and $j$. We are currently only measuring the displacement in longitude and latitude, but other methods employ more sophisticated measures \citep{Daly2000}. We will often focus on a fixed ``target'' station $i$ and compare it to a set of neighboring stations $\eta(i)$. For example, $\eta(i)$ could be the set of all stations within 100 km of station $i$. Let $r(i,t)$ denote the total precipitation reported by station $i$ on day $t$. If there is no report from station $i$ on day $t$, we set $r(i,t) = \bot$ to indicate that the value is missing. We aggregate precipitation over 24-hour periods. This period was chosen based on a spatial correlation analysis (see Figure ~\ref{fig:tradeoff}), which showed that this gives a good tradeoff between maximizing inter-station correlations while not losing too many days due to missing values. Let $r_{min}$ denote the amount of rain that is ``effectively zero''. If $r(i,t)\leq r_{min}$, station $i$ is considered to have measured no precipitation on day $t$. In our experiments, we set $r_{min} = 1\mathrm{mm}$.
	
	\subsection{PWSQC (FZ-filter)}
	FZ-filter \citep{Kondragunta2006,Madsen1993} is a simple yet robust QC algorithm designed specifically for the detection of blocked rain gauges. We employ the version implemented in PWSQC \citep{de2019quality}. Pseudo-code is shown in Algorithm \ref{alg:fz-filter}. FZ-filter computes the median $r_{med}$ of the rainfall reported by neighboring stations after removing missing values ($r(j,t) = \bot$). If this is non-zero ($r_{med}>r_{min})$, the day is declared to be a ``wet day''. If the target station $i$ reported zero on a wet day, this is a ``wet zero'' day for station $i$. If $i$ exhibits $z_{max}$ consecutive wet zero days (ignoring dry days), it is declared to be blocked. PWSQC (FZ-filter) doesn't require a training dataset or auxiliary dataset, and it handles missing values among the neighbors. It therefore satisfies requirements R1, R2 and R3. However, PWSQC (FZ-filter) can fail if a spatially-clustered set of rain gauges are all blocked simultaneously. Whenever more than half of the neighbors of $i$ are clogged, the median rainfall of the neighbors will be zero, the day will be declared to be a ``dry day'', and the blockage at station $i$ will go undetected.
	
	\begin{algorithm}
		\caption{FZ-filter}
		\label{alg:fz-filter}
		\begin{algorithmic}
			\STATE \textbf{Data Structures and Parameters}
			\STATE \quad $z$: $z[i]$ is number of ``wet zero'' days for station $i$
			\STATE \quad $r_{min}$: rain $\leq r_{min}$ is treated as zero
			\STATE \quad $z_{max}$: number of wet zero days before a station is
			\STATE \quad \quad declared ``blocked''
			\STATE \textbf{Procedure}
			\STATE $z[i] \leftarrow 0$ for all stations $i = 1,\ldots,n$
			\LOOP
			\STATE \textbf{read} day $t$ measurements $r(i,t)\;\; \forall{i}$ 
			\FOR{$i = 1,\ldots,n$}
			\IF{$r(i,t) > r_{min}$} \STATE $z[i] \leftarrow 0$ 
			\COMMENT{If station $i$ reports nonzero rain, reset the wet zero counter}
			\ENDIF
			\STATE $\eta_t(i) \leftarrow \{j \in \eta(i) : rain(j,t) \neq \bot\}$
			\STATE $r_{med} \leftarrow \mathbf{median}(\{r(j,t) : j \in \eta_t(i)\})$
			\IF{$r_{med} > r_{min}$ \AND $r(i,t) \leq r_{min}$} 
			\COMMENT{Wet zero day}
			\STATE $z[i] \leftarrow z[i]+1$
			\ENDIF
			\IF{$z[i] \geq z_{max}$} 
			\STATE \textbf{report gauge }$i$\textbf{ blocked}
			\ENDIF
			\ENDFOR
			\ENDLOOP
		\end{algorithmic}
	\end{algorithm}
	\vspace*{0.1in}

	\begin{algorithm}
		\caption{KR}
		\label{alg:kr}
		\begin{algorithmic}
			\STATE \textbf{Data Structures and Parameters}
			\STATE \quad $z$: $z[i]$ is number of days ``prediction error'' for station $i$
			\STATE \quad $r_{min}$: rain $\leq r_{min}$ is treated as zero
			\STATE \quad $z_{max}$: number of prediction error days before a station is declared ``blocked''
			\STATE \quad $\tau$: anomaly detection threshold
			\STATE \textbf{Procedure}
			\STATE $z[i] \leftarrow 0$ for all stations $i = 1,\ldots,n$
			\LOOP
			\STATE \textbf{read} day $t$ measurements $r(i,t)\;\; \forall{i}$ 
			\FOR{$i = 1,\ldots,n$}
			\IF{$r(i,t) > r_{min}$} \STATE $z[i] \leftarrow 0$ 
			\ENDIF
			
			\STATE $\hat{r} \leftarrow \Big( \sum_{j \neq i} \kappa(i,j) \times rain(j,t) \Big) / \sum_{j \neq i} \kappa(i,j)$
			\IF{$|\hat{r} - r(i,t)|^2 > \tau$}  \COMMENT{Prediction error day}
			\STATE $z[i] \leftarrow z[i]+1$
			\ENDIF
			\IF{$z[i] \geq z_{max}$} 
			\STATE \textbf{report gauge }$i$\textbf{ blocked}
			\ENDIF
			\ENDFOR
			\ENDLOOP
		\end{algorithmic}
	\end{algorithm}
	\vspace*{0.1in}

	\subsection{Kernel Regression}
	The Kernel Regression (KR; Algorithm~\ref{alg:kr}) method is nearly identical to FZ-filter, except that instead of computing the median rainfall of the neighbors, it applies kernel regression. We employ the squared exponential kernel:
	\begin{equation}\label{eq:rbf}
		\kappa(i,j) = \sigma_f^2\exp\left(-\frac{d(i,j)^2}{2\ell^2}\right).
	\end{equation}
	The value of $\ell$ gives the length scale of the kernel. It must be tuned to give good performance. For KR, we set $\sigma_f = 1$. Kernel regression is identical to the first pass of Barnes interpolation \citep{Barnes1964}.
	
	One difference from FZ-filter is that KR does not have a definition of dry days, so it raises an alarm if $z_{max}$ consecutive days show a squared prediction error greater than $\tau$. 
	
	Kernel regression does not depend on auxiliary data and can handle missing data from neighboring stations by excluding any station with missing values from the computation of $\hat{r}$ (see algorithm \ref{alg:kr}). Thus, kernel regression satisfies requirements R1 and R3. However, the kernel’s length-scale parameter, $\ell$, must be estimated using historical data. This challenge can be addressed by setting $\ell$ for newly installed stations to the same value as the nearest stations. In our experiment, we use the same value of $\ell$ for all stations.

	\begin{algorithm}
		\caption{GM}
		\label{alg:gm}
		\begin{algorithmic}
			\STATE \textbf{Data Structures and Parameters}
			\STATE \quad $z$: $z[i]$ is number of ``prediction error'' days for station $i$
			\STATE \quad $r_{min}$: rain $\leq r_{min}$ is treated as zero
			\STATE \quad $z_{max}$: number of prediction error days before a station is declared ``blocked''
			\STATE \quad $\tau$: anomaly detection threshold
			\STATE \quad $\boldsymbol{\mu}$, $\boldsymbol{\Sigma}$: multivariate Gaussian model fit to historical data
			\STATE \textbf{Procedure}
			\STATE $z[i] \leftarrow 0$ for all stations $i = 1,\ldots,n$
			\LOOP
			\STATE \textbf{read} day $t$ measurements $r(i,t)\;\; \forall{i}$ 
			\FOR{$i = 1,\ldots,n$}
			\IF{$r(i,t) > r_{min}$} \STATE $z[i] \leftarrow 0$ 
			\ENDIF
			
			\STATE $\hat{r} \leftarrow \mu_i + \boldsymbol{\Sigma}_{i,-i}\boldsymbol{\Sigma}^{-1}_{-i,-i}(\boldsymbol{rain}_{-i}(t-\boldsymbol{\mu}_{-i})$
			\STATE $\sigma^2 \leftarrow \boldsymbol{\Sigma}_{i,i} - \boldsymbol{\Sigma}_{i,-i}\boldsymbol{\Sigma}_{-i,-i}^{-1}\boldsymbol{\Sigma}_{-i,i} $
			\IF{$(\hat{r} - r(i,t))^2/\sigma^2 > \tau$} 
			\STATE $z[i] \leftarrow z[i]+1$
			\COMMENT{Prediction error day}
			\ENDIF
			\IF{$z[i] \geq z_{max}$} 
			\STATE \textbf{report gauge }$i$\textbf{ blocked}
			\ENDIF
			\ENDFOR
			\ENDLOOP
		\end{algorithmic}
	\end{algorithm}
	\vspace*{0.1in}
	
	\subsection{Joint Gaussian Model} \label{subsec:ch1_gm}
	The joint Gaussian method (GM), fits a multivariate Gaussian distribution to the daily historical rainfall observed over the entire network. Let $\boldsymbol{\mu}$ and $\boldsymbol{\Sigma}$ be the parameters of the joint Gaussian distribution over all the stations in the network. The conditional distribution of station $i$ given the rest of the stations is a Gaussian with mean $\mu_{it}$ and variance $\sigma_{it}^2$, computed as follows \citep{bishop2006pattern, murphy2012machine}:
	\begin{align*}
		\mu_{it} &= \mu_i + \boldsymbol{\Sigma}_{i,-i}\boldsymbol{\Sigma}_{-i,-i}^{-1}(\textbf{\textit{rain}}_{-i}(t- \boldsymbol{\mu}_{-i}) \\
		\sigma^2_{it} &= \boldsymbol{\Sigma}_{i,i} - \boldsymbol{\Sigma}_{i,-i}\boldsymbol{\Sigma}_{-i,-i}^{-1}\boldsymbol{\Sigma}_{-i,i},
	\end{align*}
	where $\boldsymbol{\Sigma}_{-i,-i}$ is the covariance matrix $\boldsymbol{\Sigma}$ without row $i$ and column $i$, $\boldsymbol{\Sigma}_{i,-i}$ is the $i$th row of the covariance matrix with the $i$th column deleted, 
	and $\textbf{\textit{rain}}_{-i}(t)$ is the vector of rainfall observed at all weather stations on day $t$, excluding station $i$. For any other station $j$ that did not report rain on day $t$, its entries are also deleted from $\boldsymbol{\mu_{-i}}$ and $\boldsymbol{\Sigma}_{i,-i}$, and the $j$th row and $j$ column are deleted from $\boldsymbol{\Sigma}_{-i,-i}$.

	Like PWSQC (FZ-filter) and KR, GM does not rely on auxiliary data and can manage missing neighbors. Thus, GM meets requirements R1 and R3. However, before fault detection can be performed on a new station $j$, GM requires an estimate of the mean rainfall $\mu_j$ and the covariance of that station with the rest of the network (inserting row $j$ and column $j$ into $\boldsymbol{\Sigma}$). This necessitates some historical data, which means that GM does not meet Requirement 2.

	\begin{algorithm}
		\caption{CMM}
		\label{alg:cmm}
		\begin{algorithmic}
			\STATE \textbf{Data Structures and Parameters}
			\STATE \quad $z$: $z[i]$ is number of ``prediction error'' days for station $i$
			\STATE \quad $r_{min}$: rain $\leq r_{min}$ is treated as zero
			\STATE \quad $z_{max}$: number of prediction error days before a station is declared ``blocked''
			\STATE \quad $\tau_i$: anomaly detection threshold for station $i$
			\STATE \quad $\forall i: \phi_i, \omega_i$ model parameters for each station
			\STATE \quad $\sigma_i^2$: fitted residual variance for station $i$
			\STATE \quad $\epsilon$: parameter to prevent division by zero
			\STATE \textbf{Procedure}
			\STATE $z[i] \leftarrow 0$ for all stations $i = 1,\ldots,n$
			\LOOP
			\STATE \textbf{read} day $t$ measurements $r(i,t)\;\; \forall{i}$ 
			\FOR{$i = 1,\ldots,n$}
			\IF {$r(i,t) \neq \bot$\AND $\forall j\in \eta(i) \; r(j,t) \neq \bot$}
			\STATE $\hat{p}_1 \leftarrow \frac{1}{1 + \exp{-(\phi^i_0 + \phi_1^{i\top} \boldsymbol{rain}_{\eta(i)}(t)})}$
			\STATE $\mu_i \leftarrow \omega_0^i + \sum_{j\in \eta(i)} \omega_j^i \log(r(j,t)+\epsilon)$
			\STATE $d_1 \leftarrow \hat{p}_1 \text{Normal}(\log(r(i,t) + \epsilon); \mu_i, \sigma^2_i)$
			\COMMENT{$\text{Normal}$ is the Gaussian density}
			\IF {$r(i,t) \leq r_{min}$} 
			\STATE $s \leftarrow -\log \min\{ 1 - \hat{p}_1, d_1\}$
			\ELSE \STATE $s \leftarrow -\log d_1$
			\ENDIF
			\IF{$s > \tau_i$} 
			\STATE $z[i] \leftarrow z[i]+1$
			\COMMENT{Prediction error day}
			\ENDIF
			\IF{$z[i] \geq z_{max}$} 
			\STATE \textbf{report gauge }$i$\textbf{ blocked}
			\ENDIF
			\ENDIF
			\ENDFOR
			\ENDLOOP
		\end{algorithmic}
	\end{algorithm}
	\vspace*{0.1in}
	
	\subsection{Conditional Mixture Model} \label{subsec:ch1_cmm}
	The conditional mixture model (CMM) QC method, introduced by \citet{zemicheal2020conditional}, combines two sub-models: A classifier for each target station $i$ that predicts the probability of rain (versus no rain) at the target station and a log-linear regression model that predicts the amount of rain at the target station. Both models base their predictions on the rain reported at a fixed set of neighboring stations $\eta(i)$. 
	
	To fit these models for station $i$, we construct a training set $D_i$ consisting of pairs $r(i,t)$ and $\boldsymbol{rain}_{\eta(i)}(t)$, where, as before, $\boldsymbol{rain}_{\eta(i)}(t)$ is a vector of length $|\eta(i)|$ of the precipitation values observed at the neighboring stations $\eta(i)$ on day $t$. The data set for station $i$ consists of approximately 365 days over the history of the weather network where $r(i,t)$ and $\boldsymbol{rain}_{\eta(i)}(t)$ were all observed. 
	
	Let $y(i,t)=1$ if $r(i,t)>r_{min}$ (i.e., a non-trivial amount of rain was observed at station $i$ and time $t$). Let $y(i,t)=0$ otherwise. Define the logistic regression model as
	\begin{equation} \label{eq:cmm-logistic}
		p_\phi^i(y(i,t) | \boldsymbol{rain}_{\eta(i)}(t)) = \frac{1}{1 + \exp{-(\phi^i_0 + \phi_1^{i\top} \boldsymbol{rain}_{\eta(i)}(t)})},
	\end{equation}
	where $\phi^i_0$ is the intercept term and $\phi^i_1$ is a vector of length $|\eta(i)|$ corresponding to the number of neighboring stations. Let $\phi^i = (\phi^i_0, \phi^i_1)$ be the parameter vector for the logistic regression model $p^i_\phi$ for station $i$. Let $\hat{p}_1(i,t) = p^i_\phi(y(i,t)=1|\boldsymbol{rain}_{\eta(i)}(t))$ be a shorthand for the predicted probability of rain at station $i$ and day $t$. 
	
	After fitting this logistic regression model to $D_i$, ZeMicheal and Dietterich then fit a log linear model to minimize the following likelihood function
	\begin{equation}
		\mathcal{L}(\omega^i) = \sum_i \hat{p}_1^i(s,t)\left[\log(r(i,t)+\epsilon) - (\omega_0^i + \sum_{j\in \eta(i)} \omega_j^i \log(r(j,t)+\epsilon))\right]^2,
	\end{equation}
	where $\omega^i$ gathers together $\omega_0^i$ and $\omega_j^i$ parameters into a vector. This is ordinary least-squares regression with three changes. First, we take the logarithm of the observed precipitation to transform the distribution to be more Gaussian. Second, we add a small positive amount $\epsilon$ to the precipitation to avoid taking the logarithm of zero. Third, the squared difference between the log of observed rain $r(i,t)$ and the linear regression model is weighted by the predicted probability of rain computed by the logistic regression model. Finally, we define $\sigma^2_i$ as the mean of the (unweighted) squared residuals on a log scale. 
	
	We can view the linear regression model as providing the mixing coefficients for a model with two components. The first component is an impulse at zero that integrates to $1-\hat{p}_1(i,t)$, and the second component is a Gaussian distribution with mean $\mu_i = \omega_0^i + \sum_{j\in \eta(i)} \omega_j^i \log(r(j,t)+\epsilon)$ and variance $\sigma^2_i$. We can write the combined model as
	\begin{multline}
		\label{eq:CMM}
		\qquad P(R(i,t)=r \mid \boldsymbol{rain}_{\eta(i)}(t))  = 
		\begin{cases}
			1 - \hat{p}_1(i,t) & r = 0 \\
			\hat{p}_1(i,t)  \mathcal{N}(\log(r + \epsilon); \mu_i, \sigma^2_i) & r > 0.
		\end{cases}
	\end{multline}
	
	ZeMicheal and Dietterich found experimentally that extending the $r=0$ case to include the log-linear density (evaluated at zero) improved the results. Hence, the full conditional mixture model is
	\begin{multline}
		\label{eq:CMM_zemicheal}
		P(R(i,t)=r \mid \boldsymbol{rain}_{\eta(i)}(t))  = 
		\begin{cases}
			\min\{1 - \hat{p}_1(i,t),\; \hat{p}_1(i,t) \mathcal{N}(\log(\epsilon); \mu_i, \sigma^2_i)\} & r = 0 \\
			\hat{p}_1(i,t)  \mathcal{N}(\log(r + \epsilon); \mu_i, \sigma^2_i) & r > 0.
		\end{cases}\qquad
	\end{multline}
	
	A new observation $(r(i,t), \boldsymbol{rain}_{\eta(i)}(t))$ is scored by computing the negative log likelihood 
	$- \log P(R(i,t) = r(i,t) \mid \boldsymbol{rain}_{\eta(i)}(t))$ and signalling an alarm if it exceeds a station-specific threshold $\tau_i$. These thresholds are tuned based on the training data. Algorithm~\ref{alg:cmm} shows the pseudo-code for CMM.

	CMM has been deployed as ``RainQC'' on the TAHMO weather network since 2019 and produces a report of potential precipitation sensor anomalies each day. Unfortunately, it suffers from the weaknesses presented in section \ref{sec:ch1_intro}. First, it violates R2 because it requires historical data for each station and its neighbors to fit its regression model. Roughly half of the TAHMO stations currently lack sufficient historical data and, hence, cannot be checked using this model. Second, for a given target station $i$, if any of its neighbors $\eta(i)$ fail to report precipitation on a given day, the regression model cannot be evaluated. In such cases, we say that the model is not ``data complete''. On a typical day, more than half of the CMM models cannot be evaluated, as they are not data complete. Third, CMM computes a QC score on each individual day separately, rather than combining them a longer time window, which generally leads to higher false alarm and missed alarm rates.

	\section{Datasets} \label{sec:ch1_datasets}
\begin{figure}[H]
		\includegraphics[width=1.0\linewidth]{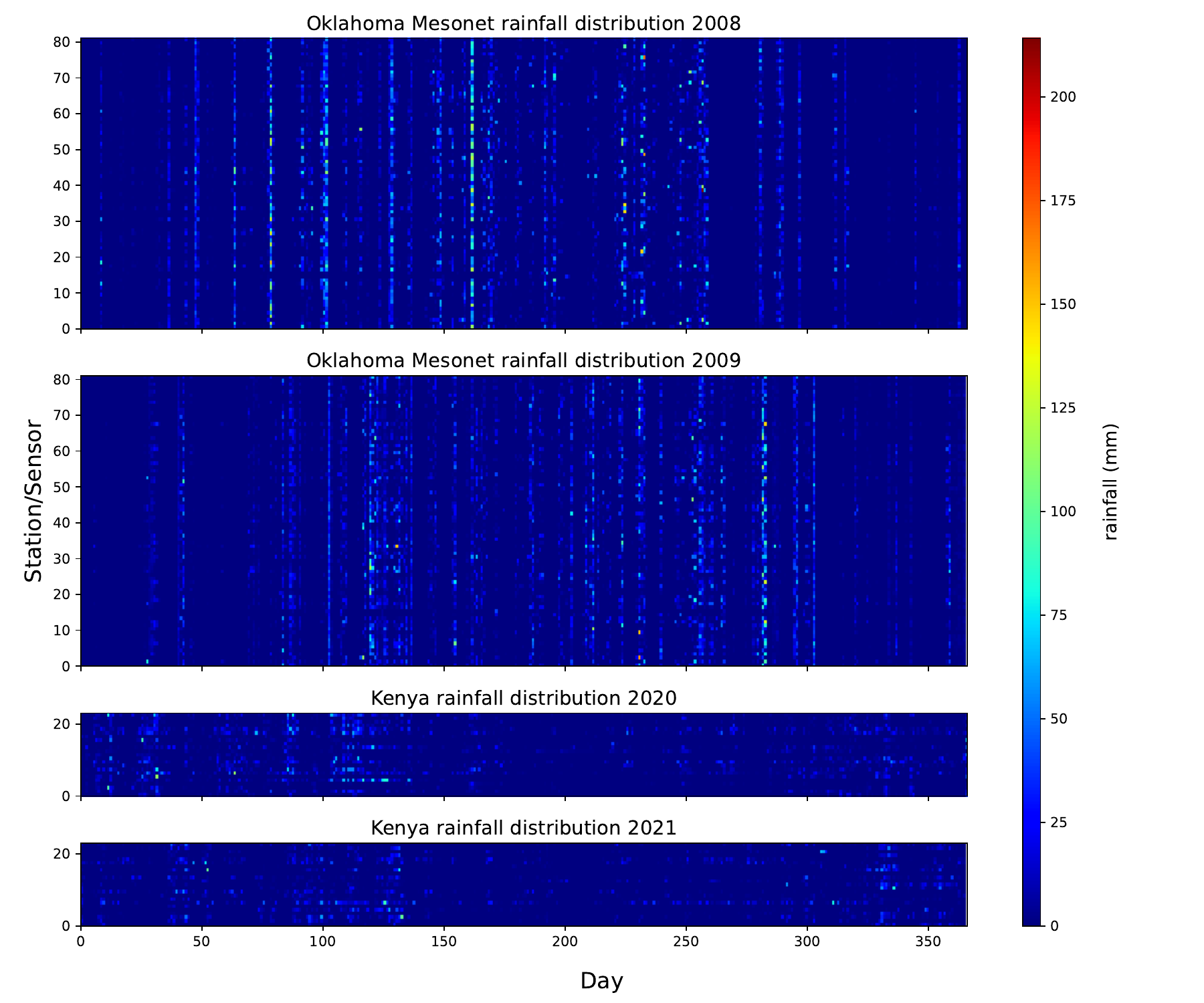}
		\caption{Historical training and testing rainfall distribution in the Oklahoma Mesonet and the Kenya dataset.}
		\label{fig:ch1_rainfall_distribution}
	\end{figure}

	This study utilizes rainfall datasets from the Oklahoma Mesonet and the TAHMO weather networks to evaluate the quality control (QC) methods. The Oklahoma Mesonet dataset is highly reliable, benefiting from a rigorous QC process and prompt repair or replacement of faulty gauges. Consequently, the measurements from this network are considered accurate, allowing for robust assessment of the QC algorithms. The Oklahoma network features a relatively uniform distribution of stations across the state, which experiences numerous large-scale rain events throughout the year (see Figure \ref{fig:ch1_rainfall_distribution}, upper two panels).
	
	In contrast, the TAHMO dataset consists of more densely clustered stations across a diverse landscape, with less frequent and more localized rain events (see Figure \ref{fig:ch1_rainfall_distribution}, lower two panels). Due to the less stringent QC processes and slower repair or replacement times, the reliability of the TAHMO measurements is comparatively lower.

	\subsection{TAHMO Dataset} \label{subsec:ch1_ke_datase}
	The weather stations in TAHMO are equipped with ATMOS 41 sensors manufactured by METER. These sensors monitor a range of weather variables, including precipitation, air temperature, atmospheric pressure, relative humidity, solar radiation, and wind speed and direction. Most ATMOS 41 sensors in TAHMO provide measurements at a 5-minute resolution. These measurements pass through three stages of quality control: (a) rule-based checks based on \cite{Fiebrich2020}, (b) CMM, and (c) comparison of monthly total precipitation against satellite data products. Faulty observations are assigned quality flags based on these checks and reports created by field technicians when they visit the stations to make repairs.

	To obtain data for our experiments, we retrieved  the raw rainfall data and corresponding quality flags for all stations in Kenya. We chose this subset because of Kenya's relatively high station density and its more robust quality control process. We removed all flagged observations confirmed by network managers after inspection.  We then selected two years (2020 and 2021) with the most complete data. Due to a high proportion of missing values across the TAHMO network, only 23 stations in Kenya had sufficiently complete data to use with all of the methods (especially CMM). We train the QC methods (specifically GM and CMM) on the data from 2020, and we test QC performance on the data from 2021.
	
	\begin{figure}
		\centering
		\includegraphics[width=0.59\linewidth]{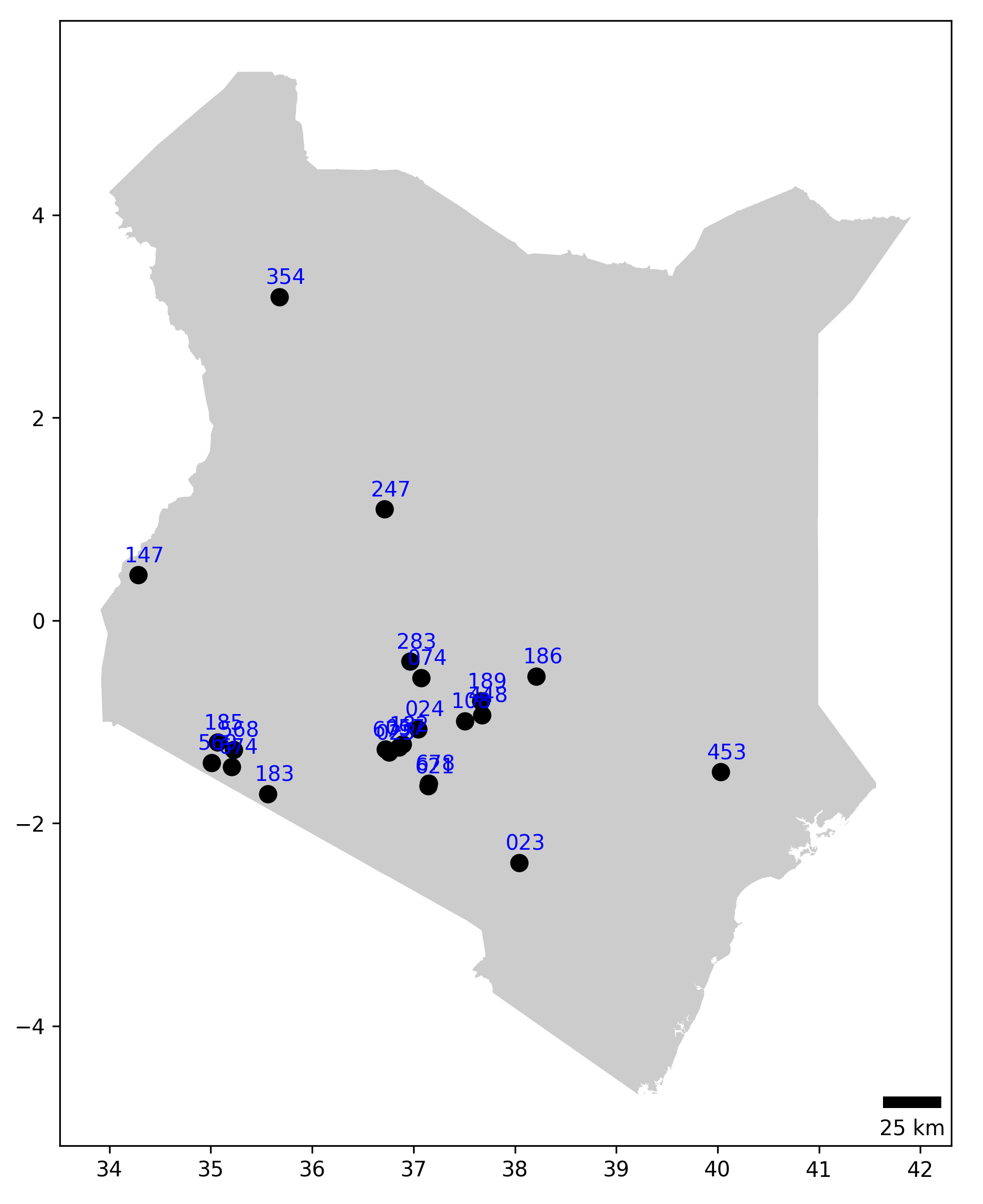}
		\caption{Geographical distribution of the 23 stations in the Kenya test dataset.}\label{fig:ch1_ok_kenya_network}
	\end{figure}

	\subsection{Oklahoma-Mesonet Dataset} \label{subsec:ch1_ok_datase}
	The Oklahoma Mesonet (\url{http://www.mesonet.org}) is a multi-purpose network established to provide research-quality data in real time. It is comprised of more than 110 surface observing stations operated by the University of Oklahoma (OU) and Oklahoma State University (OSU) (McPherson et al 2007; Brock et al. 1995; Elliott et al. 1994). Remote stations send data every 5 minutes to an operations center, located at the Oklahoma Climatological Survey (OCS), for data quality assurance. The Oklahoma Mesonet is renowned for outstanding quality assurance system, including rain gauge maintenance and calibration. When a gauge appears to be malfunctioning, the data are quickly flagged as questionable or missing, and the gauge is usually repaired or replaced within a few days \citep{mcpherson2007statewide}. We purchased the raw data (with quality flags) from Oklahoma Mesonet for the years 2008 and 2009. After discarding the stations that have flagged or missing data, we are left with a set of $81$ weather stations. We use data from 2008 for training of the QC methods and data from 2009 for testing the various QC methods.
	
	\begin{figure}[H]
		\centering
		\includegraphics[width=1.0\linewidth]{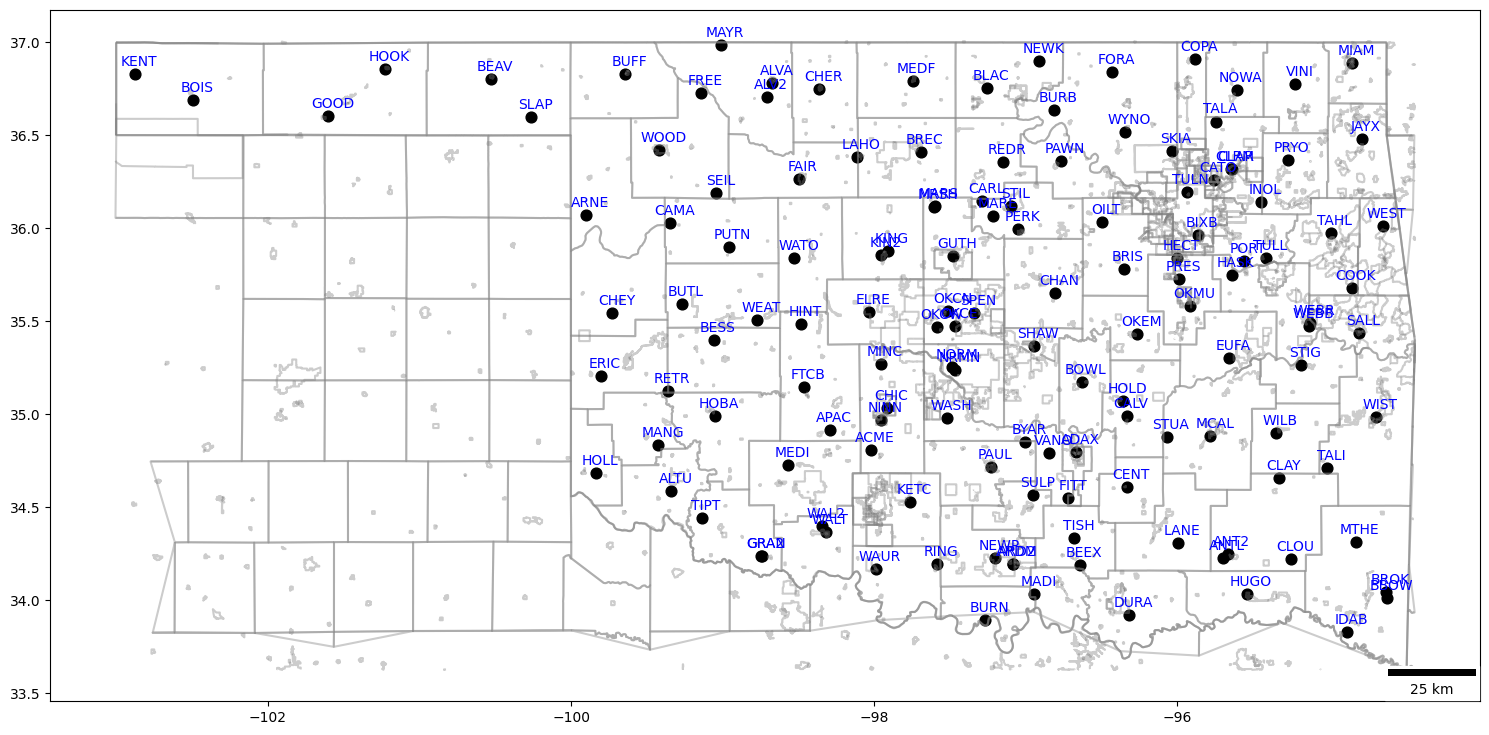}
		\caption{Geographical distribution of stations in the Oklahoma Mesonet test dataset.}
		\label{fig:ch1_ok_mesonet_network}
	\end{figure}

	\subsection{Synthetic Rain Gauge Faults}
	
	To evaluate the ability of the QC methods to detect blocked rain gauges, we insert sequences of zero values into the test data to simulate a completely blocked rain gauge. From the onset of the fault, we set all subsequent measurements for that station through the rest of the year to 0.0, under the assumption that the station's rain gauge is not repaired during the study period. We consider two configurations of faults: one where the faulty stations are randomly distributed across the network, which we refer to as ``spatially random faults'' (SRF), and a second one where the faulty stations are all neighbors of one another. We refer to the latter configuration as ``spatially clustered faults'' (SCF), and we divide it into two sub-configurations: one where the cluster of faulty stations is surrounded by working stations (SCF-IN) and one where the stations in the cluster are at the periphery of the network, which we denote by SCF-OUT.

	\section{Experiments} \label{sec:ch1_experiments}
	In this section, we describe the setup of the experiments performed, and we present the evaluation metrics used to compare the methods.
	
	\subsection{Setup} \label{subsec:ch1_exp_setup}
	Let $\mathcal{S}$ denote the set of $n$ stations $\{\text{S}_1,\ldots,\text{S}_n\}$ in the weather network, each having a rain gauge. Let $D_{\text{train}}$ be the set of historical rainfall data recorded over the network, and let $D_{\text{test}}$ be set of potentially faulty rainfall measurement for which quality control is to be performed. Our objective is to identify all the faulty sensors in $D_{\text{test}}$ in a timely manner, given $D_{\text{train}}$. 
	
	The various models are trained as follows. For KR, we fit the kernel length scale parameter $\ell$ to minimize the sum of squared errors when predicting the observed rainfall on $D_{\text{train}}$. For GM, we fit the joint Gaussian parameters $\boldsymbol{\mu}$ and $\boldsymbol{\Sigma}$ to $D_{\text{train}}$. 
	For CMM, we determine the neighboring stations $\eta(i)$ for each station $i$ by sorting the stations by their distance to $i$ (using longitude and latitude only), retaining only those whose distance is $\leq$ 100 km, and selecting the up to 5 nearest stations in that list. We then fit the CMM parameters ($\phi_i$, $\omega_i$, and $\tau_i$).
	
	\subsection{Evaluation metrics} \label{subsec:ch1_evaluation_metric}
\begin{figure}[H]
		\centering
		\includegraphics[width=0.9\linewidth]{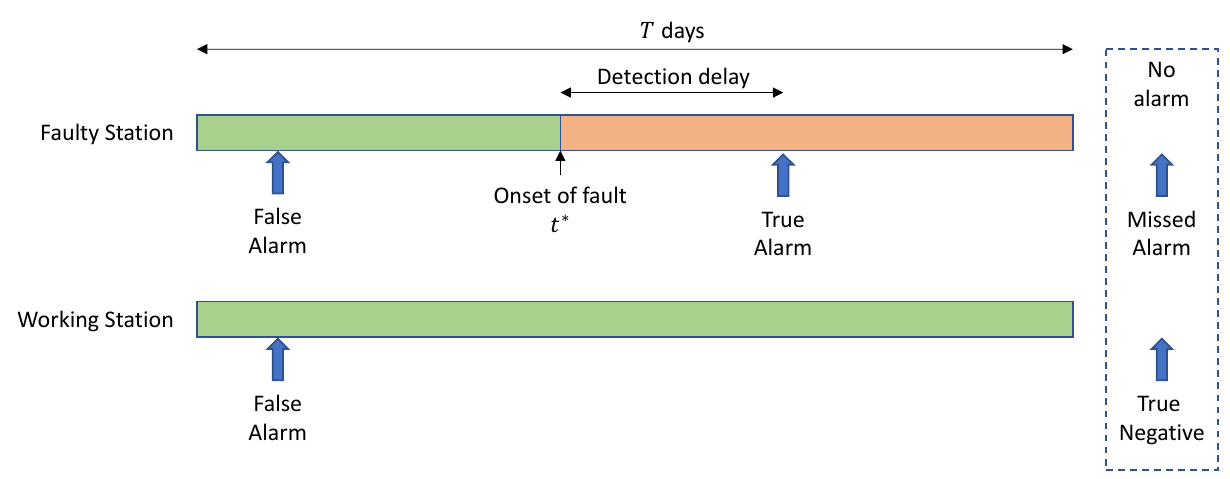}
		\caption{Definitions of TP, FP, FN, and TN. The rectangles indicate time intervals during the year. During green intervals, the station is working properly. During red intervals, it is faulty. The arrows indicate time points at which a QC method might introduce an alarm. The two arrows in the dashed box at right indicate the status if no alarm is declared during the year.}
		\label{fig:ch2-definitions}
	\end{figure}

	One might suppose that because QC methods are similar to machine learning classifiers, we could treat each day as a data point and apply the standard evaluation metrics such as accuracy, precision, recall, and Area Under the Receiver Operating Characteristic curve (AUROC) to measure QC performance. However, these are not appropriate for our network operations setting. Our goal is not to measure the expected accuracy of the methods on a daily basis, but rather to detect broken sensors as early as possible so that investigation and repair can be performed. This leads us to introduce different definitions and compute different metrics. let $t^*$ be the time of onset (start time) of the fault. Figure~\ref{fig:ch2-definitions} illustrates the new definitions. 
	\begin{itemize}
		\item False Positive Alarm. This occurs when the QC method declares an alarm before the onset of the fault for a faulty station $t^*$ or when it declares \textit{any} alarm for a working station.
		\item False Negative Alarm. This occurs when the QC method fails to declare an alarm for a faulty station (indicated in the figure by an alarm off the right end). For the false negative case, we compute the detection delay as $\Delta d = T - t^*$, the number of days between the onset of the fault $t^*$ and the end of the time period $T$. In our simulation experiments, each fault extends to the end of the year.
		\item True Positive Alarm. This occurs when the QC method declares an alarm for a faulty station at any time $t \geq t^*$, the onset of the fault. When a true positive alarm is declared, we also measure the detection delay $\Delta d = t - t^*$, which is the number of days between the onset of the fault on day $t^*$ and the alarm on day $t$.
		\item True Negative Alarm. This occurs when the QC method never declares an alarm for a working station (indicated in the figure by an alarm off the right end).
	\end{itemize}
	A QC method will typically declare multiple alarms. We consider only the earliest such alarm, as this is the key signal needed by network managers to open a trouble ticket.

	For a given setting of $\tau$ and a given QC method, we will compute three metrics: the false alarm rate (FAR), missed alarm rate (MAR), and average detection delay (ADD). To define these, let $FP(\tau), FN(\tau), TP(\tau),$ and $TN(\tau)$ be the sets of stations induced by the QC method with threshold $\tau$. Then the three metrics are defined as
	\begin{align}
		FAR(\tau) &= \frac{|FP(\tau)|}{|FP(\tau)| + |TP(\tau)|} \\
		MAR(\tau) &= \frac{|FN(\tau)|}{|FN(\tau)| + |TP(\tau)|} \\
		ADD(\tau) &= \frac{1}{|FN(\tau)| + |TP(\tau)|} \sum_{i\in FN(\tau)\cup TP(\tau)} \Delta d_i.
	\end{align}
	The FAR is the fraction of alarms that are false alarms. The MAR is the fraction of faulty stations that were false negatives. And ADD is the average detection delay for the faulty stations. 
	
	There are tradeoffs among these three metrics. If the QC method is too sensitive, it will tend to produce more false alarms but also smaller detection delays. If the QC method is not sensitive enough, it will produce longer delays and missed alarms (which are effectively infinite delays). We will visualize this tradeoff using a slightly modified version of the Activity Monitoring Operating Characteristic curve (AMOC) \cite{fawcett1999activity}. The AMOC curve plots the false alarm rate as a function of the average detection delay. For each setting of the detection threshold $\tau$ of a QC method, we compute the $FAR(tau)$ and $ADD(tau)$ and plot one point. The perfect AMOC curve detects all faults with zero delay while raising no false alarms. For the KR and GM methods, the value of $z_{min}$ (the number of days of fault evidence before an alarm is raised) is held constant in these curves. To visualize the effect of $z_{min}$, we can plot a separate AMOC curve for each value of $z_{min} = 1,\ldots,7$. The FZ-filter method does not have a detection threshold $\tau$. The only way to trade off false alarms versus missed alarms and detection delay is to vary $z_{min}$. Hence, the AMOC curves for FZ-filter are obtained by varying $z_{min}$. 
	
	Algorithm \ref{alg:amoc} presents pseudo-code for efficiently computing the AMOC curve. The basic idea is to start with $\tau=+\infty$ (conceptually) and all alarms are either TP or FN. As we iterate through the list of alarms in descending value of $\tau$, alarms can switch from TP to FP or from FN to TP to FP. As an alarm changes sets, the total delay $d_{total}$ is updated appropriately. 
	
	\begin{algorithm}
		\caption{AMOC}
		\label{alg:amoc}
		\renewcommand{\algorithmicrequire}{}
		\renewcommand{\algorithmicensure}{}
		\begin{algorithmic}[1]
			\REQUIRE \textbf{Inputs:} $A$ : set of alarm tuples $\langle i, t, s\rangle$ where $i$ indexes the station, $t$ is the date of the alarm, and $s$ is the alarm score
			\REQUIRE \quad $n$: number of stations in the weather network
			\REQUIRE \quad $T$: length of time period for the dataset
			\REQUIRE \quad $t^*_i\; i=1,\ldots,n$ time of inserted fault. If no fault is inserted, $t^*_i = T+1$
			\ENSURE \textbf{Output:} R: set of points $(FAR(\tau), ADD(\tau), MAR(\tau))$ giving the false alarm, average detection delay, and missed alarm rates for threshold $\tau$.
			\STATE \textbf{Procedure}
			\STATE \quad $a_i: i=1,\ldots,n$ time of first alarm for station $i$
			\STATE \quad $d_{total}$: total detection delay of true positives
			\STATE \quad $FP, FN, TP, TN$: number of stations in each set
			\STATE \textbf{Initializations:} All stations are either FN or TN
			\STATE \quad $a_i \gets T+1 \;\; \forall i$
			\STATE \quad $F \gets \{i : t^*_i \leq T\}$ \COMMENT{set of faulty stations}
			\STATE \quad $d_{total} \gets \sum_{i \in F} T - t^*_i$
			\STATE \quad $FN \gets |F|;  TN \gets n - FP; FP \gets 0; TP \gets 0; R \gets \{\}$
			
			\STATE Sort input $A$ in decreasing order of $s$ values
			\FOR{$\langle i, t, s\rangle \in A$}
			\IF {$i \in F$} \COMMENT{$i$ is faulty}
			\IF {$a_i = T+1$} \COMMENT{$i$ was a FN}
			\IF{$t \geq t^*_i$}  \COMMENT{FN becomes TP}
			\STATE $d_{total} \gets d_{total} -(T - t^*_i) + (t - t^*_i)$
			\STATE $FN \gets FN - 1;\; TP \gets TP + 1$
			\ELSIF{$t < t^*_i$} \COMMENT{FN becomes FP}
			\STATE $d_{total} \gets d_{total} - (T - t^*_i)$
			\STATE $FN \gets FN - 1;\; FP \gets FP + 1$
			\ENDIF
			\ELSIF {$a_i \geq t^*_i$} \COMMENT{$i$ was a TP}
			\IF {$t \geq t^*_i$} \COMMENT{$i$ remains a TP but with shorter delay}
			\STATE $d_{total} \gets d_{total} - (a_i - t^*_i) + (t - t^*_i)$
			\ELSIF{$t < t^*_i$} \COMMENT{TP becomes FP}
			\STATE $d_{total} \gets d_{total} - (a_i - t^*_i)$
			\STATE $TP \gets TP - 1;\; FP \gets FP + 1$
			\ENDIF
			\ELSIF{$a_i < t^*_i$} \COMMENT{FP remains a FP}
			\ENDIF
			\ELSE \COMMENT{$i$ is a non-faulty station}
			\IF {$a_i = T + 1$} \COMMENT{TN becomes FP}
			\STATE $TN \gets TN - 1;\; FP \gets FP + 1$
			\ELSE \COMMENT{FP remains FP}
			\ENDIF
			\ENDIF
			\STATE $a_i \gets t$
			\STATE $R \gets R \cup \{\langle FP / (FP + TP), d_{total} / (FN + TP), FN / (FN + TP) \rangle\}$
			\ENDFOR
			\RETURN{} R
		\end{algorithmic}
	\end{algorithm}

	We visualize QC performance with two additional graphs. The first, called the FAR-MAR graph, plots the false alarm rate versus the missed alarm rate as we vary the detection threshold $\tau$. The ideal method will have a setting of $\tau$ that achieves zero for both the false alarms and missed alarms. The second graph, called the MAR5 graph, plots the false alarm rate and average detection delay for the setting of $\tau$ that achieves a missed alarm rate of 5\%. Only one point is plotted for each method and each value of $z_{min}$. 
	
	\section{Results} \label{sec:ch1_results}
	In the context of network operations, we would like to have a QC method that achieves 5\% missed alarm rate and less than 20\% false alarm rate even if this leads to additional detection delay. We need the low missed alarm rate in order to achieve high data quality. We want reasonable false alarm rates, because there is a very significant cost to sending a technician to visit a weather station. An unnecessary visit is a waste of scarce resources.
	
	TAHMO's quarterly QC process using satellite data offers a 90-day detection delay, so our focus is on methods that offer 90 day or less detection delay with a less than 20\% false alarm rate. All results represent the average performance across 20 runs.
	\begin{figure}[H]
		\setlength\tabcolsep{2pt}
		\centering
		\begin{tabular}{ccc}
			\includegraphics[width=0.48\textwidth]{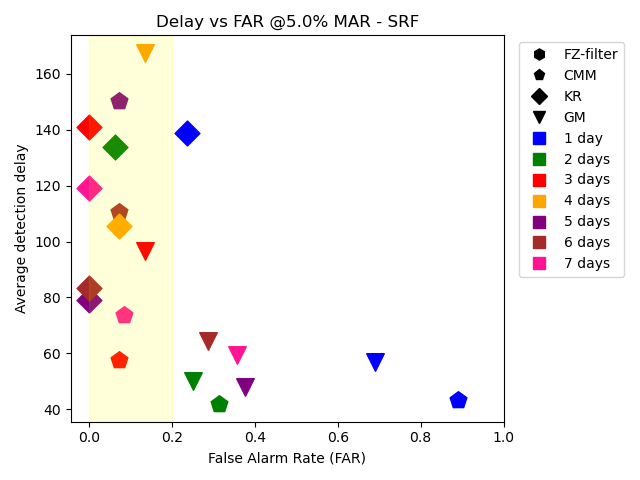} &
			\includegraphics[width=0.48\textwidth]{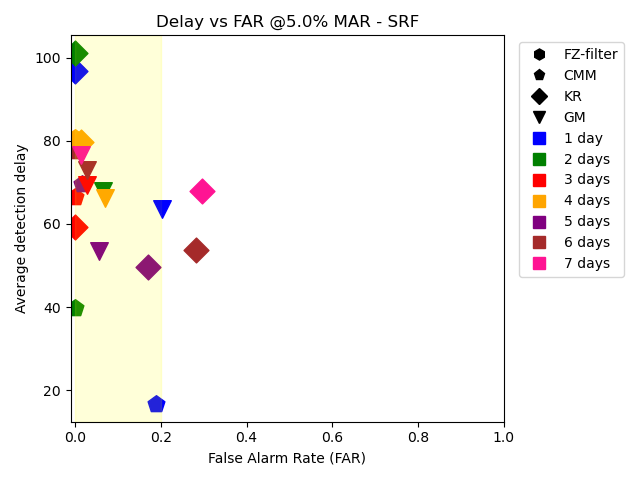} \\
			\includegraphics[width=0.48\textwidth]{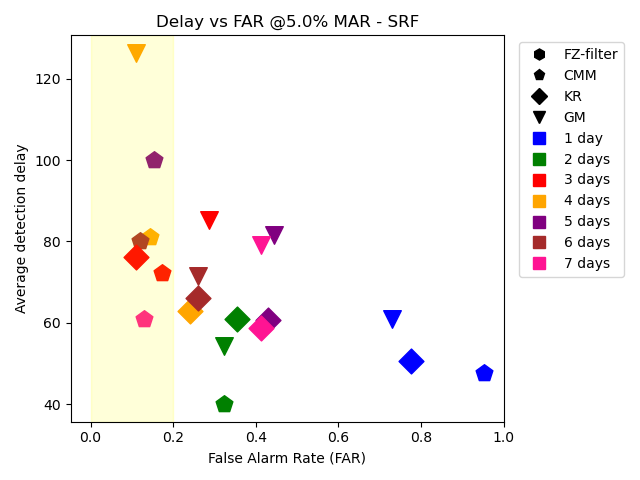} &
			\includegraphics[width=0.48\textwidth]{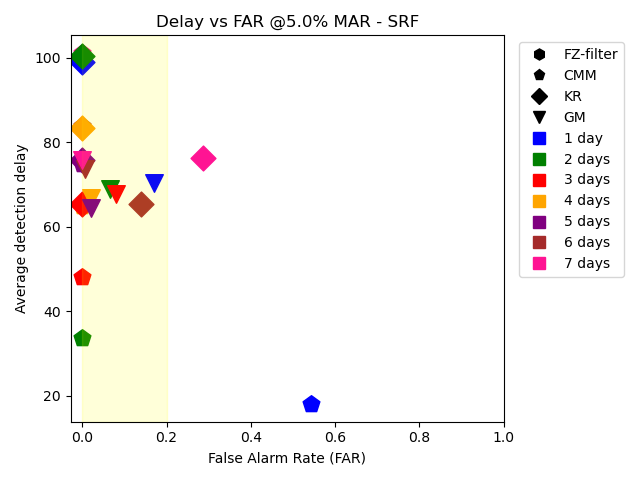} \\
		\end{tabular}
		\caption{Detection performance on the TAHMO Kenya and Oklahoma networks with 5\% and 10\% faulty rain gauges that are spatially random. Each point plots the FAR and average detection delay for a choice of $z_{min}$ that achieves 5\%MAR. Shapes correspond to different methods; colors correspond to different values of $z_{min}$. Yellow shading highlights the region of 0-20\% FAR. Plots in the left column show performance on the TAHMO Kenya network, and plots in the right column show performance on the Oklahoma network. Rows from top to bottom represent 5\% and 10\% faulty stations.}
		\label{fig:ch1_performance_MAR05_SRF_5_10}
	\end{figure}

	\subsection{Detection Delay} \label{sec:ch1_results-detection-delay}
	Performance of the algorithms were initially assessed under conditions that simulate a common real-world ``faultiness'' state of weather networks, e.g., some percentage of the total stations in the network are experiencing faults. In this set of experiments, faulty stations are chosen randomly (spatially random faults) and QC performance is assessed at four levels of faultiness, simulating 5\%, 10\% 15\%, and 20\% of the stations having faulty rain-gauges. The 5\% fault level would be considered higher than what would be typical in the Oklahoma Mesonet. There is more variation in the``faultiness'' of the TAHMO weather network, and the entire 5\% to 20\% fault range reflects rain gauge fault levels that have been experienced over time.
	
	We begin our analysis with a MAR5 plot comparing performance for 5\% and 10\% faulty stations as shown in Figure~\ref{fig:ch1_performance_MAR05_SRF_5_10}. In TAHMO 5\% and Mesonet 5\% and 10\%, there is always a setting of KR that achieves near-zero FAR and 5\% MAR. However, the detection delays can be quite high. If we consider the Pareto frontier (low and to the left), there is a setting of CMM that gives good performance in all four cases, especially in Mesonet 10\%. The GM is never on the Pareto frontier, and there are no points for the FZ filter, because there was not a setting of $z_{min}$ that achieved 5\% MAR. All methods struggle with TAHMO 10\% fault rate, which probably reflects the relatively small size of the clean data that we were able to collect.
	
	\begin{figure}[H]
		\setlength\tabcolsep{2pt}
		\centering
		\begin{tabular}{ccc}
			\includegraphics[width=0.48\textwidth]{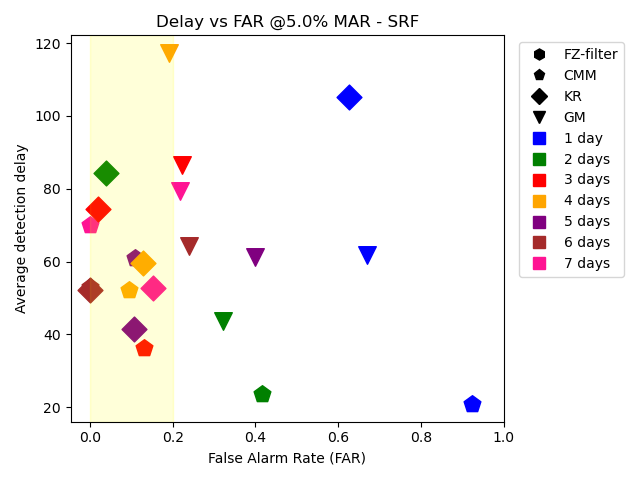} & 
			\includegraphics[width=0.48\textwidth]{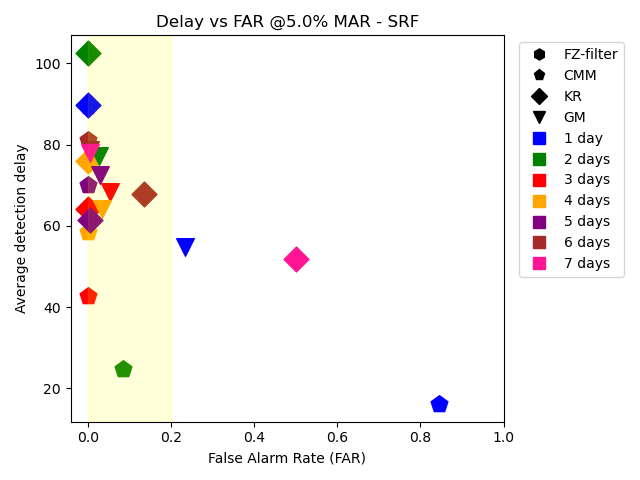} \\
			\includegraphics[width=0.48\textwidth]{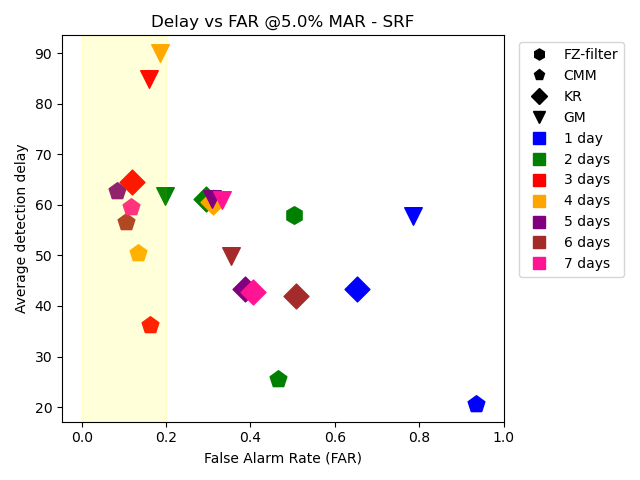} & 
			\includegraphics[width=0.48\textwidth]{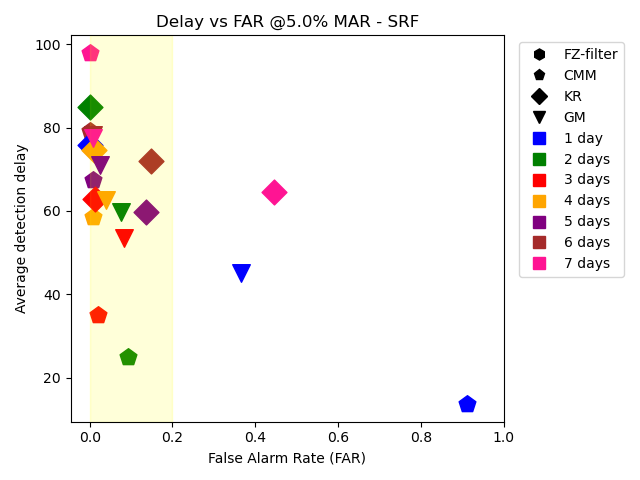} \\
		\end{tabular}
		\caption{Detection performance on the TAHMO Kenya and Oklahoma networks with 15\% and 20\% faulty rain gauges that are spatially random. Plots in the left column show performance on the TAHMO Kenya network and plots in the right column show performance on the Oklahoma network. Rows from top to bottom represent 15\% and 20\% faults.}
		\label{fig:ch1_performance_MAR05_SRF_15_20}
	\end{figure}
	
	Let us now consider the 15\% and 20\% fault rates, shown in the MAR5 graph of Figure~\ref{fig:ch1_performance_MAR05_SRF_15_20}. For TAHMO 15\%, KR achieves the best performance with zero FAR and a detection delay around 53 days. In the other three settings, CMM is the best method, and we note that even in the TAHMO 15\% case, there is a configuration of CMM that also achieves 0 FAR, but with a detection delay of 71 days. As before, GM is never on the Pareto frontier.
	
	Upon first glance, the behavior of the KR and GM methods as we change $z_{min}$ jumps around rather than following a predictable trend. The explanation for this is that the threshold for $\tau$ is tuned separately for each value of $z_{min}$. As $z_{min}$ increases, this typically increases the missed alarm rate, which requires $\tau$ to decrease to restore 5\% MAR. 
	
	\begin{figure}[H]
		\setlength\tabcolsep{2pt}
		\centering
		\begin{tabular}{ccc}
			\includegraphics[width=0.48\textwidth]{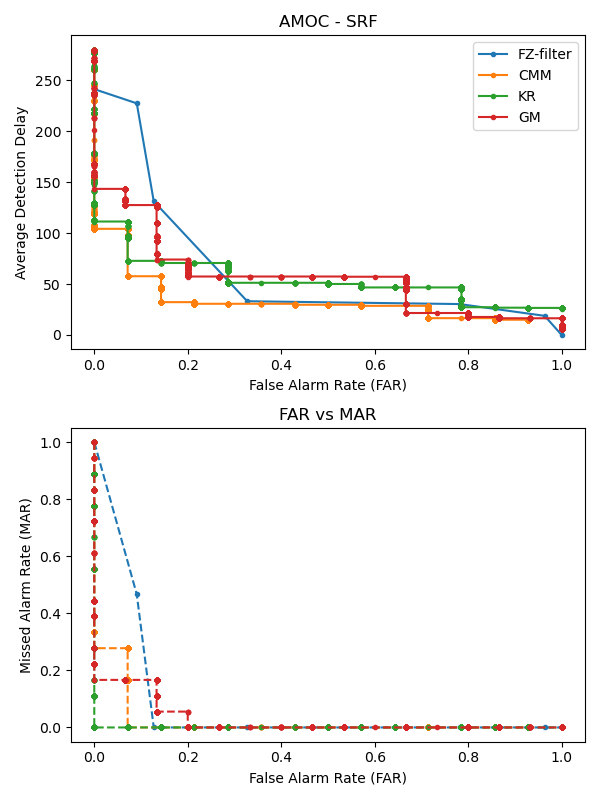} &
			\includegraphics[width=0.48\textwidth]{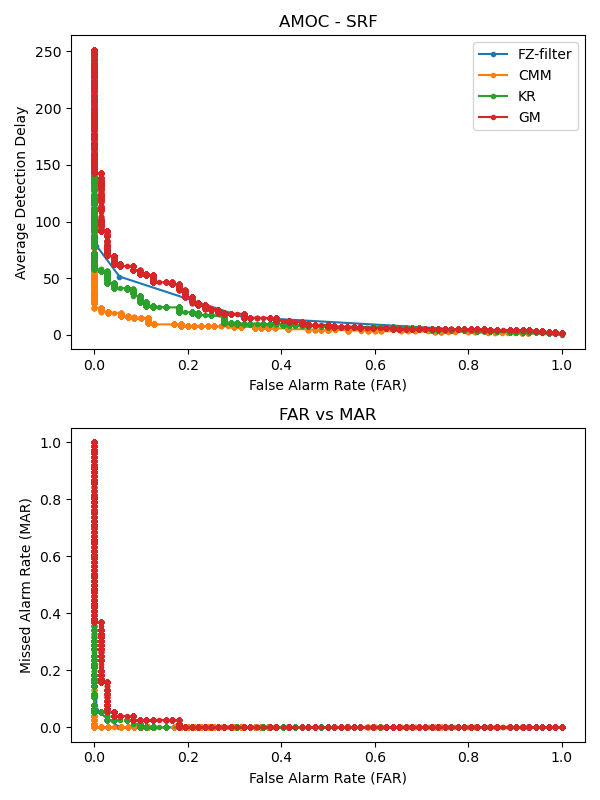} \\ 
		\end{tabular}
		\caption{AMOC and FAR-MAR trade-off of algorithm performance for the TAHMO Kenya and Oklahoma networks with 5\% faulty rain gauges that are spatially random and 3 days of evidence. Plots in the left column show performance on the TAHMO Kenya network and plots in the right column show performance on the Oklahoma network.}
		\label{fig:ch1_performance_MAR05_SRF_AMOC}
	\end{figure}
	
	We can gain insight into the behavior of FZ-filter by examining the AMOC and FAR-MAR plots. Figure~\ref{fig:ch1_performance_MAR05_SRF_AMOC} shows these graphs for $z_{min}=3$. In the TAHMO network, with its less frequent and more clustered rain events, the best performance of the FZ-filter in the FAR $\leq$ 0.20 band is at 130 days of detection delay with approximately 15\% FAR. Looking at 15\% FAR in the corresponding FAR-MAR trade-off plot, we see that the FZ-filter does not miss any alarms with a 0\% MAR. The FZ-filter performs better in the Oklahoma setting, with an in-band detection delay of about 51 days, also with a 0\% MAR. A similar process is employed for fault levels 10\%, 15\%, and 20\%.
	
	\begin{table}[H]
		\centering    
		\begin{tabular}{c|cccc|cccc}        
			Faults &  FZ$_{ok}$ &  CMM$_{ok}$ & KR$_{ok}$ & GM$_{ok}$ & FZ$_{ke}$ & CMM$_{ke}$ & KR$_{ke}$ & GM$_{ke}$ \\
			\hline                  
			5\%&   50 & \bf10 & \underline{16} & \underline{16} & 130 & \bf20 & \underline{70} & 75 \\
			10\%&  55 & \underline{12} & 13 & \bf10 & 160 & \bf50 & \underline{70} & 112 \\
			15\%&  55 & \underline{12} & 15 & \bf10 & --  & \bf25 & \underline{34} & -- \\
			20\%&  55 & \bf12 & 15 & \bf12 & --  & \bf30 & \underline{60} & 62 \\
		\end{tabular}    
		\captionsetup{justification=centering}
		\caption{Lowest detection delay at 5\%-20\% faults (FAR $\leq$ 0.20, MAR $\leq$ 0.05) for Oklahoma and Kenya networks. The boldface and underline highlight the best and the second best performing QC methods.}
		\label{tab:dd-vs-spr-fault-level}
	\end{table}
	
	\begin{table}[H]
		\centering    
		\begin{tabular}{c|cccc|cccc}        
			Delay&  FZ$_{ok}$ &  CMM$_{ok}$ & KR$_{ok}$ & GM$_{ok}$ & FZ$_{ke}$ & CMM$_{ke}$ & KR$_{ke}$ & GM$_{ke}$\\
			\hline
			15&  0.65&  0.15& 0.50&  0.60 & 0.97 & -- & 0.78 & --\\
			30&  0.22&  0.00&  0.03& 0.45 & 0.87 & 0.15 & 0.45 & --\\
			45&  0.17&  0.00&  0.03& 0.22 & 0.30 & 0.15 & 0.45 & --\\
			60&  0.03&  0.00&  0.00& 0.07 & 0.25 & 0.15 & 0.15 & 0.45\\
			75&  0.01&  0.00&  0.00& 0.04 & 0.21 & 0.15 & 0.15 & 0.40\\
			90&  0.00&  0.00&  0.00& 0.03 & 0.20 & 0.07 & 0.15 & 0.40\\
		\end{tabular}    
		\captionsetup{justification=centering}
		\caption{Lowest false alarm rate at 5\% faults, 0.05 MAR and selected average Detection Delays. Oklahoma and Kenya networks are noted by the column name subscript}
		\label{tab:dd-vs-fpr-ke}
	\end{table}
	
	By restricting our attention to configurations with FAR $\leq 0.20$ and MAR $\leq 0.05$, we can compute the best average detection delay and summarize performance in Table~\ref{tab:dd-vs-spr-fault-level}. If there is no detection delay value given, it indicates that the given algorithm could not meet the desired FAR and MAR constraints for any value of detection delay. 
	When performance is assessed in the Oklahoma setting, we can see that the FZ-filter offers the worst detection delay while the CMM, KR, and GM methods all offer similar detection delays across the various fault levels. Things are a bit different in the TAHMO Kenya setting, as not all algorithms are able to achieve points that satisfy our constraints. Generally, CMM offers the lowest delay with KR providing the next best delay. For some settings, GM offers detection delays comparable to KR, but for 10\% faults, the delay is much larger, and at 15\% faults, GM was not able to satisfy the constraints. While these results represent the averaging of multiple runs of the experiments, it may be that there are certain fault patterns that preclude better performance for some of the algorithms. In particular, the TAHMO Kenya setting may offer more challenging fault distributions for the algorithms because it has fewer stations, they are also more clustered, and there are fewer rain events.
	
	We also see in Table \ref{tab:dd-vs-fpr-ke}, presenting the lowest FAR at various detection delays when 5\% of the stations are faulty, that every automated QC method works better in the Oklahoma setting versus in the Kenya setting. While this work does not offer a thorough analysis of these differences, we posit that Oklahoma with a larger number of more uniformly distributed stations and more frequent and larger scale rain events (see figure \ref{fig:ch1_rainfall_distribution}) offers a less challenging blocked rain gauge environment for the algorithms. In the Kenya setting, there are far fewer stations and they mostly fall into one large cluster centered on the primary agricultural zone of the country. Rainfall is less frequent and is also more clustered, occurring mainly in the agricultural area and along the coast. While QC performance in the large agricultural cluster may be acceptable, the density of TAHMO stations is too low/sparse to allow any of these neighbor-based methods to succeed.

	\subsection{Robustness to Distant Neighborhoods} \label{sec:ch1_results-distance}
	When designing a weather station network, we want to know how far apart the stations can be. The non-uniform spatial distribution of weather stations in the TAHMO network may also be affecting QC performance, as some stations have few or no neighbors within an acceptable radius. In such cases, KR and GM rely on neighbors that may be quite far away, which may damage their performance. 
	
	To understand QC performance as the distance between stations increases, we experimented on both the TAHMO Kenya and Oklahoma Mesonet by constraining the selected neighbors to be no closer than a specified distance $d_{min}$ (km) from the target station. For example, when $d_{min}=10$ km, all the stations within $10$ km radius around the target station are ignored when performing quality control. This simulates a sparse network in which the fault detection signal is farther away, and hence weaker. We analyze the ADD, MAR and FAR for the values of $d_{min}$ equal to 0 km, 5 km, 10 km, 15 km, 30 km, and 50 km. 
	
\begin{figure}[H]
		\setlength\tabcolsep{2pt}
		\centering
		\begin{tabular}{ccc}
			\includegraphics[width=0.48\textwidth]{ke-delay_vs_far-SRF-0.05_MAR-0.15_faulty.png} &
			\includegraphics[width=0.48\textwidth]{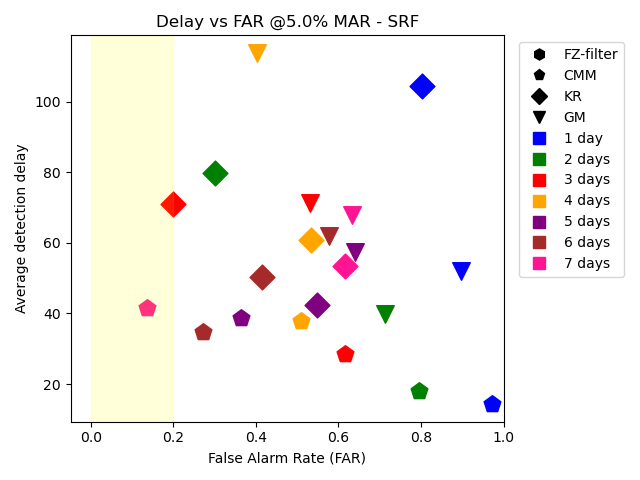} \\
			\includegraphics[width=0.48\textwidth]{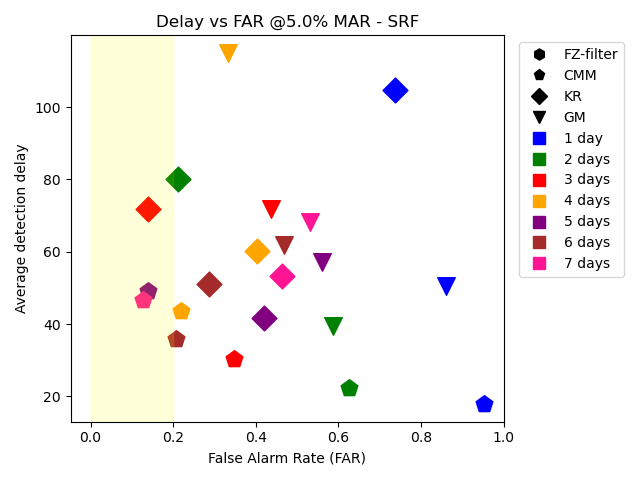} &
			\includegraphics[width=0.48\textwidth]{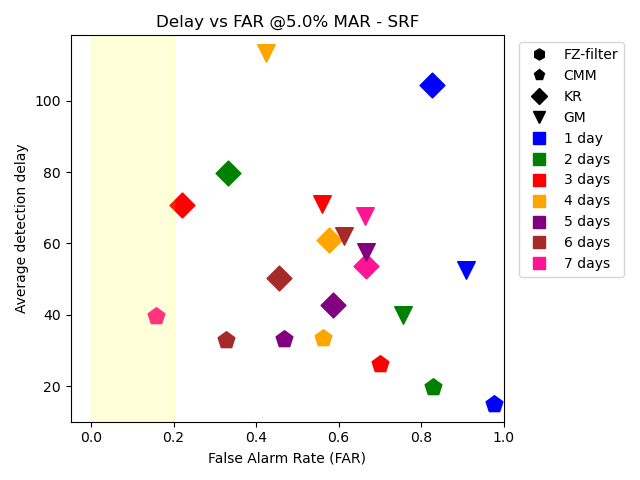} \\
			\includegraphics[width=0.48\textwidth]{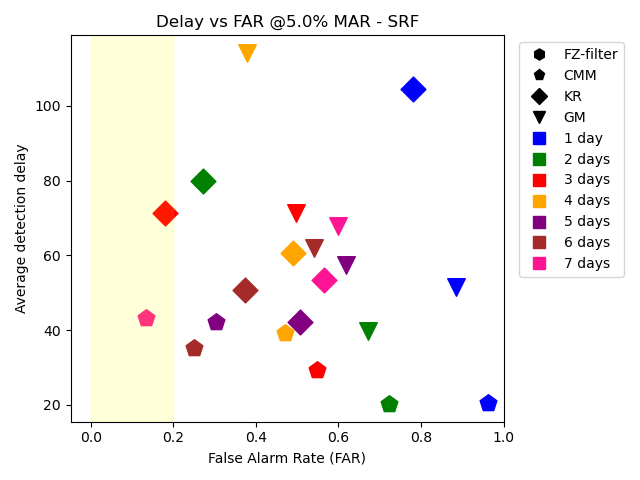} &
			\includegraphics[width=0.48\textwidth]{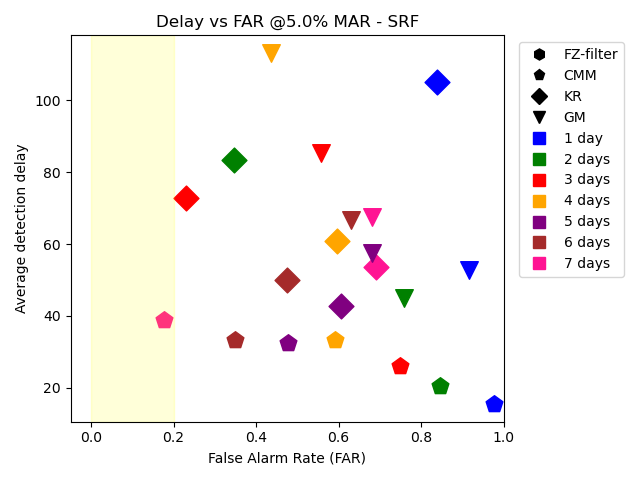} 
		\end{tabular}
		\caption{Effect of more distant neighbors on the TAHMO network. The plots on the left column top to bottom show the performance when the neighbors are chosen with minimum distance to the target station of 0 km, 5 km and 10 km, and the plots on the right (top to bottom) show performance when the neighbors are at a minimum distance of 15 km, 30 km and 50 km. Fault rate is $15\%$; spatially randomly distributed.}
		\label{fig:performance_ke_increasing_dist_neighbors}
	\end{figure}

\begin{figure}[H]
		\setlength\tabcolsep{2pt}
		\centering
		\begin{tabular}{ccc}
			\includegraphics[width=0.48\textwidth]{ok-delay_vs_far-SRF-0.05_MAR-0.15_faulty.png} &
			\includegraphics[width=0.48\textwidth]{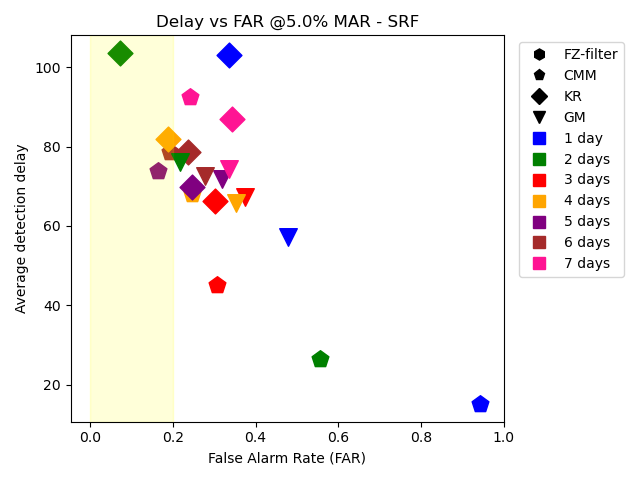} \\
			\includegraphics[width=0.48\textwidth]{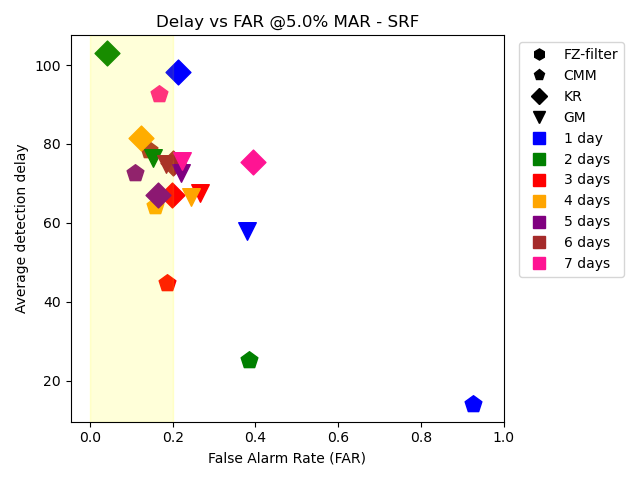} &
			\includegraphics[width=0.48\textwidth]{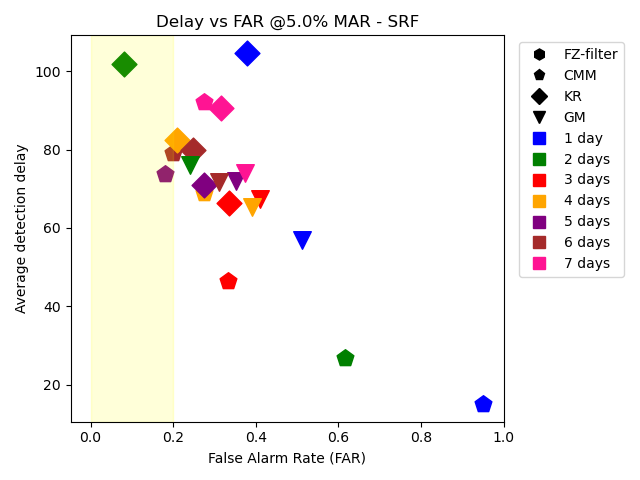} \\
			\includegraphics[width=0.48\textwidth]{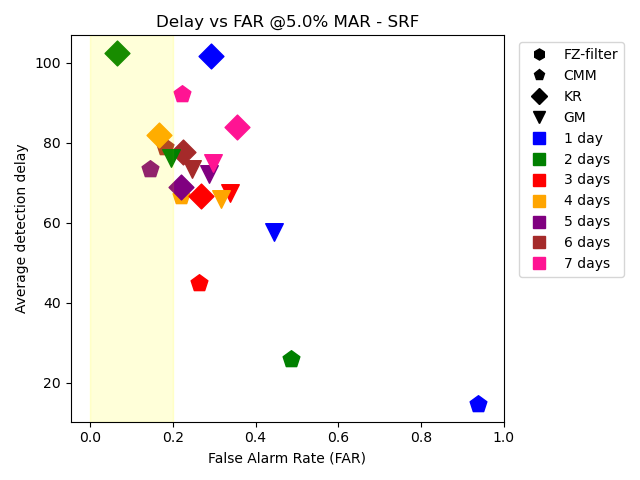} &
			\includegraphics[width=0.48\textwidth]{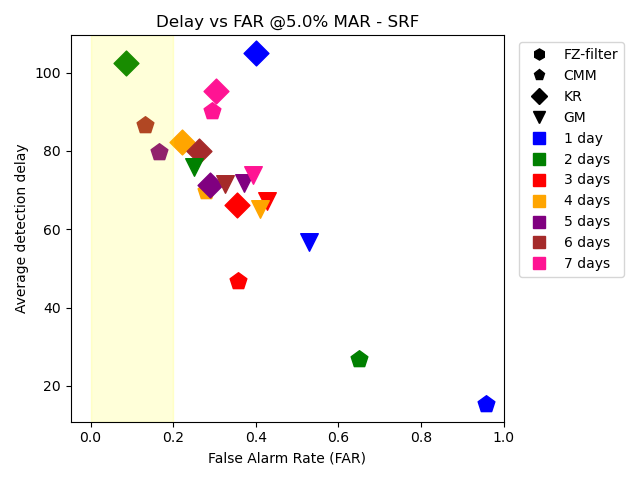} 
		\end{tabular}
		\caption{Effect of more distant neighbors on Mesonet. The plots on the left column top to bottom show the performance when the neighbors are chosen with minimum distance to the target station of 0 km, 5 km and 10 km, and the plots on the right (top to bottom) show performance when the neighbors are at a minimum distance of 15 km, 30 km and 50 km. Fault rate is $15\%$; spatially randomly distributed.}
		\label{fig:performance_ok_increasing_dist_neighbors}
	\end{figure}

	Figure \ref{fig:performance_ke_increasing_dist_neighbors} and \ref{fig:performance_ok_increasing_dist_neighbors} present the trade-off between ADD and FAR at 5\% MAR, when 15\% of the stations in both weather networks are randomly faulty. In the TAHMO Kenya network, at $d_{min}=0$ km, only CMM is functional within the FAR $\leq 20\%$ and MAR $= 5\%$ performance space, achieving 52 days ADD. At $d_{min}=5$ km, KR at 3 days of evidence and 73 days ADD is present into the constraint performance space. Above $d_{min}=10$ km (up to $d_{min}=50$ km), the detection delay for both KR and CMM remained mostly unchanged, while the FAR increased gradually with $d_{min}$. 
	
	In the Oklahoma Mesonet network, the average detection delay slightly rises as the distance increases from 0 km to 50 km. The false alarm rate on the other hand shows a more rapid increase with the distance. For instance, at $d_{min}=0$ km, all methods achieve less 70 days ADD within the FAR $\leq 10\%$ and MAR $= 5\%$ performance band, with CMM, KR and GM achieving 23 days, 62 days, and 64 days ADD respectively. At $d_{min}=5$ km, only KR with 2 days of evidence is in the performance band, with 104 days ADD (above the 90 days satellite comparison). Within the wider FAR $\leq 20\%$ and MAR $= 5\%$ performance space, CMM, KR and GM achieve 44 days, 67 days, and 73 days ADD respectively. And when $d_{min}=15$ km and above, only KR and CMM are able to perform detection within the FAR $\leq 20\%$ and MAR $= 5\%$ performance space.
	
	The above detection performance suggests that as the neighbors become more distant, the predictive power of the QC methods decreases just like the correlation between target and its neighbors.

	\begin{table}
		\centering    
		\begin{tabular}{c|cccccc}        
			Method & $0$km & $5$km & $10$km & $15$km & $30$km & $50$km \\
			\hline                  
			GM$_{ok}$ & 66 & 75 & 78 & - & - & - \\
			KR$_{ok}$ & 62 & 68 & 82 & 82 & 103 & 103 \\
			CMM$_{ok}$ & \bf 27 & \bf 46 & \bf 74 & \bf 75 & \bf 75 & \bf 78 \\
			Fz-filter$_{ok}$ & - & - & - & - & - & - \\
			\hline 
			GM$_{ke}$ & 118 & - & - & - & - & - \\
			KR$_{ke}$ & 41 & 73 & 73 & 73 & - & - \\
			CMM$_{ke}$ & \bf 37 & \bf 48 & \bf 44 & \bf 42 & \bf 40 & \bf 38 \\
			Fz-filter$_{ke}$ & - & - & - & - & - & - \\
		\end{tabular}    
		\captionsetup{justification=centering}
		\caption{Average detection delay in the Oklahoma Mesonet and the TAHMO Kenya networks for increasingly distant neighbors ($d_{min}=$0km, 5km, 10km, 15km, 30km and 50km) at a maximum MAR of 5\%, and a maximum FAR of 20\%. The best performing QC method in boldface.}
		\label{tab:ch1_ok_n_ke-distant_neighbors-srf}
	\end{table}

	\subsection{Robustness in the Presence of Clustered Anomalies} \label{sec:ch1_results-clustered-anomalies}
	Groups of sensors can become faulty simultaneously as the result of a local source of dust, a sandstorm, a violent thunderstorm, or some other natural catastrophe. We therefore need to evaluate the robustness of each QC method when groups of sensor become faulty simultaneously. To do this, we inject faults into groups of neighboring stations such that the onset of the fault is the same for all of the stations in the group. We then evaluate the ADD and FAR at $5\%$ MAR. We study two types of clusters: the first type is when all the faulty stations are surrounded by nominal stations, which we refer to as spatially clustered faults inside (SCF-IN), and the second type is when at least one of the faulty stations is at the periphery of the network. For example, a cluster of faulty stations at the Kenyan coast. We refer to the latter type of clustered faults as SCF-OUT. For the TAHMO Kenya network, we simulate clusters of size 2 to 5 (8.7\% to 21.7\% of the network), and for the Oklahoma Mesonet, we study clusters of size 2 to 8 (2.4\% to 9.8\% of the network).
	
	We did not expect any of the methods to do well with clustered faults. In particular, for clogged stations in the center of a cluster of clogged rain gauges, most of the neighbors will be clogged, so it will appear to be a dry day with all sensors working. Because all four methods analyze each station separately, there is no capability to perform iterative inference in which the stations at the edge of the cluster are diagnosed as clogged, and then when the stations at the center are scored, they would ignore the rain values reported by their immediate neighbors and attend to stations farther away.  
	
\begin{figure}[H]
		\setlength\tabcolsep{2pt}
		\centering
		\begin{tabular}{ccc}
			\includegraphics[width=0.48\textwidth]{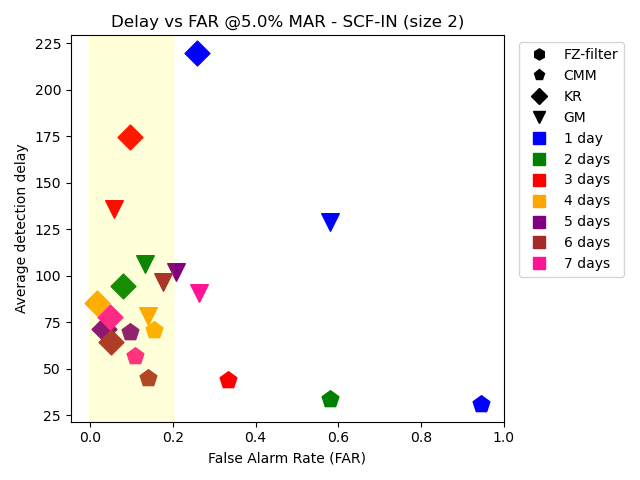} &
			\includegraphics[width=0.48\textwidth]{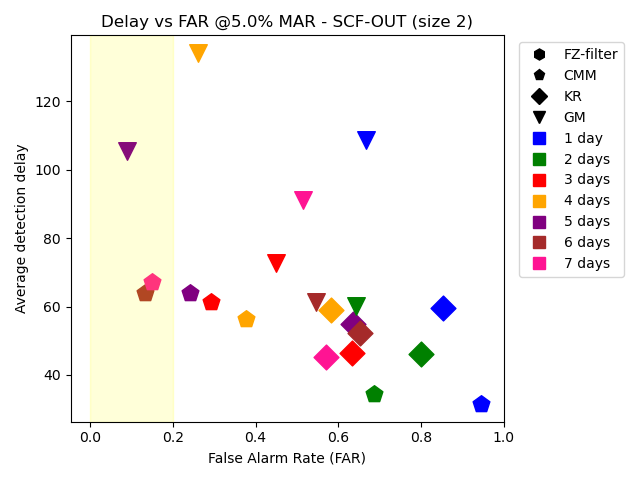} \\
			\includegraphics[width=0.48\textwidth]{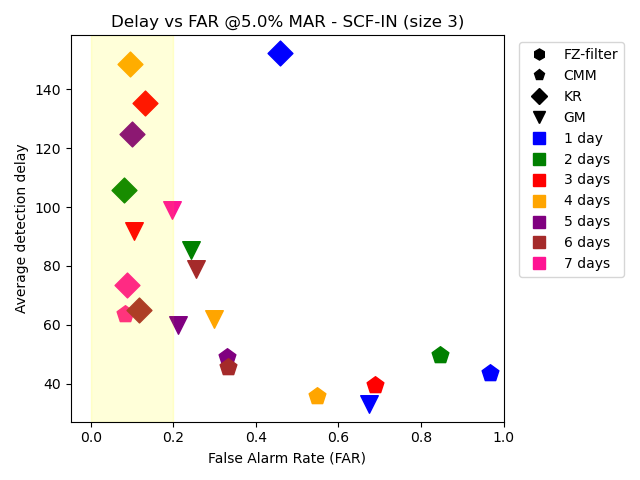} &
			\includegraphics[width=0.48\textwidth]{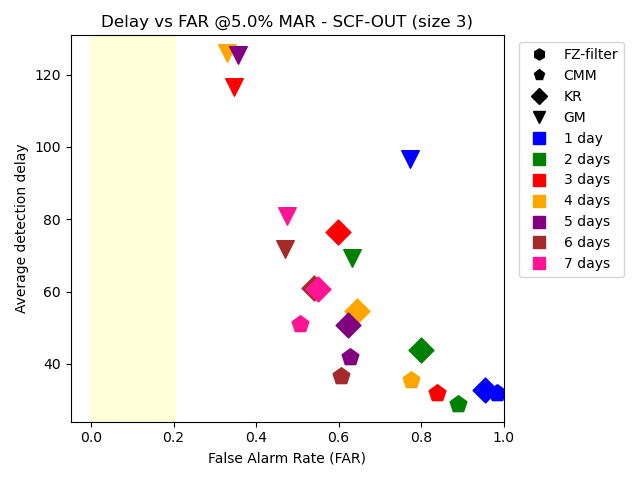} \\
			\includegraphics[width=0.48\textwidth]{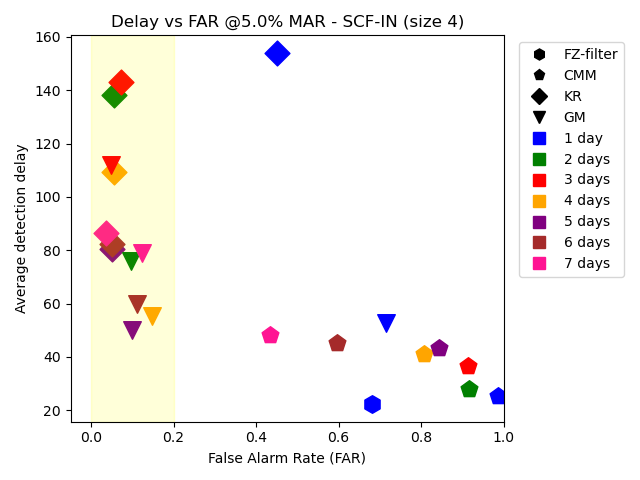} &
			\includegraphics[width=0.48\textwidth]{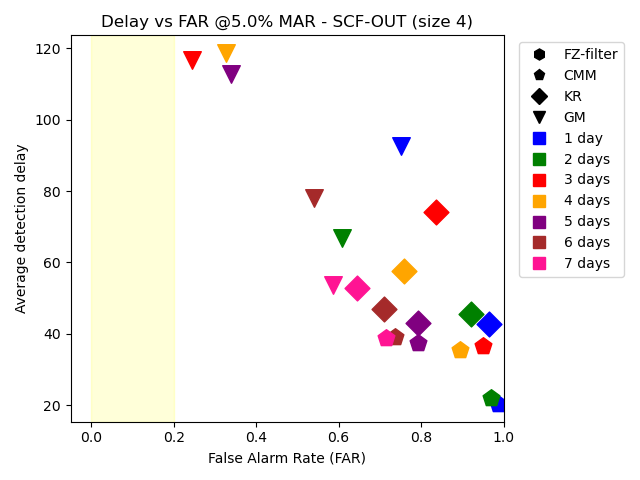} 
		\end{tabular}
		\caption{Effect of spatially-clustered faults on detection performance for the TAHMO Kenya network with $15\%$ faulty stations. The plots on the left column show the performance when the cluster of faults are inside the area (SCF-IN) and the plots on the right are when some stations are at the periphery (SCF-OUT).}
		\label{fig:performance_ke_0.15_faulty_clustered_faults}
	\end{figure}

\begin{figure}[H]
		\setlength\tabcolsep{2pt}
		\centering
		\begin{tabular}{ccc}
			\includegraphics[width=0.48\textwidth]{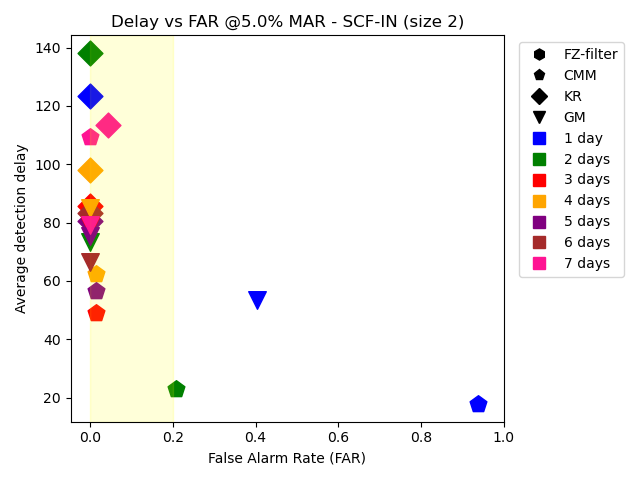} &
			\includegraphics[width=0.48\textwidth]{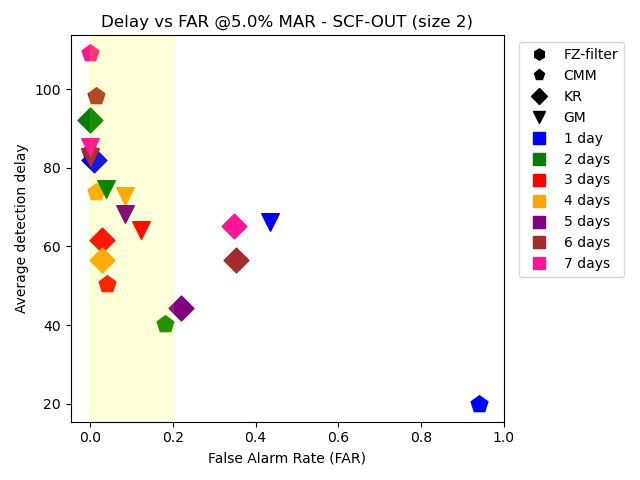} \\
			\\
			\includegraphics[width=0.48\textwidth]{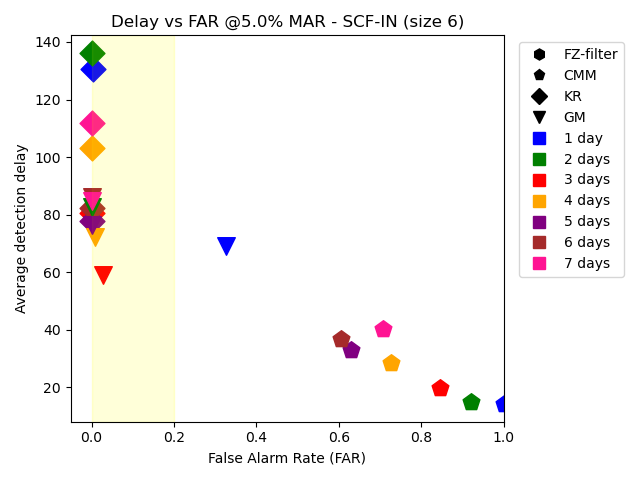} &
			\includegraphics[width=0.48\textwidth]{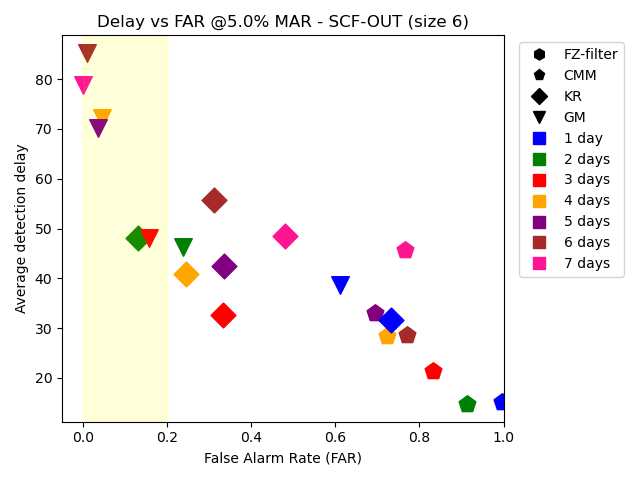} \\
			\includegraphics[width=0.48\textwidth]{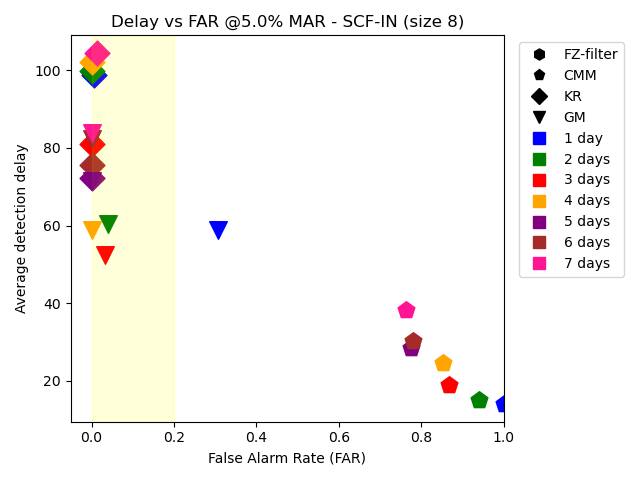} &
			\includegraphics[width=0.48\textwidth]{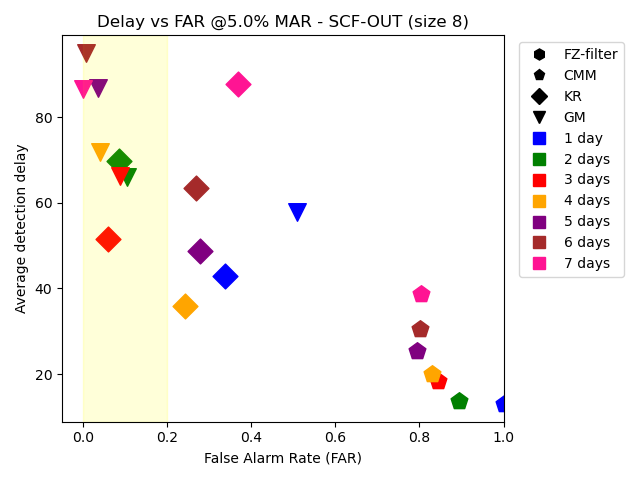} 
		\end{tabular}
		\caption{Effect of spatially-clustered faults on the detection performance for the Oklahoma Mesonet network with $15\%$ faulty stations. The plots on the left column show the performance when the cluster of faults are inside the area (SCF-IN) and the plots on the right are when some stations are at the periphery (SCF-OUT).}
		\label{fig:performance_ok_0.15_faulty_clustered_faults}
	\end{figure}

	The results for clustered faults in the TAHMO Kenya network and the Oklahoma Mesonet are presented in Figures~\ref{fig:performance_ke_0.15_faulty_clustered_faults} and \ref{fig:performance_ok_0.15_faulty_clustered_faults} respectively.  The first surprise is that in both networks, as long as the clusters are SCF-IN clusters in the interior of the networks, performance is still quite good. The false alarm rates have increased slightly compared to spatially-random faults, but KR and CMM still perform quite well, and both appear on the Pareto frontier within the region of interest. A second surprise is that when the clusters become big, the GM method achieves the lowest ADD while maintaining low FAR. This may reflect its more sophisticated handling of covariances compared to KR. Because KR normalizes its weights, even distant stations can be given high weight, whereas in the joint Gaussian distribution, distant stations always have low weight and the conditional variance increases. However, we expected to see this phenomenon in the minimum-distance experiments of the preceding section, but we did not. 
	
	The SCF-OUT setting is much more difficult. In Kenya, only clusters of size 2 can be handled at all, and this is with significant false alarm rates. For clusters of size 3 and 4, no method can achieve FAR $\leq 0.2$. 
	
	The story in Oklahoma is similar. For SCF-IN clusters, all methods achieve low false alarm rates, and GM again excels for larger clusters (size 6 and 8). For SCF-OUT clusters, the situation is better than in Kenya. For small clusters, CMM and KR lie on the Pareto frontier. But for larger clusters, CMM fails, and KR is generally the best method. 
	
	Overall, we notice that CMM, which uses only up to 5 nearest neighbors for fault detection, is strong when the cluster size is small. But as the cluster size increases and more neighboring stations become affected, we see that CMM falls apart. We suspect that CMM may still be more robust if we increased the number of neighbors that it can use. However, such an increase of neighbors will lead to a decline in the number of stations that can be diagnosed due to the problem of missing data. This highlights the trade-off between number of stations that can be diagnosed and the number of neighbors. Once the size of the SCF-OUT faults reaches 5 stations, CMM doesn't have any signal to draw on. Only global/dynamic methods such as KR and GM can pull signal from the larger network to diagnose target stations. Additionally, we observe in general that in both TAHMO Kenya and Oklahoma Mesonet, as the size of the cluster increases, so does the average detection delay needed to achieve 5\% MAR.
	

	\begin{table}
		\centering    
		\begin{tabular}{c|ccccccc|ccccccc}  
			& \multicolumn{7}{c|}{FAR $\leq 20\%$} & \multicolumn{7}{c}{FAR $\leq 10\%$} \\ 
			Method & 2 & 3 & 4 & 5 & 6 & 7 & 8 & 2 & 3 & 4 & 5 & 6 & 7 & 8 \\
			\hline                
			GM$_{IN}$ & \underline{68} & \underline{57} & \underline{76} & \underline{52} & \bf 60 & \bf 55 & \bf 54 & 68 & 57 & 76 & \underline{52} & \bf 60 & \bf 55 & \bf 54 \\
			KR$_{IN}$ & 84 & 76 & 88 & 70 & \underline{77} & \underline{83} & \underline{74} & 84 & 76 & 88 & 70 & \underline{77} & \underline{83} & \underline{74} \\
			CMM$_{IN}$ & \bf50 & \bf37 & \bf37 & \bf38 & - & - & - & \bf50 & \bf37 & \bf67 & \bf 46 & - & - & - \\
			Fz-filter$_{IN}$ & - & - & - & - & - & - & - & - & - & - & - & - & - & - \\
			\hline 
			GM$_{OUT}$ & 65 &\ \bf 53 & 65 & 55 & \bf 48 & \underline{55} & \underline{67} & 70 & \bf 75 & 65 & 62 & \bf 69 & \bf 72 & \bf 67 \\
			KR$_{OUT}$ & \underline{57} & - & \underline{43} & \underline{55} & \bf 48 & \bf33 & \bf 52 & \underline{57} & - & \bf43 & \bf 55 & - & \underline{83} & - \\
			CMM$_{OUT}$ & \bf40 & \underline{55} & \bf36 & \bf 48 & - & - & - & \bf 50 & \underline{82} & \underline{50} & \underline{60} & - & - & - \\
			Fz-filter$_{OUT}$ & - & - & - & - & - & - & - & - & - & - & - & - & - & - \\
		\end{tabular}    
		\captionsetup{justification=centering}
		\caption{Average detection delay at a maximum MAR of 5\%, and a maximum FAR of 20\% for spatially clustered faults SCF-IN and SCF-OUT in the Oklahoma network. The columns show the cluster size and the rows are the QC methods. The boldface and underline highlight the best and the second best performing QC methods.}
		\label{tab:ch1_ok-add_vs_far-scf}
	\end{table}
	
	\begin{table}
		\centering    
		\begin{tabular}{c|cccc|cccc|cccc}   
			& \multicolumn{4}{c|}{FAR $\leq 20\%$} & \multicolumn{4}{c|}{FAR $\leq 10\%$} & \multicolumn{4}{c}{FAR $\leq 5\%$} \\
			Method & 2 & 3 & 4 & 5 & 2 & 3 & 4 & 5 & 2 & 3 & 4 & 5 \\
			\hline                  
			GM$_{IN}$ & 81 & 93 & \bf49 & 104 & 137 & 93 & 113 & 161 & - & - & - &  \\
			KR$_{IN}$ & \underline{65} & \underline{65} & \underline{64} & \bf 87 & \bf 65 & \underline{73} & \bf 64 & 102 & \bf 71 & - & \bf 83 & 102 \\
			CMM$_{IN}$ & \bf45 & \bf 63 & - & - & \underline{70} & \bf63 & - & - & - & - & - & - \\
			Fz-filter$_{IN}$ & - & - & - & - & - & - & - & - & - & - &  &  \\
			\hline 
			GM$_{OUT}$ & 108 & - & - & - & 108 & - & - & - & - & - & - & - \\
			KR$_{OUT}$ & - & - & - & - & - & - & - & - & - & - & - & - \\
			CMM$_{OUT}$ & \bf 66 & - & - & - & - & - & - & - & - & - & - & - \\
			Fz-filter$_{OUT}$ & - & - & - & - & - & - & - & - & - & - & - & - \\
		\end{tabular}    
		\captionsetup{justification=centering}
		\caption{Average detection delay at a maximum MAR of 5\%, and a maximum FAR of 20\% for spatially clustered faults SCF-IN and SCF-OUT in the TAHMO Kenya network. The columns show the cluster size and the rows are the QC methods. The boldface and underline highlight the best and the second best performing QC methods.}
		\label{tab:ch1_ke-add_vs_far-scf}
	\end{table}

	\section{Discussion and Conclusion}  \label{sec:ch1_conclusion}
	
	Timeliness in the detection of broken rain gauges is important. Flood and drought indicators/alarms are more valuable when the data is accurate and reflects recent precipitation events. Crop insurance and other agricultural programs may depend on accurate rain measurement in specific locales. Timeliness of broken gauge detection impacts timeliness of repair. As budgets are limited, false alarms must be reduced to minimize repair costs. The Oklahoma Mesonet is able to detect blocked rain gauges rapidly with a low false alarm rate by comparing the sensor-reported rainfall with real-time Doppler radar estimates of rainfall. Additionally, Oklahoma Mesonet sensors are regularly rotated, with sensors in the field brought back to the lab, cleaned and recalibrated, and then redeployed to replace sensors in a different location.

	One of the central questions of this work is how well we can detect blocked rain gauges using only the sensor readings themselves. The results show that for spatially-random faults, CMM can achieve detection delays of $\leq 30$ days for most configurations in both TAHMO Kenya and Oklahoma Mesonet. KR is also very strong on Mesonet, but its detection delays are in the 34-70 range on TAHMO Kenya (see Table~\ref{tab:dd-vs-spr-fault-level}). One factor affecting performance is the distance between sensors in the network and the degree to which rain events are wide-spread versus being more local. Rainfall events in Kenya are more local, and many of the TAHMO stations are close together (see Figure~\ref{fig:ch1_rainfall_distribution}). This gives acceptable performance. Rainfall events in Oklahoma are more global, and the Mesonet stations are farther apart. This also gives good performance. Our experiments with removing nearby stations show that if the Mesonet stations were further apart, none of our methods work very well, despite the global nature of the rainfall events.

	The experiments in Section \ref{sec:ch1_results-distance} explored the performance differences between algorithms that use predominantly local information (the CMM), often set at training time, and algorithms that are able to use the entire network of stations. Essentially these experiments simulate scenarios in which a target station's local neighbors, within some distance $d$, are not available. The question then arises of how informative the more distant neighbors are in helping to detect that the target's rain gauge is blocked. It also follows that, as there is less local information available to the algorithms, there would also need to be larger scale rain events (or multiple scattered events) to provide useful data to the algorithms. For instance, if we are limiting a target's neighbors to those $\geq$ 50 km from the target and the only rain event during the testing period is a 5 km-diameter rain event at a neighbor 10 km from the target, the more distant neighbors will be uninformative (as is the excluded neighbor).
	
\begin{figure}
		\setlength\tabcolsep{2pt}
		\centering
		\begin{tabular}{ccc}
			\includegraphics[width=0.48\textwidth]{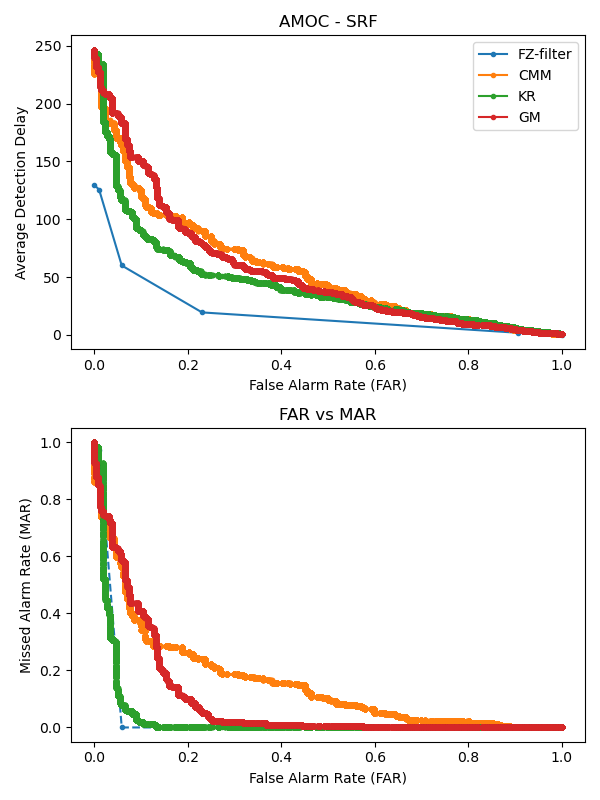} &
			\includegraphics[width=0.48\textwidth]{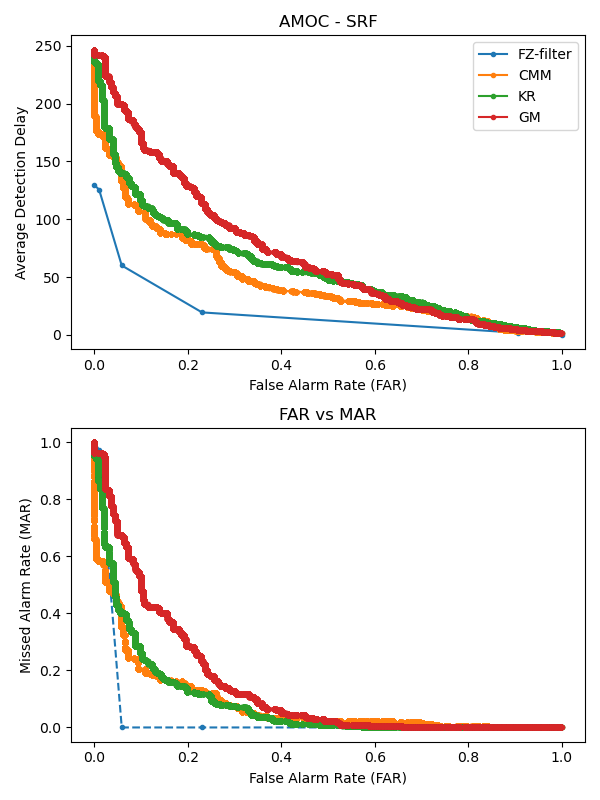} \\
		\end{tabular}
		\caption{AMOC and FAR-MAR trade-off of algorithm performance for the Oklahoma network with 15\% faulty rain gauges that are spatially random, neighbors $\leq$ 30 km are excluded, for two and three days of evidence. Plots in the left column show performance with two days of evidence and and plots in the right column show performance with three days of evidence}
		\label{fig:performance_MAR15_SRF_30KM_AMOC}
	\end{figure}

	As with the fault level results in Section \ref{sec:ch1_results-detection-delay}, there are additional ADD vs.~MAR vs.~FAR trade-off points available outside of the fixed 5\% MAR plots. Figure \ref{fig:performance_MAR15_SRF_30KM_AMOC} highlights not just the FAR vs. MAR trade-off, but also the effect of evidence on an algorithm being able to provide satisfying performance points in the FAR $\leq$ 0.20, MAR $\leq$ 5\% constraint space. First, with two days of evidence, there are two algorithms, FZ-filter and KR, that offer points in the constraint space. FZ-filter detects at 60 days of delay with about 6\% FAR and 0\% MAR, while KR offers a similar 60 day detection delay, a higher 20\% FAR at the same 0\% MAR. 
	
	For the detection of clustered faults, the plots in Figure \ref{fig:performance_ke_0.15_faulty_clustered_faults} show that KR and GM work well in SCF-IN. But in the TAHMO Kenya network, no QC method is able to work well. This suggest that, because the rain events are more local, we need a higher density of weather stations along the periphery (especially the coast). For example, stations in Tanzania and Uganda may be used when diagnosing faults in sensors along the international borkers with those countries. Additionally, we note that CMM is particularly vulnerable to clustered faults, because of its fixed neighborhood definition.

\begin{figure}
		\centering
		\begin{tabular}{ccc}
			\includegraphics[width=0.50\textwidth]{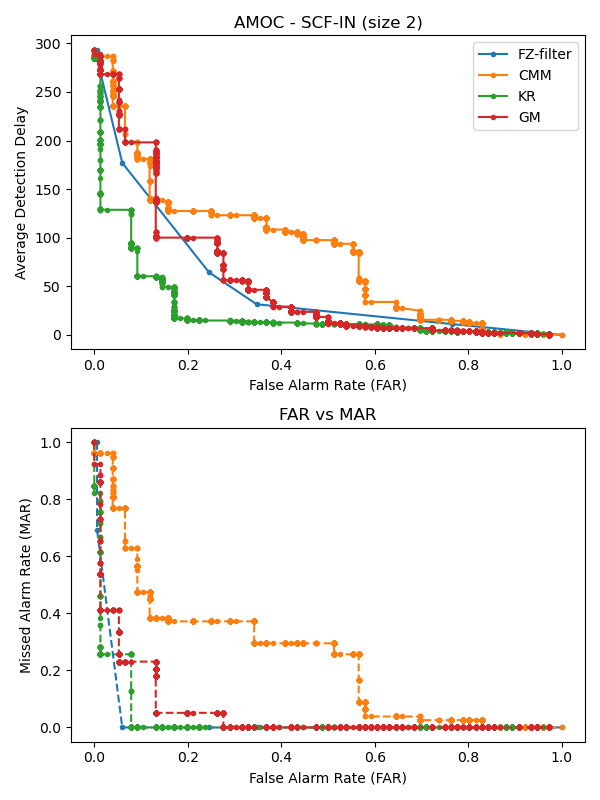} \\
		\end{tabular}
		\caption{AMOC and FAR-MAR trade-off of algorithm performance for the Kenya network with 15\% faulty rain gauges that are spatially clustered with size=2 in the interior of the network, for two days of evidence.}
		\label{fig:performance-cluster-size-2-ke-two-days}
	\end{figure}

	As we've found with most experiments, examination of MAR $\leq$ 5\% instead of fixed 5\% MAR yields additional satisfying points. Figure \ref{fig:performance-cluster-size-2-ke-two-days} shows that the KR algorithm is able to detect blocked gauges with about 15 days of detection delay with about 15\% FAR and 0\% MAR. This offers a trade-off versus the satisfying point at 5\% MAR that offers a 10\% FAR---but at about 90 days of detection delay. We also see that FZ-filter offers a satisfying point in the constraint space with 0.05 FAR, 0\% MAR, but an incredible 175 days of detection delay.
	
	As noted previously, most of our our experiments fixed MAR at 5\%  so that results were more easily comparable. As we've already discussed, it is often worthwhile to examine the entirety of the constraint space, e.g., FAR $\leq$ 20\%,  MAR $\leq$ 5\%, for satisfying points. As there are variable real-world costs associated with MAR, FAR, ADD, there are opportunities to select trade-off points that are adapted to the needs of specific regions, seasons, emergency response policies, insurance risk models, etc.
	
\begin{figure}
		\setlength\tabcolsep{2pt}
		\centering
		\begin{tabular}{ccc}
			\includegraphics[width=0.48\textwidth]{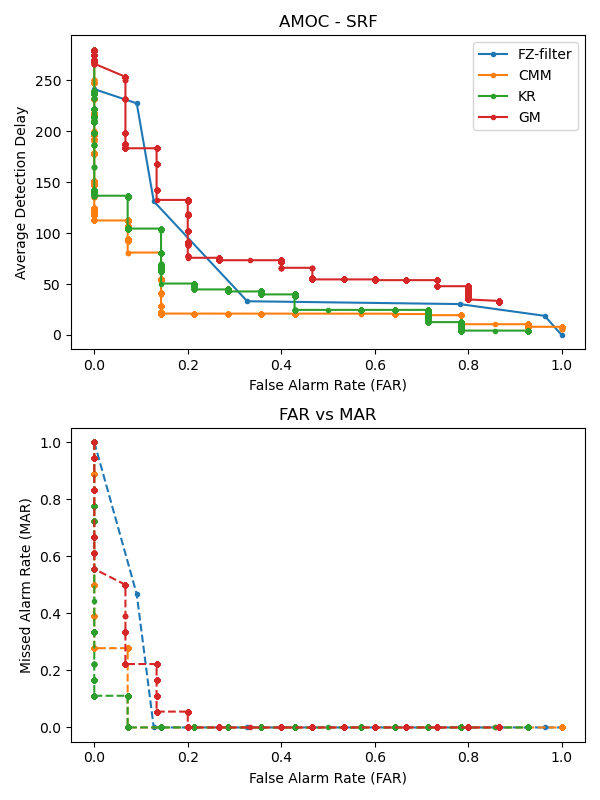} &
			\includegraphics[width=0.48\textwidth]{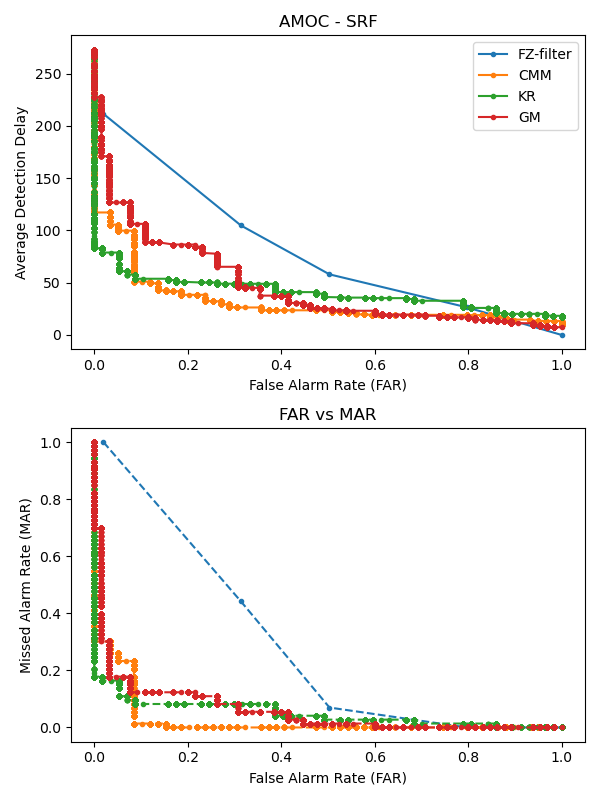} \\
		\end{tabular}
		\caption{AMOC and FAR-MAR trade-off of algorithm performance TAHMO Kenya network. The left hand plots represent 5\% faulty rain gauges that are spatially random with four days of evidence. The right hand plots represent 20\% faulty rain gauges that are spatially random with five days of evidence.}
		\label{fig:performance_AMOC_KE-SRF05-04-SRF20-05}
	\end{figure}

	The results of our experiments show that CMM achieves good detection delay with multiple days of evidence, despite the fact that  currently-deployed version alarms with only one day of evidence in TAHMO. Although one day of evidence generally produces lower detection delay, it also generates a much higher false alarm rate, mostly larger than $20\%$. For instance, in regions of higher station reliability (5\% faulty), using four days of evidence would significantly reduce the FAR to about 15\%, at the cost of approximately 25-30 days of detection delay (see Figure \ref{fig:performance_AMOC_KE-SRF05-04-SRF20-05}, upper left plot). At this 15\% FAR, we wouldn't miss any alarms as the  corresponding MAR is 0\% (see Figure \ref{fig:ch1_performance_MAR05_SRF_AMOC}, lower left plot). While the CMM FAR is 50\% higher than that of the manual satellite QC process, the reduction in detection delay from 90 days for the satellite to 30 days for the CMM method might be a worthwhile trade-off, at least in certain seasons and/or regions where there are higher risks associated with faulty rain gauges. In regions of lower station reliability (20\% faulty), we also find improvements over the currently-deployed system when we require five days of evidence. The FAR drops to about 15\% with approximately 1\% MAR, at the cost of about 50 days of detection delay (see Figure \ref{fig:performance_AMOC_KE-SRF05-04-SRF20-05}, upper and lower plots on the right). In this setting of more faulty stations, the CMM detection delay is still roughly half that of the satellite, and the reduced detection delay may offer similar benefits in exchange for more false alarms.
	
	Although we constrained our analysis of TAHMO network data to the stations in Kenya, TAHMO is a continent-scale network with over 700 stations. Many of these stations are widely dispersed, and a significant fraction of them are poorly maintained. Even more troublesome is that many stations are offline for extended periods of time because of battery or network connectivity problems. This means that the neighborhood structure of the network is not stable, so methods such as CMM that require such stability, do not work well. This suggests that KR and GM might be a better choice for many parts of the continent.

	The CMM algorithm provides a good example of a local information model that performs reasonably well under benign conditions in both the Oklahoma and TAHMO Kenya settings. The KR and GM models leverage global information and provide value when the rate of local failures is high for a given target station and local information is constrained or not available at all.

	\section{Acknowledgements} \label{sec:ch1_acknowledgements}
	We would like to acknowledge Michael Slater and Austin Kaburia for their contributions in preparing the TAHMO dataset used in this study.

	\phantom{}\clearpage
	\phantom{}\vfill
	\begin{center}
		\heading
		Integrating Probabilistic Consistency-Based Diagnosis and Joint Inference for Advanced Weather Sensor Diagnosis
	\end{center}
	\vfill
	Iren\'{e} Tematelewo and Thomas G. Dietterich
	\vfill\noindent

	\vfill
	\chapter{Integrating Probabilistic Consistency-Based Diagnosis and Joint Inference for Advanced Weather Sensor Diagnosis} \label{ch2:pcsd}

	\section{Abstract}
	Sensor networks have become instrumental in applications ranging from weather forecasting to resource exploration, thanks to their ability to collect vast amounts of high-resolution data. Despite their advantages, these networks face significant challenges related to data quality, which can be compromised due to environmental conditions, network transmission issues, and technical errors. Effective quality control (QC) mechanisms are crucial for ensuring data reliability, particularly in precipitation sensor networks, which are vital for regions dependent on rainfed agriculture.
	
	Traditional QC methods fall into two categories: rule-based and probabilistic. Rule-based methods are simple, efficient, and easy to implement but they cannot adapt to the pattern of behavior of each individual sensor. For instance, a value that is high for one location may be normal for another location. In consequence, range checks need to be very conservative to avoid causing many false alarms. Probabilistic methods on the other hand, despite being more complex, excel at detecting clusters of faulty sensors through joint reasoning, making them valuable for identifying anomalies across multiple sensors. However, both approaches have their drawbacks, including susceptibility to false alarms and missed detections.
	
	This paper introduces two general-purpose methods for sensor diagnosis that leverage joint reasoning to address the limitations of existing QC methods, particularly in scenarios where clusters of sensors may fail simultaneously. These methods aim to enhance the detection of erroneous sensor readings and improve the overall reliability of sensor network data.

	\section{Introduction} \label{sec:ch2_intro}
	
	Sensor networks have revolutionized research and predictions in various applications such as weather forecasting, environmental study, natural hazards study and resource exploration \citep{campbell2013quantity}. These networks facilitate the collection of massive amounts of data at fine resolution, offering significant benefits. However, these advantages come with the risk of compromised data quality if robust quality control mechanisms are not in place.
	
	Throughout the data collection process, the integrity of the data can be threatened at different phases. For instance, environmental conditions, such as extreme weather, can damage or disrupt sensors, network transmission issues may corrupt the data during transfer, and technical errors or hardware degradation can introduce noise and inaccuracies \cite{dereszynski2012probabilistic}. Traditionally, providing raw data with limited or no quality checks to end consumers or the public has been a common practice in the ecological community \cite{campbell2013quantity}. However, without proper quality control (QC), utilizing such data for prediction or decision making can yield misleading or consequential results. While expert knowledge can be leveraged to identify certain abnormal sensor behaviors, the scale of modern sensor networks renders manual QC impractical. Therefore automated QC procedures are essential for detecting erroneous sensor readings and ensuring data reliability. 
	
	The significance of QC in weather sensor networks, especially in precipitation sensor networks (or rain gauge networks), cannot be overstated, given the pivotal role of rainfall for humanity. Many countries in Africa, Latin America and South Asia heavily depend on rainfed agriculture for their staple food \citep{rockstrom2015agriculture, wani2009rainfed}. QC for rainfall data presents unique challenges: (i) unlike other weather variables (e.g., temperature, relative humidity, atmospheric pressure), precipitation is intermittent with a highly non-Gaussian skewed distribution \cite{Szturc2022}; (ii) the correlation between rainfall at neighboring weather stations can be remarkably low \cite{zemicheal2020conditional}; (iii) the most common correct reading is zero precipitation, which is the same reading recorded when a rain gauge is blocked (i.e., a sensor reporting zero rainfall on rainy days). Addressing these challenges is crucial for ensuring the accuracy and reliability of rainfall data in critical regions worldwide.
	
	The most prevalent methods for QC in weather data can be grouped into two categories: (1) rule-based methods, which encode expert knowledge of the variable being checked into a set of rules such that measurements that violate those rules are flagged as anomalous \citep{lewis2018rule, lewis2021quality, Kondragunta2006}; (2) probabilistic methods, which typically model the probability (or probability density) of the correctly functioning system and then use a predefined threshold to flag test observations having a low probability (or density). A drawback of these methods is that if the system is functioning correctly but in an unusual state, the probability can be low, which leads to false alarms. Conversely, if the system is faulty but in a very common state, then the probability of the observations may still be very high, and this leads to missed alarms. 
	
	Rule-based methods offer a distinct advantage in their simplicity and ease of implementation. They are computationally efficient and they mostly do not rely on historical training data, making them accessible and practical. However, their effectiveness is confined to anomalies within the realm of expert knowledge. In contrast, probabilistic models, while computationally more demanding, excel in conducting joint reasoning across multiple sensors. This capability enhances their potential to detect clusters of malfunctioning sensors (e.g., a natural catastrophe such as flood or hurricane can cause multiple sensor from a same region to report incorrect values), presenting a valuable advantage in scenarios where agreement between multiple broken sensors increases the difficulty of anomaly detection.
	
	This paper proposes two general purpose methods for sensor diagnosis via joint reasoning that aim to address the limitations of existing QC methods in settings when clusters of sensors can be simultaneously faulty.
	
	This article is structured as follows. In Sections \ref{sec:ch2_problem_setup} and section \ref{sec:ch2_background}, we formally present the problem addressed in this paper and provide a comprehensive background on consistency-based diagnosis and probabilistic fault detection. Section \ref{sec:ch2_methods} introduces the proposed methods for probabilistic consistency-based sensor diagnosis via joint reasoning. Section \ref{sec:ch2_experiments} describe the datasets used for evaluating the proposed methods, introduces the metrics used, and describes the experimental setup used to analyze each method, and presents the results. Finally, in section \ref{sec:ch2_conclusion}, we conclude the article and outline future research directions.

	\section{Problem Setup} \label{sec:ch2_problem_setup}
	
	Let $\mathcal{S} = \{\text{S}_1,\ldots,\text{S}_n\}$ be a network of $n$ weather sensors, and $\textbf{x}_t = [\text{x}_{1t},\ldots,\text{x}_{nt}]$ be a vector of measurements/observations of a variable (e.g., rainfall, air temperature, atmospheric pressure) from the sensors in $\mathcal{S}$ at time $t$. We use notation $\eta_i \subset \mathcal{S} \setminus \{\text{S}_i\}$ to denote the set of $k$ most similar sensors to S$_i$. Where similarity can be defined in terms of distance or historically correlation between two sensors.
	
\begin{figure}
		\begin{minipage}[b]{0.48\textwidth}
			\centering
			\includegraphics[width=0.98\textwidth]{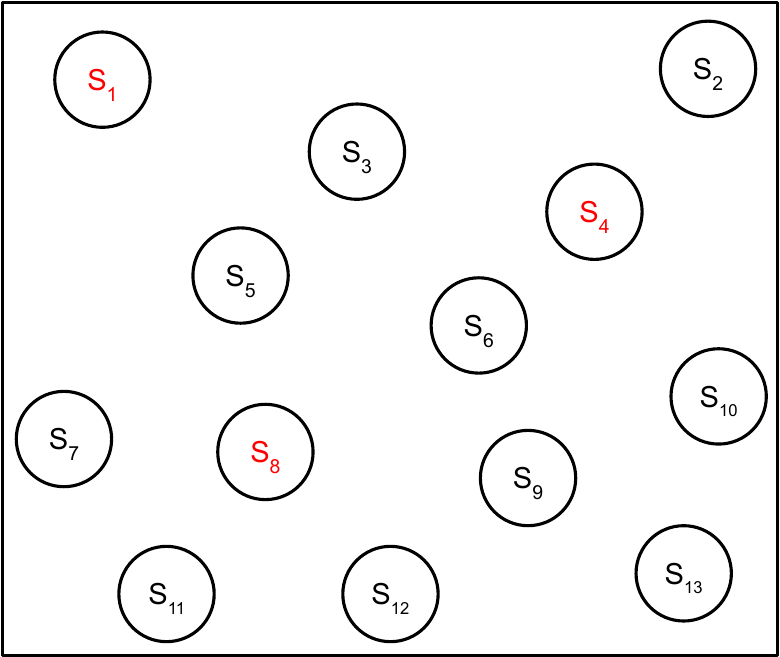}
			\captionof{figure}{Illustrative example of a network with faulty sensors that spatially randomly distributed. S$_1$, S$_4$ and S$_8$ (in red) are the faulty sensor, while the rest of the sensor are working correctly.}
			\label{fig:illutration_spatially_random_faults}
		\end{minipage}
		\hfill
		\begin{minipage}[b]{0.48\textwidth}
			\centering
			\includegraphics[width=0.98\textwidth]{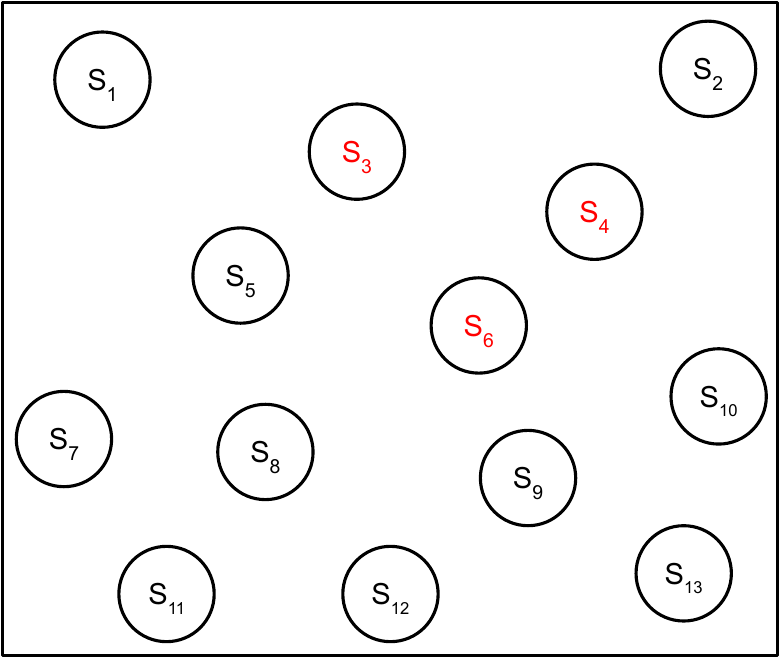}
			\captionof{figure}{Illustrative example of a network with faulty sensors that spatially clustered. S$_3$, S$_4$ and S$_6$ (in red) are the faulty sensor, while the rest of the sensor are working correctly.}
			\label{fig:illutration_spatially_clustered_faults}
		\end{minipage}
	\end{figure}

	Given historical data $\mathcal{D}_{train}=\{\textbf{X}_t\in \mathbb{R}^n\}_{t=1}^T$ of a weather variable from the sensor network $\mathcal{S}$, we are interested in diagnosing the sensors at time $t' > T$, based on observations $\mathcal{D}_{test}=\{\textbf{X}_t\in \mathbb{R}^n\}_{t=t'}^M$. We assume that all sensors are working correctly up to time $T$, and that certain sensors may be malfunctioning from time $t' > T$. We also assume that the faulty sensors (if any) can be either spatially random (i.e., independent) or spatially clustered (i.e., a group of neighboring sensors becoming faulty simultaneously) --- see Figure \ref{fig:illutration_spatially_random_faults} and \ref{fig:illutration_spatially_clustered_faults} for an illustration of spatially random and spatially clustered faulty sensors. We denote the set of broken sensors as $\Delta= \{\text{S}_i \in \mathcal{S} | \text{b}_i=1\}$, where b$_i=1$ indicates that sensor S$_i$ is functioning incorrectly, and b$_i=0$ indicates that sensor S$_i$ is working properly.
	
	We use notation $\textbf{X}_{-i}$ to represent the vector $\textbf{X}$ without index $i$ (i.e., all sensors except S$_i$), and we use notation S$_i$ interchangeably to refer to either weather station $i$ or sensor $i$ depending on the context".
	
	\section{Background} \label{sec:ch2_background}
	In this section, we introduce the theories of consistency-based diagnosis and probabilistic fault diagnosis. We then discuss the advancements and contributions of our proposed methods.

	\subsection{Consistency-Based Diagnosis}
	There are two main approaches to diagnosis. The first approach, sometimes called ``abductive'' or ``experiential'', assumes that the set of possible faults in the system is known. For every possible combination of system inputs and faults, this approach applies a model to predict the observations that will be produced by the system. The model must include both the normal behavior of the system and the behavior for each possible combination of faults. Models of faulty behavior are called ``fault models.'' For example, in medical diagnosis, each fault is a disease, and for each disease, physicians know what symptoms are likely to be observed (fever, headache, lab tests that can identify different types of bacteria, etc.). A diagnosis identifies a set of faults that explain all of the observed outputs of the system including, particularly, the symptoms. Diagnostic reasoning can be accomplished through formal logic or through probabilistic methods. Bayesian diagnosis, in particular, computes the posterior probability of each possible diagnosis given the observations.
	
	The requirement to have fault models leads to two problems. First, creating and debugging the fault models is very time-consuming. Second, it is impossible to diagnosis novel faults.
	
	The second approach, consistency-based diagnosis, refers to fault detection techniques that rely on modeling only the correct behavior of a system and detecting faults as inconsistencies between expected and observed behavior. Any inconsistency induces a diagnosis problem, which is resolved by identifying the set of system components that explain the inconsistency when they are assumed to be faulty \citep{de1987diagnosing, kleer1976local, lucas2001bayesian, reiter1987theory}. Unlike the abductive/experiential approach to diagnosis \cite{cazes2020model, flesch2007conflict, reiter1987theory}, consistency-based methods --- also referred to as diagnosis from first principles \cite{reiter1987theory} --- do not need models of each potential fault. Instead, any inconsistency between prediction and system observation is viewed as evidence of faulty behavior. This allows the system to detect novel faults that have not been anticipated by fault modeling from expert experience. 
	
	The consistency-based approach was first formalized in first-order logic by Ray Reiter \citet{reiter1987theory}. It defines a system with observations as a triple $(SD, COMPS, OBS)$, where $SD$, short for system description, is a model (expressed as a set of logical formulas) describing how the components of the system normally behave. Each logical formula that describes the behavior of a component $C$ includes a predicate $\neg AB(C)$ (short for ``$C$ is not abnormal") to say that the behavior is only predicted when $C$ is working properly \cite{mccarthy1986applications}. $COMPS$ is a finite set of components in the system. $OBS$ is a finite set of first-order formulae representing the available observations \citep{lucas2001bayesian, pulido2004possible, reiter1987theory}. An example of a system description for a precipitation sensor is the following: ``The range of rainfall collected by a normal rain gauge in $5$ minutes is between $0$ and $0.4$ inches." Which can be expressed in first-order logic as follows: \\
	\begin{align*}
		\text{RAINGAUGE}(x) \land \text{DURATION}(t)=5 \land \neg \text{AB}(x) \supset \text{RAIN}(x, t) \geq 0 \land \text{RAIN}(x, t) \leq 0.4 
	\end{align*}
	
	\noindent
	The literal $AB(x)$ is the health function, which takes value true when component $x$ is faulty and false otherwise. Likewise, the literal $\neg AB(x)$ describes a normal component when it takes value true. A system $(SD, COMPS, OBS)$ with $n$ components $COMPS=\{c_1,\ldots,c_n\}$ is considered faulty when an observation has been made that conflicts with the system description. That is, at least one of the system components must be dysfunctional. In other words, the observation contradicts the system description when all the components are assumed to be working correctly. This can be expressed as:
	\begin{align*}
		SD \land \neg AB(c_1) \land \ldots \land \neg AB(c_n) \land OBS \vdash \bot,
	\end{align*}
	where $\vdash$ is the logical entailment relation, and $\bot$ expresses a contradiction. For example, a normal rain gauge reporting negative rainfall is a contradiction.
	
	Finding the set of components that, when assumed faulty explains the contradiction constitutes a diagnosis. Which we formally define by borrowing Reiter's \cite{reiter1987theory} notation as follows:
	
	\begin{definition}[\textit{Diagnosis}] \label{def:diagnosis}
		$\Delta \subseteq COMPS$ is a diagnosis for the system $(SD, COMPS, OBS)$ \textit{if} 
		\begin{equation*}
			SD \land OBS \land \{AB(c) | c \in \Delta \} \cup \{\neg AB(c) | c \in COMPS \setminus \Delta \}
		\end{equation*}
		is consistent.
	\end{definition}
	
	From definition \ref{def:diagnosis}, the following remarks emerge:
	\begin{itemize}
		\item An inconsistency may be explained by multiple diagnoses
		\item Assuming all components to be faulty always explains any inconsistency (trivial diagnosis). 
	\end{itemize}
	With these remarks, a more practical definition of diagnosis is one that seeks to find the diagnosis $\Delta$ with the smallest cardinality. Let's call such a diagnosis a \textit{minimal diagnosis}. Since one can have multiple minimal diagnoses, algorithms similar to \citet{cazes2020model} can be used to select the most probable diagnosis. We therefore define consistency-based diagnosis as a minimal set of components that, when assumed faulty, explains the contradiction while maintaining consistency between the system description and the observation. 
	Based on proposition 3.4 from \citet{reiter1987theory}, we formally define consistency-based diagnosis as follows:
	
	\begin{definition}[\textit{Consistency-based diagnosis}] \label{def:consistency_based_diag}
		$\Delta \subseteq COMPS$ is a consistent diagnostic for the system $(SD, COMPS, OBS)$ \textit{if} $\Delta$ is a minimal set such that
		\begin{equation*}
			SD \land OBS \land \{AB(c) | c \in \Delta \} \cup \{\neg AB(c) | c \in COMPS \setminus \Delta \}
		\end{equation*}
		is consistent.
	\end{definition}
	
	\begin{figure}
		\begin{minipage}[b]{0.47\textwidth}
			\centering
			\includegraphics[width=0.6\textwidth]{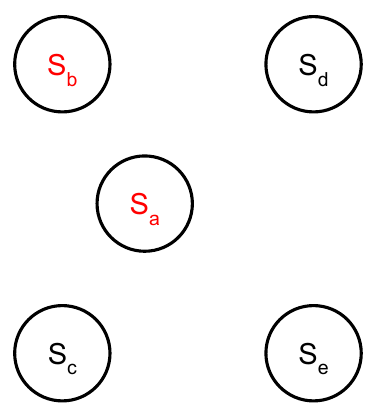}
			\captionof{figure}{Illustrative sensor network.}
			\label{fig:toy_exple_consistency_based_diag}
		\end{minipage}
		\hfill
		\begin{minipage}[b]{0.47\textwidth}
			\centering
			\begin{tabular}{c|c}
				\hline
				Sensor & neighbors \\
				\hline
				{\color{red} S$_a$} & {\color{red} S$_b$}, S$_c$ \\
				{\color{red} S$_b$} & {\color{red} S$_a$}, S$_d$ \\
				S$_c$ & {\color{red} S$_a$}, S$_e$ \\
				S$_d$ & {\color{red} S$_a$}, {\color{red} S$_b$} \\
				S$_e$ & S$_a$, S$_c$ \\ \hline
			\end{tabular}
			\captionof{table}{Neighbors of each sensor in the illustrative example.}
			\label{tab:neighbors_sat_method_illustration}
		\end{minipage}
	\end{figure}
	
	\begin{example} \label{exple:illustrative_toy_network}
		Consider Figure \ref{fig:toy_exple_consistency_based_diag}, which depicts a weather network of 5 rain gauges from 5 weather stations S$_a$, S$_b$, S$_c$, S$_d$ such that S$_c$, S$_d$ and S$_e$ are working correctly, while S$_a$ and S$_b$ are simultaneously broken. S$_a$ and S$_b$ always report zero rainfall even on rainy days. Suppose a sensor's observation can be predicted using its $2$ closest neighbors, and let $\eta_i$ be the set of neighbors of sensor S$_i$ ($i \in \{a,b,c,d,e\}$) as defined in Table \ref{tab:neighbors_sat_method_illustration}. \\ 
		
		\noindent
		Suppose the rainfall observed over the network for $5$ days is given in Table \ref{tab:consistency-based_diag_exple_obs},
		\begin{table}
			\centering
			\begin{tabular}{c c c c c c}
				\hline
				Day & Rain S$_a$ & Rain S$_b$ & Rain S$_c$ & Rain S$_d$ & Rain S$_e$ \\
				\hline
				$1$ & $0.0$ & $0.0$ & $5.1$ & $2.4$ & $3.3$ \\ 
				$2$ & $0.0$ & $0.0$ & $0.0$ & $0.0$ & $0.0$ \\ 
				$3$ & $0.0$ & $0.0$ & $3.5$ & $6.2$ & $7.0$ \\ 
				$4$ & $0.0$ & $0.0$ & $2.0$ & $4.1$ & $5.0$ \\ 
				$5$ & $0.0$ & $0.0$ & $4.0$ & $5.5$ & $1.0$ \\ \hline
			\end{tabular}
			\caption{Illustrative artificial rainfall (mm) observed over the sensor network in Figure \ref{fig:toy_exple_consistency_based_diag}}
			\label{tab:consistency-based_diag_exple_obs}
		\end{table}
		
		\noindent
		Let $SD_1=\{\forall x \in COMPS, \text{  RAINGAUGE}(x) \land \neg \text{AB}(x) \supset \text{RAIN}(x) \geq 0 \}$, and
		\begin{flalign*}
			SD_2 = \{& \forall x \in COMPS, (y, z)=\eta_x, && \\ 
			& \bigwedge_{t=1,\ldots,T} \Big( &&  \\
			& \text{RAINGAUGE}(x) \land \text{RAINGAUGE}(y) \land \text{RAINGAUGE}(z) && \\
			& \land \neg \text{AB}(x) \land \neg \text{AB}(y) \land \neg \text{AB}(z) \land \text{DAY}(t) && \\
			& \land \text{RAIN}(x, t) = 0 \land \text{RAIN}(y, t) > 0 \land \text{RAIN}(z, t) > 0 \Big) \vdash \bot && \\
			&\}
		\end{flalign*}    
		$SD_1$ describes the fact that all working rain gauges always report non-negative rainfall, while $SD_2$ states that it is a contradiction if a target rain gauge and its neighbors are all working correctly and for $T$ consecutive days, the neighbors report nonzero rainfall while the target reports zero rainfall.
		
		\noindent
		We have 
		
		\begin{itemize}
			\item $COMPS=\{\text{S}_a, \text{S}_b, \text{S}_c, \text{S}_d, \text{S}_e\}$
			\item  $SD=SD_1 \cup SD_2$
			\item $OBS = \{(\text{Rain}(\text{S}_a, t),\ldots,\text{Rain}(\text{S}_e, t)), \quad \forall t=1,\ldots,5\}$ \qquad (rainfall values from Table \ref{tab:consistency-based_diag_exple_obs})
		\end{itemize}
		
	\end{example}
	
	\noindent A minimal consistent diagnosis for the sensor network of example \ref{exple:illustrative_toy_network} is $\Delta = \{\text{S}_a, \text{S}_b\}$. 
	
	\begin{proof} 
		(i). $\text{AB}(\text{S}_a) \land \text{AB}(\text{S}_b) \land \neg \text{AB}(\text{S}_c) \land \neg \text{AB}(\text{S}_d) \land \neg \text{AB}(\text{S}_e) \nvdash \bot$ therefore, $\Delta$ is consistent. (ii). Suppose there exists $\Delta' \subseteq COMPS$ such that $\text{Card}(\Delta') < \text{Card}(\Delta)$ and $\big( \bigwedge_{\text{S}_i \in \Delta'} \text{AB}(\text{S}_i) \big) \land \big( \bigwedge_{\text{S}_j \in COMPS \setminus \Delta'} \neg \text{AB}(\text{S}_j) \big) \nvdash \bot$. $\text{Card}(\Delta') < \text{Card}(\Delta)$ implies $\text{Card}(\Delta') \leq 1$. 
		
		$\bigwedge_{\text{S}_j \in COMPS \setminus \{\text{S}_a\}} \big( \cdot \big) \land \neg \text{AB}(\text{S}_a) \vdash \bot$, $\bigwedge_{\text{S}_j \in COMPS \setminus \{\text{S}_b\}} \big( \cdot \big) \land \neg \text{AB}(\text{S}_b) \vdash \bot$ and $\bigwedge_{\text{S}_j \in COMPS \setminus \{\text{S}_a\}} \big( \cdot \big) \land \neg \text{AB}(\text{S}_b) \land \neg \text{AB}(\text{S}_b) \vdash \bot$. Thus, $\text{Card}(\Delta') \geq 2$. This is a contradiction. Therefore, $\Delta$ is minimal.
		$\bigwedge_{\text{S}_j \in COMPS \setminus \{\text{S}_a\}} \big( \cdot \big)$ represents any combination of predicates AB and/or $\neg$AB.
		
	\end{proof}

	\subsection{Probabilistic Fault Diagnosis}
	Probabilistic fault diagnosis methods leverage probabilistic models and statistical inference techniques to quantify uncertainty and make informed decisions about system conditions. These methods use fault classification heuristics to estimate the most likely fault state of system components \cite{bartha1999probabilistic}. Probabilistic graphical models such as Bayesian networks, Markov networks and logistic regression are popular examples of probabilistic fault diagnosis methods \cite{koller2009probabilistic, murphy2012machine}. Bayesian networks and Markov networks represent the relationships between variables (system components, fault types, observations) as directed and undirected graphs, respectively. Bayesian network approaches combine prior knowledge (e.g., fault prevalence) with new evidence to calculate the probability of a particular fault or combination of faults \citep{cai2017bayesian, dereszynski2011spatiotemporal, lucas2001bayesian, middleton1991probabilistic, yongli2006bayesian}. For example, consider the rain gauge network from Figure \ref{fig:toy_exple_consistency_based_diag}, and suppose that the true rainfall at weather station S$_i$ is a latent variable Z$_i$, and X$_i$ is an observation of the true rainfall Z$_i$ by S$_i$ ($i=a,\ldots,e$). Let b$_i$ be a binary random variable representing the state of sensor S$_i$. If b$_i = 0$, S$_i$ is working correctly, but if b$_i=1$, S$_i$ is faulty. An example Bayesian network representation of the system is presented in Figure \ref{fig:toy_exple_bayesian_net}. This example encodes the causal relationship between the true but latent rainfall variables $\{Z_i\}$, and the knowledge that the rain gauge reading $X_i$ (the observed rainfall) is caused by the true rainfall $Z_i$ and the sensor state $b_i$. From this Bayesian network, fault diagnosis can be performed by computing the posterior distribution over state variables $\{b_i\}$ given the observations. The posterior distribution can be derived from the joint distribution $P(\text{Z}_a,\ldots,\text{Z}_e,\text{X}_a,\ldots,\text{X}_e,\text{b}_a,\ldots,\text{b}_e)$, using the Bayes rule as follows: 
	
	\begin{figure}
		\centering
		\includegraphics[width=0.6\textwidth]{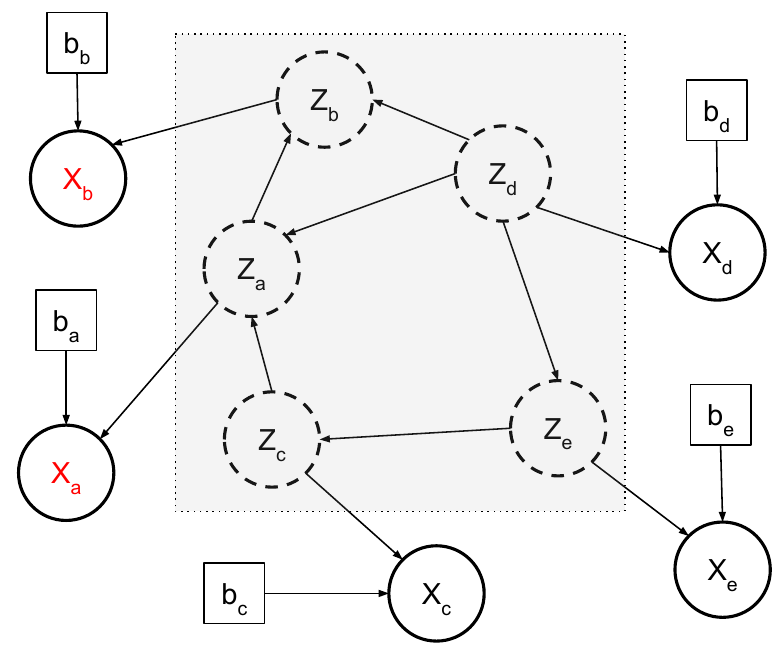}
		\captionof{figure}{Illustrative Bayesian network.}
		\label{fig:toy_exple_bayesian_net}
	\end{figure}
	
	\begin{align}
		P(\textbf{b}|\textbf{X}) = \frac{P(\textbf{b}) \int_{\textbf{Z}}P(\textbf{X},\textbf{Z}|\textbf{b})d 
			\textbf{Z}}{\int_{\textbf{Z}} P(\textbf{Z},\textbf{X})d 
			\textbf{Z}},
	\end{align}
	with $\textbf{b}=[\text{b}_a,\ldots,\text{b}_e]$, $\textbf{X}=[\text{X}_a,\ldots,\text{X}_e]$ and $\textbf{Z}=[\text{Z}_a,\ldots,\text{Z}_e]$.
	$P(\textbf{b})$ is the prior belief about the sensors' state before data is observed, and $P(\textbf{b}|\textbf{X})$ is the posterior belief. The posterior belief assigns a probability to every possible setting of $\mathbf{b}$. There are exponentially-many such settings, so some algorithms only seek the highest probability diagnosis
	\begin{equation}
		\mathbf{b}^* = \argmax_{\mathbf{b}} P(\mathbf{b} \mid \mathbf{X}).
	\end{equation}
	
	Our description of probabilistic diagnosis has assumed that the system is modeled in the causal direction. It is possible to approach diagnosis in the anti-causal direction by formulating it as a classification problem. The observations $\{X_i\}$ are the inputs, and the state variables $\{b_i\}$ are the outputs. To avoid the combinatorial explosion of possible outputs, it is customary to predict the presence of each fault $b_i$ separately. Hence, if there are $n$ components, we must fit $n$ separate classifiers. Probabilistic classifiers, such as logistic regression, are typically employed  \citep{murphy2012machine, olsson2014explaining, pandya2014fault, zemicheal2020conditional}. 
	
	A major advantage of probabilistic methods for fault diagnosis is the uncertainty quantification and the confidence in the prediction provided by probability values. But the disadvantage of existing methods for probabilistic diagnosis is that they require fault models. In particular, note that $P(X_i \mid b_i=1)$ is the fault model for component $i$. The central contribution of this chapter is to introduce probabilistic methods for consistency-based diagnosis.

	\section{Proposed Methods} \label{sec:ch2_methods}
	
	In this section, we describe the two proposed probabilistic consistency-based sensor diagnosis (PCSD) methods. In section \ref{subsec:pcsd_sat}, we present a Boolean satisfiability constrained optimization-based method (PCSD-SAT), and we describe an iterative belief update-based approach in section \ref{subsec:pcsd_ji}. We propose two versions of the latter approach, one based on a kernel regression model, that we name PCSD-JIKR, and the other one, PCSD-JIGM, based on a joint Gaussian model. We give the intuition behind each method and we describe the corresponding algorithms.

	\subsection{Probabilistic Consistency-based Sensor Diagnosis using SAT-solvers (PCSD-SAT)} \label{subsec:pcsd_sat}
	
	Consider the sensor network $\mathcal{S}=\{\text{S}_1,\ldots,\text{S}_n\}$, and let $\textbf{X}_t$ be a vector of observations throughout the network at time $t$. We will model the joint distribution of $\textbf{X}_t$ as a multivariate Gaussian distribution with mean $\bm{\mu}$ and covariance $\bm{\Sigma}$:
	\begin{align*}
		\textbf{X}_t \sim \mathcal{N}(\bm{\mu}, \bm{\Sigma}).
	\end{align*}
	
	Let $\eta_i \subseteq \mathcal{S} \setminus \{\text{S}_i\}$ be the set of $k$ neighbors of S$_i$, and $\textbf{x}_{\eta_i}^t$ be the vector of observations at $\eta_i$ at time $t$. The consistency of an observation x$_i^t$ from sensor S$_i$ at time $t$ can be evaluated using the conditional distribution $P(\text{x}_i^t|\textbf{x}_{\eta_i}^t)$, which is a Gaussian distribution with parameters $\hat{\mu}_i^t$ and $\hat{\sigma}^2_{i}$, defined in equation \ref{eq:mu_cond_gaussian} and \ref{eq:sigma_cond_gaussian} \cite{bishop2006pattern, murphy2012machine}
	\begin{align} 
		\hat{\mu}_i^t &= \mu_i + \bm{\Sigma}_{i,\eta_i}\bm{\Sigma}_{\eta_i,\eta_i}^{-1}(\textbf{x}_{\eta_i}^t-\bm{\mu}_{\eta_i}) \label{eq:mu_cond_gaussian} \\
		\hat{\sigma}^2_{i} &= \Sigma_{i,i} - \bm{\Sigma}_{i,\eta_i}\bm{\Sigma}_{\eta_i,\eta_i}^{-1}\bm{\Sigma}_{\eta_i,i}, \label{eq:sigma_cond_gaussian}
	\end{align}
	where $\bm{\mu}_{\eta_i}$ and $\bm{\Sigma}_{\eta_i,\eta_i}$, which are subsets of $\bm{\mu}$ and $\bm{\Sigma}$, represent the mean vector and the covariance matrix of S$_i$'s neighbors, $\eta_i$.
	
	Let $F(\text{x}_i^t \mid \textbf{x}_{\eta_i}^t)$ denote the conditional cumulative distribution function, and let $q_i^t=\min\{F(\text{x}_i^t \mid \textbf{x}_{\eta_i}^t), 1-F(\text{x}_i^t \mid \textbf{x}_{\eta_i}^t\}$ be the one-sided tail probability of x$_i^t$. Define $p_i^t=2q_i^t$ to be the two-sided $p$-value.

	\begin{definition}[$\alpha$-\textit{consistency} and $\alpha$-\textit{inconsistency}] \label{def:alpha_consistency}
		Let S$_i$ be a weather station with observation x$_i^t$ at time $t$, $\eta_i$ the set of neighbors of S$_i$, \textbf{x}$_{\eta_i}^t$ the observations from $\eta_i$ at time $t$, and $p_i^t$ be the two-sided $p$-value of x$_i^t$ conditioned on \textbf{x}$_{\eta_i}^t$. A sensor's observation x$_i$ from S$_i$ is $\alpha$-\textit{consistent} if $p_i^t \geq \alpha$ (for $\alpha \in (0, 1)$). Otherwise, it is called $\alpha$-\textit{inconsistent}. 
	\end{definition}

	Because of neighbor conditioning, x$_i$ can be $\alpha$-inconsistent without being a faulty observation. This may happen when one or more of S$_i$'s neighbors are faulty. Therefore, an $\alpha$-inconsistent reading should be interpreted as an indication that one or more of the sensors involved in $P(\text{x}_{it} \mid\eta_i)$ is faulty (i.e., S$_i$ and/or any of its neighbors can be the culprit). Finding the set of sensors that, when removed, restore $\alpha$-consistency is the goal of consistency-based diagnosis. In order to find those sensors, we propose to evaluate the $\alpha$-consistency of all possible combinations of sensors in $\{\text{S}_i\} \cup \eta_i$. Those evaluations can be summarized into a truth table, and subsequently expressed as boolean formula $f(\text{S}_i)$ that evaluates to true for any boolean assignment to the variables that leads to $\alpha$-consistency. Let b$_i$ be the state of sensor S$_i$, such that b$_i$ takes value $1$ when S$_i$ is broken, and $0$ otherwise. Any boolean assignment that sets $b_i=1$ is alpha-consistent.
	
	Once all assignments have been evaluated for each sensor S$_i$ ($\forall i=1,\ldots,n$), we create a boolean formula $f(\text{S}_i)$ from the $\alpha$-consistent assignments. We can do joint reasoning to find the set of broken sensors over the entire network by finding the minimum number of sensors that, when assigned to the faulty state (b$_i=1$) satisfy all the boolean constraints. This is equivalent to solving following optimization problem:
	
	\begin{align} \label{eq:sat_optimization}
		\begin{split}
			\min & \sum_{i=1}^{n} b_i \\
			\text{such that } & f(\text{S}_i) = True, \quad \text{for all } i = 1,\ldots,n.
		\end{split}
	\end{align}
	
	The constraints $f(\text{S}_i)$ can be further transformed into binary integer constraints, making the constraint satisfiability optimization problem a linear integer programming (ILP) optimization problem, which can be solved by existing ILP solvers \cite{cplex2009v12, gurobi}. Example \ref{exple:toy_sat_method} illustrates a step-by-step application of the method.
	
	\begin{example} \label{exple:toy_sat_method}
		Consider the illustrative rain gauge network of example \ref{exple:illustrative_toy_network}, and let x$_a$, x$_b$, x$_c$, x$_d$ and x$_e$ be the observed values from sensors S$_a$, S$_b$, S$_c$, S$_d$ and S$_e$ respectively. Let b$_a$, b$_b$, b$_c$, b$_d$ and b$_e$ be their status, where b$_i = 0$ ($i \in \{a, b, c, d, e\}$) if S$_i$ is working correctly, and b$_i = 1$ otherwise. $\eta_i$ is the set of neighbors of S$_i$ as presented in Table \ref{tab:neighbors_sat_method_illustration}. 
		\\
		\noindent
		Sensor S$_a$ can be diagnosed by reasoning over x$_a$, x$_b$, and x$_c$ through the evaluation of all combinations of boolean assignments to (b$_a$, b$_b$, b$_c$). Let $p_{a \mid b,c}$ be the p-value of x$_a$ given x$_b$ and x$_c$, and let $\alpha$ be a pre-defined threshold for consistency check. The $\alpha$-consistency of each combination (b$_a$, b$_b$, b$_c) \in \{0, 1\}^3$ can be evaluated using the p-value as follows:
		\begin{itemize}
			\item (b$_a$, b$_b$, b$_c$) = (0, 0, 0): $p_{\text{x}_a \mid \text{x}_b, \text{x}_c} < \alpha$ $\implies$ $\alpha$-inconsistent.
			\item (b$_a$, b$_b$, b$_c$) = (0, 0, 1): $p_{\text{x}_a \mid \text{x}_b} \geq \alpha$  $\implies$ $\alpha$-consistent. 
			\item (b$_a$, b$_b$, b$_c$) = (0, 1, 0): $p_{\text{x}_a \mid \text{x}_c} < \alpha$ $\implies$ $\alpha$-inconsistent.
			\item (b$_a$, b$_b$, b$_c$) = (0, 1, 1): $p_{\text{x}_a} = 1$ $\implies$ $\alpha$-consistent by definition.
			\item All (b$_a$, b$_b$, b$_c$) = (1, *, *) are $\alpha$-consistent by definition ($* \in \{0,1\}$).
		\end{itemize}
		
		\noindent
		This evaluation of Boolean assignments to (b$_a$, b$_b$, b$_c$) can be represented as a truth table as shown in Table \ref{tab:truth_table_sa}. Similarly, the truth tables resulting from evaluating the $\alpha$-consistency of S$_b$, S$_c$, S$_d$ and S$_e$ are presented in Table \ref{tab:truth_table_sa}, \ref{tab:truth_table_sb}, \ref{tab:truth_table_sc}, \ref{tab:truth_table_sd} and \ref{tab:truth_table_se} respectively. \\
		
		\noindent
		The truth table of S$_a$ (Table \ref{tab:truth_table_sa}) can be further simplified as follows:
		\begin{flalign*}
			f(S_a) &= (\neg b_a \land \neg b_b \land b_c) \lor (\neg b_a \land b_b \land b_c) \lor (b_a \land \neg b_b \land \neg b_c) \lor (b_a \land \neg b_b \land b_c)  \\
			&\quad \lor (b_a \land b_b \land \neg b_c) \lor (b_a \land b_b \land b_c) &&
		\end{flalign*}
		
		\noindent
		This can be rewritten as
		\begin{flalign}
			f(S_a) &= \bar{b}_a \bar{b}_b b_c + \bar{b}_a b_b b_c + b_a \bar{b}_b \bar{b}_c + b_a \bar{b}_b b_c + b_a b_b \bar{b}_c + b_a b_b b_c && \nonumber \\
			&= b_a + b_c.  \qquad (\text{see appendix } \ref{appx:boolean_formula_reduction}) &&  \label{eq:f(s_a)}
		\end{flalign}
		
		\noindent
		Similarly, we have
		\begin{flalign}
			&f(S_b) = b_b + b_d && \label{eq:f(s_b)}  \\
			&f(S_c) = b_a + b_c \qquad (\text{same as } f(S_a)) && \label{eq:f(s_c)}  \\
			&f(S_d) = b_a b_b + b_d  && \label{eq:f(s_d)} \\
			&f(S_e) = b_a + b_e. && \label{eq:f(s_e)}
		\end{flalign}
		
		\noindent
		The problem of diagnosing S$_a$, S$_b$, S$_c$, S$_d$ and S$_e$ can be defined as identifying the minimum number of sensors that, when assigned a broken status, satisfy all the Boolean formulas $f(S_i)$, $\forall i \in \{a, b, c, d, e\}$. This is equivalent to the following constrained optimization problem:
		
		\begin{align} \label{eq:sat_optimization_example}
			\begin{split}
				\min & \quad \sum_{i \in \{a, b, c, d, e\}} b_i \\
				\text{s.t.}: & \quad b_a + b_c \geq 1    \\
				& \quad b_b + b_d \geq 1    \\
				& \quad b_a b_b + b_d \geq 1    \\
				&\quad b_a + b_e \geq 1    \\
				& \quad b_i \in \{0, 1\}, \quad \text{for } i = a, b, c, d, e.
			\end{split}
		\end{align}
		
		\noindent
		The minimizers of \ref{eq:sat_optimization_example} are (b$_a$, b$_b$, b$_c$, b$_d$, b$_e$) = (1, 1, 0, 0, 0) and (b$_a$, b$_b$, b$_c$, b$_d$, b$_e$) = (1, 0, 0, 1, 0). For blocked rain gauge detection, the most likely diagnostic can be selected by reasoning over the precipitation amount recorded by the faulty sensors. This leads to (b$_a$, b$_b$, b$_c$, b$_d$, b$_e$) = (1, 1, 0, 0, 0) as final diagnostic.
		
		As we have written it, the optimization problem \ref{eq:sat_optimization_example} involves products of variables. By introducing intermediate variables $\{y_j\}$, it can be expressed as a binary integer linear programming problem as shown in \ref{eq:sat_optim_example_ilp}.
		
		\begin{align} \label{eq:sat_optim_example_ilp}
			\begin{split}
				\min  & \quad \sum_{i \in \{a, b, c, d, e\}} b_i \\
				\text{s.t}: & \quad b_a + b_c \geq 1    \\
				& \quad b_b + b_d \geq 1    \\
				& \quad b_a + b_e \geq 1    \\
				& \quad b_a - y_1 \geq 0    \\
				& \quad b_b - y_1 \geq 0    \\
				& \quad b_d + y_1 \geq 1    \\
				& \quad b_i \in \{0, 1\}, \quad \text{for } i = a, b, c, d, e  \\
				& \quad y_1 \in \{0, 1\}.
			\end{split}
		\end{align}
		
		Algorithm~\ref{alg:pcsd_sat} presents the pseudo-code for PCSD-SAT. 
	\end{example}

	\begin{algorithm}[H]
		\caption{\textsc{PCSD-SAT}}
		\label{alg:pcsd_sat}
		\begin{algorithmic}[1]
			\STATE \textbf{Data Structures and Parameters}
			\STATE \quad \textbf{b}: b$_i$ is the status of sensor S$_i$
			\STATE \quad $\bm{\mu}$: $\mu_i$ is the mean value of x$_i$
			\STATE \quad $\bm{\Sigma}$: covariance matrix
			\STATE \quad $\tau$: anomaly detection threshold
			\STATE \quad $\alpha$: confidence level for $\alpha$-consistency
			\STATE \quad $k$: number of neighbors
			\STATE \textbf{Procedure}
			\STATE \quad Fit $\bm{\mu}$ and $\bm{\Sigma}$ to the training dataset
			\STATE \quad $\mathcal{Q} \gets \varnothing$ \COMMENT{set of all constraints}
			\FOR{$i = 1,\ldots,n$}
			\STATE $\eta_i \gets \text{GetNeighbors}(i, k)$ \COMMENT{$k$ nearest neighbors of S$_i$}
			\STATE $\mathcal{C} \gets \varnothing$ \COMMENT{set of consistent assignments of b$_i$s}
			\FOR{(b$_i$, b$_{\eta_{i,1}}$, \ldots, b$_{\eta_{i,k}}$) $\in \{0, 1\}^{k+1}$}
			\STATE $w \gets \{\text{b}_j: \text{b}_j = 0, \forall \text{b}_j \in (\text{b}_i, \text{b}_{\eta_{i,1}}, \ldots, \text{b}_{\eta_{i,k}})\}$ 
			\IF{$w = \varnothing$ or card$(w)=1$}
			\STATE $\mathcal{C} \gets \mathcal{C} \cup (\text{b}_i, \text{b}_{\eta_{i,1}}, \ldots, \text{b}_{\eta_{i,k}})$
			\STATE $p \gets 0$
			\ELSE
			\STATE $\bm{\tilde{\Sigma}} \gets \bm{\Sigma}_w$
			\IF{b$_i \in w$}
			\STATE $\hat{\mu}_i \gets \mu_i + \bm{\tilde{\Sigma}}_{i,-i}\bm{\tilde{\Sigma}}_{-i,-i}^{-1}(\textbf{x}_{-i}-\bm{\mu}_{-i})$
			\STATE $\hat{\sigma}^2_i \gets \bm{\tilde{\Sigma}}_{i,i} - \bm{\tilde{\Sigma}}_{i,-i}\bm{\tilde{\Sigma}}_{-i,-i}^{-1}\bm{\tilde{\Sigma}}_{-i,i}$
			\STATE $p \gets 2\min\{F(\text{x}_i; \hat{\mu}_i, \hat{\sigma}^2_i), 1 - F(\text{x}_i; \hat{\mu}_i, \hat{\sigma}^2_i) \}$ 
			\ELSE
			\STATE $j \gets Index(w[0])$ \COMMENT{index of any b$_j \in w$}
			\STATE $\hat{\mu}_j \gets \mu_j + \bm{\tilde{\Sigma}}_{j,-j}\bm{\tilde{\Sigma}}_{-j,-j}^{-1}(\textbf{x}_{-j}-\bm{\mu}_{-j})$
			\STATE $\hat{\sigma}^2_j \gets \bm{\tilde{\Sigma}}_{j,j} - \bm{\tilde{\Sigma}}_{j,-j}\bm{\tilde{\Sigma}}_{-j,-j}^{-1}\bm{\tilde{\Sigma}}_{-j,j}$
			\STATE $p \gets 2\min\{F(\text{x}_j; \hat{\mu}_j, \hat{\sigma}^2_j), 1 - F(\text{x}_j; \hat{\mu}_j, \hat{\sigma}^2_j) \}$  
			\ENDIF
			\IF{$p > \alpha$}  
			\STATE $\mathcal{C} \gets \mathcal{C} \cup (\text{b}_i, \text{b}_{\eta_{i,1}}, \ldots, \text{b}_{\eta_{i,k}})$  \COMMENT{$(\text{b}_i, \text{b}_{\eta_{i,1}}, \ldots, \text{b}_{\eta_{i,k}})$ is consistent}
			\ENDIF
			\ENDIF
			\algstore{amoc_algo}
		\end{algorithmic}
	\end{algorithm}
	
	\begin{algorithm}[H]                     
		\begin{algorithmic} [1]              
			\algrestore{amoc_algo}
			
			\ENDFOR
			\STATE $\mathcal{Q} \gets \mathcal{Q} \cup \mathcal{C}$
			\ENDFOR
			\STATE $\textbf{b} \gets \text{Solve}(\textbf{b}, \mathcal{Q})$
			\STATE $\Delta \gets \{\text{S}_i: \text{b}_i = 1, \quad \forall \text{b}_i \in \textbf{b} \}$
			\RETURN{} $\Delta$
		\end{algorithmic}
	\end{algorithm}
	
	\begin{algorithm}
		\caption{\textsc{Solve}}
		\label{alg:solve}
		\begin{algorithmic}[1]
			\STATE \textbf{Data Structures and Parameters}
			\STATE \quad \textbf{b}: vector of boolean variables
			\STATE \quad $\mathcal{Q}$: set of sets of boolean strings that evaluate to 1 (in a truth table)
			\STATE \quad QMC: Quine McCluskey procedure for boolean expression simplification
			\STATE \quad Bool2Int: boolean formula to integer constraints (see eq. \ref{eq:sat_optimization_example} to \ref{eq:sat_optim_example_ilp})
			\STATE \quad ILP: integer linear programming solver
			\STATE \textbf{Procedure}
			\STATE \quad $\mathcal{I} \gets \varnothing$  \COMMENT{set of all integer constraints}
			\FOR{$\mathcal{C} \in \mathcal{Q}$}
			\STATE $\text{bf} \gets \text{QMC}(\mathcal{C})$ 
			\STATE $\text{i} \gets \text{Bool2Int}(\text{bf})$ 
			\STATE $\mathcal{I} \gets \mathcal{I} \cup \text{i}$
			\ENDFOR
			\STATE $\text{obj} \gets \min \sum \text{b}_i$ \COMMENT{objective function}
			\STATE $\textbf{b} \gets \text{ILP}(\text{obj}, \mathcal{I})$   \COMMENT{ILP returns an assignment to $\textbf{b}$ that minimizes obj}
			\RETURN{} $\textbf{b}$
		\end{algorithmic}
	\end{algorithm}
	
	In the next section, we introduce PCSD-JIKR and PCSD-JIgm, two iterative approaches to probabilistic consistency based diagnosis via joint reasoning.

	\begin{table}
		\begin{tabular}{ccc}
			\begin{minipage}[t]{.31\linewidth}
				\centering
				\captionof{table}{Truth table of S$_a$}
				\label{tab:truth_table_sa}    
				\begin{tabular}{|c|c|c|c|}
					\hline
					$\bm{\text{b}_a}$ & $\text{b}_b$ & $\text{b}_c$ & consist. \\ \hline
					0 & 0 & 0 & 0 \\ \hline
					0 & 0 & 1 & 1 \\ \hline
					0 & 1 & 0 & 0 \\ \hline
					0 & 1 & 1 & 1 \\ \hline
					1 & $\ast$ & $\ast$ & 1 \\ \hline
				\end{tabular}
			\end{minipage} &
			\begin{minipage}[t]{.31\linewidth}
				\centering
				\captionof{table}{Truth table of S$_b$}
				\label{tab:truth_table_sb}    
				\begin{tabular}{|c|c|c|c|}
					\hline
					$\bm{\text{b}_b}$ & $\text{b}_a$ & $\text{b}_d$ & consist. \\ \hline
					0 & 0 & 0 & 0 \\ \hline
					0 & 0 & 1 & 1 \\ \hline
					0 & 1 & 0 & 0 \\ \hline
					0 & 1 & 1 & 1 \\ \hline
					1 &$\ast$ & $\ast$ & 1 \\ \hline
				\end{tabular}
			\end{minipage} &
			\begin{minipage}[t]{.31\linewidth}
				\centering
				\captionof{table}{Truth table of S$_c$}
				\label{tab:truth_table_sc}    
				\begin{tabular}{|c|c|c|c|}
					\hline
					$\bm{\text{b}_c}$ & $\text{b}_a$ & $\text{b}_e$ & consist. \\ \hline
					0 & 0 & 0 & 0 \\ \hline
					0 & 0 & 1 & 0 \\ \hline
					0 & 1 & 0 & 1 \\ \hline
					0 & 1 & 1 & 1 \\ \hline
					1 & $\ast$ & $\ast$ & 1 \\ \hline
				\end{tabular}
			\end{minipage} \\
			\begin{minipage}[t]{.32\linewidth}
				\centering
				\captionof{table}{Truth table of S$_d$}
				\label{tab:truth_table_sd}    
				\begin{tabular}{|c|c|c|c|}
					\hline
					$\bm{\text{b}_d}$ & $\text{b}_a$ & $\text{b}_b$ & consist. \\ \hline
					0 & 0 & 0 & 0 \\ \hline
					0 & 0 & 1 & 0 \\ \hline
					0 & 1 & 0 & 0 \\ \hline
					0 & 1 & 1 & 1 \\ \hline
					1 & $\ast$ & $\ast$ & 1 \\ \hline
				\end{tabular}
			\end{minipage} &
			\begin{minipage}[t]{.32\linewidth}
				\centering
				\captionof{table}{Truth table of S$_e$}
				\label{tab:truth_table_se}    
				\begin{tabular}{|c|c|c|c|}
					\hline
					$\bm{\text{b}_e}$ & $\text{b}_c$ & $\text{b}_d$ & consist. \\ \hline
					0 & 0 & 0 & 1 \\ \hline
					0 & 0 & 1 & 1 \\ \hline
					0 & 1 & 0 & 1 \\ \hline
					0 & 1 & 1 & 1 \\ \hline
					1 & $\ast$ & $\ast$ & 1 \\ \hline
				\end{tabular}
			\end{minipage} & \\
		\end{tabular}
		\caption{Truth tables. The first column of each subtable is the Boolean state of the target sensor. The symbol $\ast$ means any value in \{0,1\}, and the ``consist.''~column is the consistency of the target with those of its neighbors having a Boolean state of 0.}
	\end{table}

	\subsection{Probabilistic Consistency-based Sensor Diagnosis via Iterative Joint Inference (PCSD-JI)} \label{subsec:pcsd_ji}
	
	In the previous section, we employ a hard p-value threshold to determine whether a station is faulty. In this section, we take a softer, incremental approach. We assign a variable $\theta_i$ to each station S$_i$, that quantifies our belief that S$_i$ is working correctly. We use these $\theta_i$ values to modify the consistency p-value computation so that if station S$_j$ is a neighbor of station S$_i$ with $\theta_j < 1$, then the observation x$_j$ is down-weighted when computing the p-value of S$_i$. 
	
	We initialize each $\theta_i := 1$, and then iteratively (and in parallel) update the $\theta_i$ values to shrink those whose observation x$_i$ has a small consistency p-value. The hope is that this iterative computation will converge to a good diagnostic solution where the broken stations have $\theta < \tau$, for a learned threshold $\tau$, while the working stations have $\theta > \tau$.
	
	Consider a sensor network $\mathcal{S} = \{\text{S}_1,\ldots,\text{S}_n\}$, and let $\theta_i \in [0, 1]$, $i=1,\ldots,n$ be our belief about the state of sensor S$_i$, such that $\theta_i=1$ if we are certain that S$_i$ is working correctly, and $\theta_i=0$ otherwise. Let X$_i$ be a random variable representing the variable measured by S$_i$ (e.g., precipitation, air temperature), and $\theta'_i = 2 \min \{F(\text{x}_i; \mu_i, \sigma^2_i), 1-F(\text{x}_i; \mu_i, \sigma^2_i)\}$ be the probability of sensor S$_i$ reporting a value more extreme than the observed value x$_i$ (i.e., the 2-sided $p$-value). Let $F$ be the CDF of the predictive distribution $P(\text{X}_i|\textbf{x}_{-i})$, a Gaussian model parameterized by mean $\mu_i$ and variance $\sigma^2_i$. Given that the rest of the sensors are conditioning variables in the predictive distribution, and knowing that some of those conditioning variables could be observations from faulty sensors, we weigh the contribution of each variable (neighbor) relative to the level of belief about its state. That is, we want the contribution of each conditioning variable X$_j$ ($\forall j \neq i$) to be proportional to our belief about its working state. Once the predictive distribution is estimated for each sensor, we re-compute the $p$-value, update the $\theta$ values for only the sensors whose observations are $\alpha$-inconsistent and repeat the process until convergence of $\theta$. A predefined threshold $\tau$ (empirically selected) can be applied to the learned $\theta_i$ (for all $i=1\ldots,n$) to flag faulty sensors.
	
	We propose two methods for estimating the parameters of the predictive distribution $P(\text{X}_i \mid \textbf{x}_{-i})$ that integrates the belief about the state of the sensors $\theta_i$ for $i=1,\ldots,n$ into the estimate. Section \ref{subsubsec:ch2_kernel_reg} describes the first method, PCSD-JIKR which uses a kernel regression model and the second method, PCSD-JIGM which is based on a joint Gaussian model is presented in section \ref{subsubsec:ch2_cond_gaussian}.
	
	\subsubsection{Kernel Regression Model (PCSD-JIKR)} \label{subsubsec:ch2_kernel_reg}
	
	In this method, we compute the conditional distribution of $\text{X}_i$ using kernel regression modified to incorporate the belief parameters $\theta_i$. We model the conditional distribution as a Gaussian with mean $\mu_i$ and variance $\sigma_i$, computed as follows:
	\begin{align}
		\mu_i &= \left.\sum_{j \neq i} \theta_j \kappa(i,j) \text{x}_j \middle/ \sum_{j \neq i} \theta_j \kappa(i,j) \right. \\
		\sigma^2_i &= \left.\sum_{j \neq i} \theta_j \kappa(i,j) (\text{x}_j - \mu_i)^2 \middle/ \sum_{j \neq i} \theta_j \kappa(i,j)\right.,
	\end{align}
	where $\theta_j$ is the belief about the state of sensor S$_j$, $\kappa(i,j) = \sigma_f^2\exp\left(-\frac{d(i,j)^2}{2\ell^2}\right)$ is a radial basis function (RBF) kernel with scale parameter $\ell$, and $d(i,j)$ is the distance between locations of sensors $S_i$ and $S_j$. We set $\sigma_f = 1$ throughout this paper and tune $\ell$ on training data.
	
	The sensors with small $\theta$ values make smaller contributions to the estimated mean $\mu$ and variance $\sigma^2$. This attenuates the influence of faulty neighbors in the prediction of the expected value recorded by a given target sensor. 
	
	As described above, the $\theta$ values are initialized to 1 and then iteratively shrunk towards the computed $p$-values. The pseudo-code of the iterative procedure is presented in algorithm \ref{alg:pcsd_kernel_reg}. Note that $\theta_i$ is only updated when the p-value $p_i$ is smaller than the pre-defined significance level $\alpha$. This prevents all of the $\theta$ values from collapsing to zero. Note also that the sensors with biggest standard errors $|\frac{\text{x}_i-\mu_i}{\sigma^2_i}|$ are more aggressively updated.
	
	\begin{algorithm}
		\caption{\textsc{PCSD-JIKR}}
		\label{alg:pcsd_kernel_reg}
		\begin{algorithmic}[1]
			\STATE \textbf{Data Structures and Parameters}
			\STATE \quad $\bm{\theta}$: $\theta_i$ is the status of sensor S$_i$
			\STATE \quad $\kappa$: $\kappa(i,j)$ is the kernel between S$_i$ and S$_j$
			\STATE \quad $\tau$: fault detection threshold
			\STATE \quad $\alpha$: confidence level for $\alpha$-consistency
			\STATE \quad $\lambda$: learning rate
			\STATE \textbf{Procedure}
			\STATE \quad $\theta_i \gets 1$ for all $i=1,\ldots,n$  \COMMENT{initialization}
			\STATE \quad $\theta'_i \gets 1$ for all $i=1,\ldots,n$ \COMMENT{initialization}
			\REPEAT
			\FOR{$i = 1,\ldots,n$}
			\STATE $\hat{\mu}_i \leftarrow \left.\sum_{j \neq i} \theta_j \kappa(i,j) \text{x}_j \middle/ \sum_{j \neq i} \theta_j \kappa(i,j)\right.$
			\STATE $\hat{\sigma}^2_i \leftarrow \left.\sum_{j \neq i} \theta_j  \kappa(i,j) (\text{x}_j - \hat{\mu}_i)^2 \middle/ \sum_{j \neq i} \theta_j  \kappa(i,j) \right.$
			\STATE $p_i \gets 2\min\{F(\text{x}_i; \hat{\mu}_i, \hat{\sigma}^2_i), 1 - F(\text{x}_i; \hat{\mu}_i, \hat{\sigma}^2_i) \}$ 
			\STATE $\theta'_i \gets 1 - \max\{\alpha - p_i, 0\}$
			\ENDFOR
			\FOR{$i = 1,\ldots,n$}
			\STATE $\theta_i \gets \theta_i + \lambda (\theta'_i - \theta_i)$ \COMMENT{Move $\theta_i$ toward $\theta'_i$}
			\ENDFOR
			\UNTIL{$\bm{\theta}$ converges}
			\STATE $\Delta \gets \{\text{S}_i: \theta_i < \tau, \quad \forall i = 1,\ldots,n \}$
			\RETURN{} $\Delta$
		\end{algorithmic}
	\end{algorithm}

	\subsubsection{Joint Gaussian Model (PCSD-JIGM)} \label{subsubsec:ch2_cond_gaussian}
	
	PCSD-JIGM is identical to PCSD-JIKR with the kernel regression replaced by a joint multivariate Gaussian model. To describe the method, we must explain how to incorporate the belief variables $\{\theta_i\}$ into the computation of the conditional Gaussian $P(\text{X}_i \mid \textbf{x}_{-i})$ for sensor S$_i$.
	
	We first estimate the joint Gaussian on historical (fault-free) data to obtain $\bm{\mu}$ and $\bm{\Sigma}$. Let $(\theta_1, \ldots, \theta_n)$ be the values of the belief variables at a given point in the PCSD-JI iterations. In the multivariate Gaussian, the importance of each conditioning variable in the estimation of the conditional mean and variance is controlled by the structure of the covariance matrix. To reduce the importance of variable $X_j$, we can increase its variance as follows. Let $\textbf{D}$ be an $n\times n$ matrix whose diagonal element $D_{i,i}$ is $1/\theta_i$ for all $i$ and whose off-diagonal elements are all 1. The conditional mean and variance can then be computed as
	\begin{align}
		\bm{\tilde{\Sigma}} &\gets \textbf{D} \odot \bm{\Sigma}\\
		\hat{\mu}_i &\gets \mu_i + \bm{\tilde{\Sigma}}_{i,-i}\bm{\tilde{\Sigma}}_{-i,-i}^{-1}(\textbf{x}_{-i}-\bm{\mu}_{-i}) \\
		\hat{\sigma}^2_i &\gets \bm{\tilde{\Sigma}}_{i,i} - \bm{\tilde{\Sigma}}_{i,-i}\bm{\tilde{\Sigma}}_{-i,-i}^{-1}\bm{\tilde{\Sigma}}_{-i,i},
	\end{align}
	where $\odot$ denotes element-wise multiplication. 
	
	The rationale for this approach is the following. We want a mechanism that can decrease the impact of variable $X_j$ on the estimated conditional distribution without altering the covariance structure of the joint Gaussian. We can do this by adding zero-mean Gaussian noise $\delta \sim \mathcal{N}(0, \epsilon)$ to $X_j$ to obtain a modified variable $\tilde{X}_j$. This has the following effects:
	\begin{align}
		\mathbb{E}[\tilde{\text{X}}_j] &= \mathbb{E}[\text{X}_j] + \mathbb{E}[\delta_j] = \mathbb{E}[\text{X}_j] \label{eq:mean_added_gaussian_noise} \\ 
		\text{Cov}(\tilde{\text{X}}_j, \text{X}_k) &= \text{Cov}(\text{X}_j + \delta, \text{X}_k) \label{eq:cov_added_gaussian_noise} \\ 
		&= \text{Cov}(\text{X}_j, \text{X}_k) + \text{Cov}(\delta_j, \text{X}_k) \nonumber \\ 
		&= \text{Cov}(\text{X}_j, \text{X}_k) \nonumber  \\
		\text{Var}(\tilde{\text{X}}_j) &= \text{Var}(\text{X}_j) + \text{Var}(\delta_j) = \sigma_j^2 + \epsilon \label{eq:var_added_gaussian_noise}
	\end{align}
	
	Equation \ref{eq:mean_added_gaussian_noise} shows that the mean is unaffected (because the noise has zero mean), and Equation \ref{eq:cov_added_gaussian_noise} shows that the covariance with any other sensor S$_k$ is unchanged (because the added noise is uncorrelated with $X_k$, so $\text{Cov}(\delta_j, X_k) = 0$). However, Equation \ref{eq:var_added_gaussian_noise} shows that the variance of $X_j$ increases as desired. In PCSD-JIGM, rather than adding $\epsilon$, we divide by $\theta_j$ so that the variance goes to $+\infty$ as $\theta_j \rightarrow 0$, at which point the effect of $X_j$ would be completely removed. 
	
	Algorithm~\ref{alg:pcsd_gaussian} gives the pseudo-code for PCSD-JIGM. 
	\begin{algorithm}
		\caption{\textsc{PCSD-JIGM}}
		\label{alg:pcsd_gaussian}
		\begin{algorithmic}[1]
			\STATE \textbf{Data Structures and Parameters}
			\STATE \quad $\bm{\theta}$: $\theta_i$ is the status of sensor S$_i$
			\STATE \quad $\bm{\mu}$: $\mu_i$ is the mean value of x$_i$
			\STATE \quad $\bm{\Sigma}$: covariance matrix
			\STATE \quad $\tau$: anomaly detection threshold
			\STATE \quad $\alpha$: confidence level for $\alpha$-consistency
			\STATE \quad $\lambda$: learning rate
			\STATE \textbf{Procedure}
			\STATE \quad $\theta_i \gets 1$ for all $i=1,\ldots,n$  \COMMENT{initialization}
			\STATE \quad $\theta'_i \gets 1$ for all $i=1,\ldots,n$ \COMMENT{initialization}
			\STATE \quad Fit $\bm{\mu}$ and $\bm{\Sigma}$ to the training dataset
			\STATE $\quad \textbf{D} \gets$ an $n \times n$ matrix of all 1s
			\REPEAT
			\FOR{$i = 1,\ldots,n$}
			\STATE $\textbf{D}[j,j] \gets 1/\theta_j$ for $j=1,\ldots,n$
			\STATE $\bm{\tilde{\Sigma}} \leftarrow \textbf{D} \odot \bm{\Sigma}$  \COMMENT{element-wise multiplication}
			\STATE $\hat{\mu}_i \gets \mu_i + \bm{\tilde{\Sigma}}_{i,-i}\bm{\tilde{\Sigma}}_{-i,-i}^{-1}(\textbf{x}_{-i}-\bm{\mu}_{-i})$
			\STATE $\hat{\sigma}^2_i \leftarrow \bm{\tilde{\Sigma}}_{i,i} - \bm{\tilde{\Sigma}}_{i,-i}\bm{\tilde{\Sigma}}_{-i,-i}^{-1}\bm{\tilde{\Sigma}}_{-i,i}$
			\STATE $p_i \gets 2\min\{F(\text{x}_i; \hat{\mu}_i, \hat{\sigma}^2_i), 1 - F(\text{x}_i; \hat{\mu}_i, \hat{\sigma}^2_i) \}$ 
			\STATE $\theta'_i \gets 1 - \max\{\alpha - p_i, 0\}$
			\ENDFOR
			\FOR{$i = 1,\ldots,n$}
			\STATE $\theta_i \gets \theta_i + \lambda (\theta'_i - \theta_i)$
			\ENDFOR
			\UNTIL{$\bm{\theta}$ converges}
			\STATE $\Delta \gets \{\text{S}_i: \theta_i < \tau, \quad \forall i = 1,\ldots,n  \}$
			\RETURN{} $\Delta$
		\end{algorithmic}
	\end{algorithm}

	\section{Experiments and Results} \label{sec:ch2_experiments}
	
	We empirically assess the proposed probabilistic consistency-based diagnosis methods on the task of detecting blocked rain gauges. Blockages, caused by debris, can obstruct the funnel hole or restrict the sensor's measurement of rainfall. A completely blocked gauge will inaccurately report zero rainfall, during rainy days. Our evaluation utilizes rainfall data from two weather sensor networks, testing the methods in two scenarios: one with spatially random distribution of faulty sensors and another with spatially clustered faults. We compare the performance of our methods against four baseline quality control (QC) methods.
	
	Before discussing the results, we will outline the evaluation setup, detail the datasets and metrics employed, and introduce the baseline methods against which we will compare the performance.
	
	\subsection{Datasets} \label{subsec:ch2_dataset}
\begin{figure}
		\includegraphics[width=1.0\linewidth]{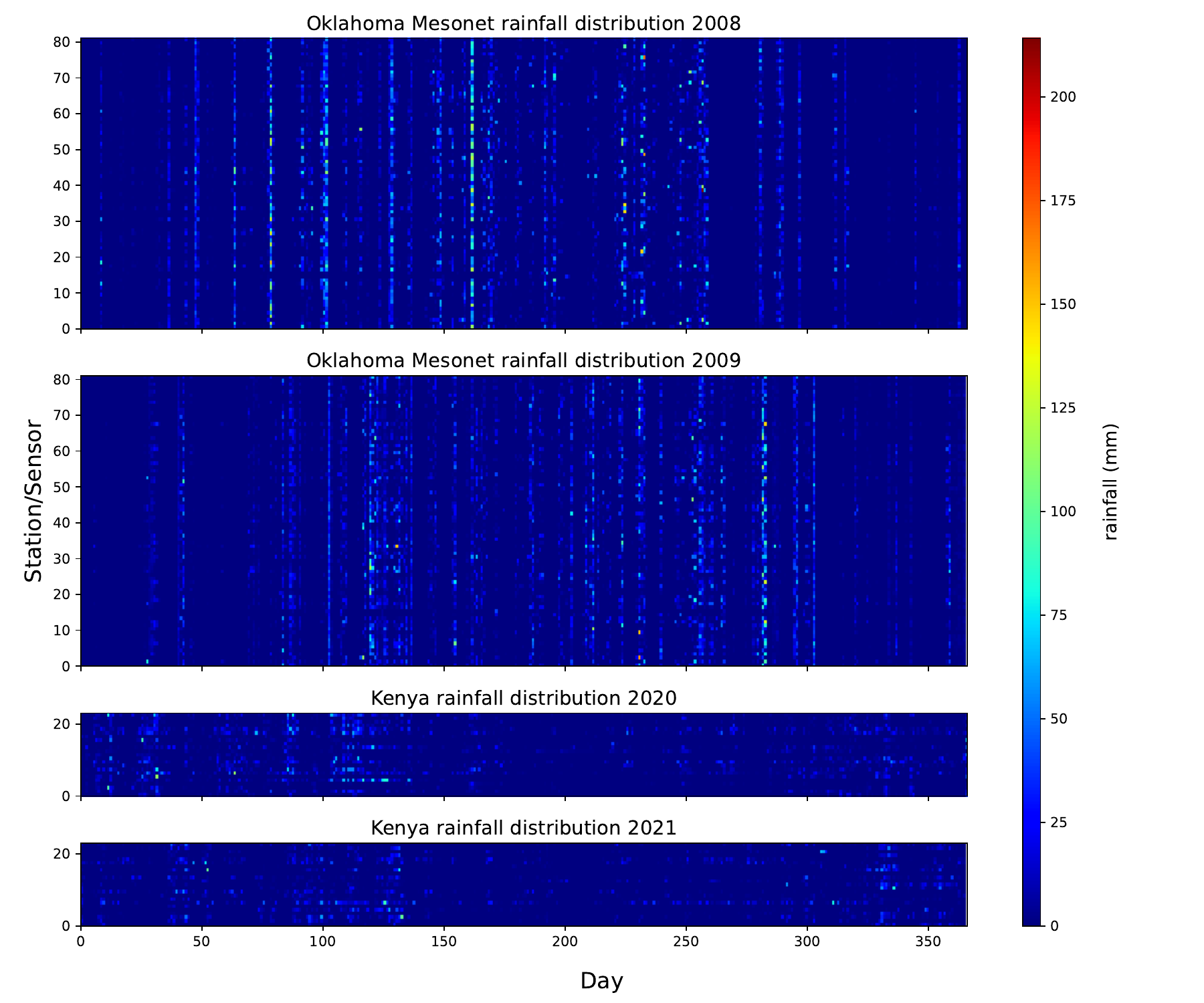}
		\caption{Historical training and testing rainfall distribution in the Oklahoma Mesonet and the Kenya dataset.}
		\label{fig:ch2_rainfall_distribution}
	\end{figure}

	Precipitation data from two weather networks are used to evaluate the performance of the QC methods described in Section \ref{sec:ch2_methods}. The first dataset contains weather data from $81$ weather stations drawn from the Oklahoma Mesonet dataset \citep{brock1995oklahoma, mcpherson2007statewide}, collected in $2008$ and $2009$. The Oklahoma Mesonet dataset contains measurements for several weather variables including precipitation, air temperature, atmospheric pressure, relative humidity, wind speed, wind direction, and solar radiation. The observations are reported every $5$ minutes with a time stamp along with a quality flag indicating instances of anomalous measurements or obvious sensor failures --- these sensor failures include out of band readings (e.g., $-999$ mm of rainfall) and other type of fault. Our experimental approach is to simulate clogged rain gauges by replacing the observed precipitation with zeros. Hence, we discard observations from a station for any day in which any weather variable was flagged as bad. The Oklahoma Mesonet has excellent quality control procedures \cite{mcpherson2007statewide}, so we can be confident that the remaining observations are correct (up to sensor measurement resolution). The second dataset consists of precipitation measurements from $23$ weather stations from the Trans‐African Hydro‐Meteorological Observatory (TAHMO) \citep{selker-2014-tahmo} network collected in Kenya during the years $2020$ and $2021$. As with the Oklahoma data, these observations have been carefully checked to ensure that they are valid. Figure \ref{fig:ch2_rainfall_distribution} presents a distribution of rainfall in training and test datasets from the Oklahoma Mesonet and the TAHMO Kenya weather networks.
	
	We simulate fully and partially blocked rain gauges by inserting artificial faults into clean data in a way that reflects the expected behavior of blocked rain gauges. For the simulation of fully blocked rain gauges, we randomly replace a contiguous fraction of the observed values at a small proportion $\beta$ (e.g., $\beta = 15 \%$) of all rain gauges with zeros. 
	
\begin{figure}
		\centering
		\subfloat[]{\includegraphics[width=0.49\textwidth]{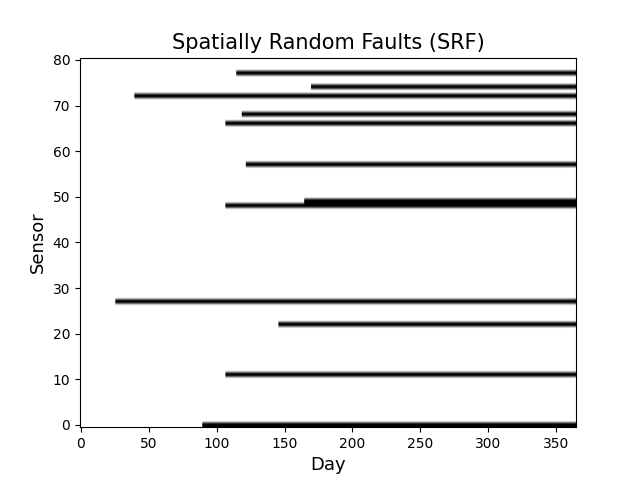}
			\label{fig:ok_example_spatially_random}}
		\hfill
		\subfloat[]{\includegraphics[width=0.49\textwidth]{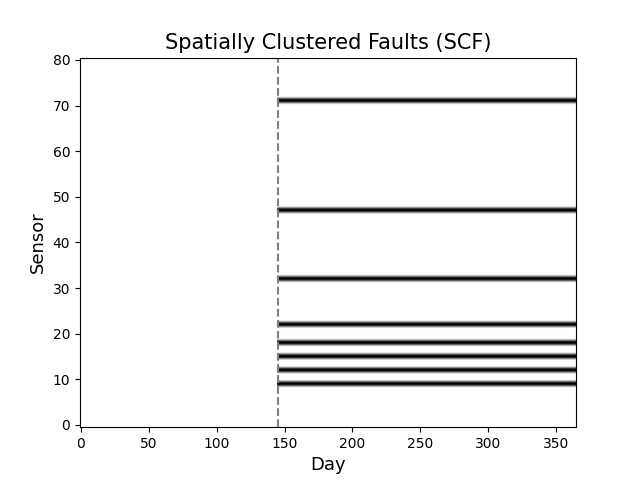}\label{fig:ok_example_spatially_clustered}}
		\caption{Distribution of faulty sensors in the Oklahoma Mesonet dataset. (a) Spatially random blocked rain gauges. (b) Spatially clustered blocked rain gauges with cluster size 8 and fault onset day $145$.}
	\end{figure}

	The faulty sensors can either exhibit spatially random or spatially clustered faults. For spatially random faults (SRF), faulty sensors are randomly selected and the onset dates of the anomalies are independent. Conversely, for spatially clustered faults (SCF), faulty sensors are geographical neighbors, with a single fault onset date applied within each cluster. We evaluate clustered faults under two scenarios, one where the cluster of faulty stations is surrounded by correctly working stations, which we refer to as SCF-IN, and a second scenario when at least one of the faulty sensors in the cluster is at the periphery of the network. We refer to the latter type of cluster faults as SCF-OUT. For both SCF-IN and SCF-OUT, we simulate different cluster sizes ranging from $2$ to $5$ for the TAHMO Kenya dataset and from $2$ to $8$ for the Oklahoma Mesonet dataset. Figure \ref{fig:ok_example_spatially_random} and \ref{fig:ok_example_spatially_clustered} illustrate examples of spatially random faults with $12$ faulty sensors out of $81$, and spatially clustered faults of size $8$ ($8$ faulty sensors out of the $81$) in the Oklahoma Mesonet test set.
	
	\subsection{Evaluation Metrics}
	Precision, recall, true positive rate (TPR), false positive rate (FPR), and the area under the receiver operating characteristic curve (AUROC) are the key metrics commonly used for binary classification. While these metrics are suitable when the objective is to maximize the prediction accuracy, our quality control context prioritizes the early detection of malfunctioning sensors to facilitate timely repair. In this scenario, the first alarm raised by a QC system serves as the sole indicator for sensor network managers to initiate a trouble ticket. Therefore, we evaluate the QC methods studied in this paper based on their average detection delay ($\bar{\Delta}$) and false alarm rate (FAR) at a fixed missed alarm rate (MAR). The detection delay is the time span between the onset of fault to the time the QC system raises an alarm. If an alarm is raised on a sensor that has no fault, or if the sensor is faulty but the alarm is raised prior to the onset of the fault, this is considered a false alarm. If an alarm is never raised on a faulty sensor, we consider this a missed alarm. Formally, let $t_i \in \mathbb{N}$ be the time of onset of the fault in sensor S$_i$, such that $t_i=-\infty$ if S$_i$ is never faulty, and let $\hat{t}_i$ be the predicted fault onset time for S$_i$. Let $f_i \in \{0, 1\}$ and $m_i \in \{0, 1\}$ indicate if there is a false alarm or a missed alarm for S$_i$, such that $f_i=1$ if the QC algorithm raised a false alarm on S$_i$, and $f_i=0$ otherwise. Similarly, $m_i=1$ if the QC algorithm missed a true alarm on S$_i$, and $m_i=0$ otherwise. Note that $f_i=1$ and $m_i=1$ cannot be true at the same time for the same sensor. With these definitions, we can compute the average detection delay, false alarm rate, and missed alarm rates as
	
	\begin{equation}
		\bar{\Delta} = \frac{\sum_{i=1}^n 1_{\{t_i \neq -\infty\}} \times \max (\hat{t}_i - t_i, 0)}{\sum_i^n 1_{\{t_i \neq -\infty\}}}
	\end{equation}
	\begin{equation}
		\text{FAR} = \frac{\sum_{i=1}^n f_i}{\sum_{i=1}^n 1_{\{t_i = -\infty\}}} 
	\end{equation}
	\begin{equation}
		\text{MAR} = \frac{\sum_{i=1}^n m_i}{\sum_{i=1}^n 1_{\{t_i \neq -\infty\}}} 
	\end{equation}
	
	where $1_{\{\text{z}\}}$ takes value $1$ when z is true, and $0$ otherwise.
	
	\subsection{Baseline Methods}
	We compare the joint reasoning methods PCSD-SAT, PCSD-JIKR, and PCSD-JIGM with the following baseline QC methods.
	
	\begin{itemize}
		\item Joint Gaussian model (GM): fits a joint Gaussian model to the historical data $\mathcal{D}_{train}$ observed at all stations and employs the conditional distribution of each station conditioned on the rest of the network to predict the expected rainfall. Faults are detected by applying a threshold to the likelihood of observed rainfall under the conditional distribution.
		\item Conditional mixture model (CMM) \cite{zemicheal2020conditional}: fits a log-linear regression to the historical data $\mathcal{D}_{train}$ observed at each station and up to five neighboring stations. The probability of zero rain is handled by a second logistic regression model for predicting the presence or absence of rain.
		\item PWSQC (Fz-filter) \cite{de2019quality}: first calculates the median rainfall reported by neighboring sensors, excluding any missing values. If the median is greater than zero, the day is classified as a ``wet day'', else it is classified as a ``dry day''. If a target sensor S$_i$ reports zero rainfall on a wet day, it is considered a ``wet zero'' day for sensor S$_i$. If S$_i$ experiences $z_{max}$ consecutive wet zero days (not counting dry days), it is declared to be blocked. This method is specifically designed only to detect clogged rain gauges.
		\item Kernel Regression (KR): a neighbor regression method that estimates the expected rainfall at each station as a weighted average of the neighbors' observation. The weights are inversely proportional to the distance between target station and each neighbor. We implement a version neighbor regression similar to \citep{mcpherson2007statewide}, but with a radial basis function (RBF) kernel for regression weight computation.
	\end{itemize}

	\subsection{Experiment Setup}
	
	We introduce simulated faults into $20$ copies of the two datasets described in section \ref{subsec:ch2_dataset} for both spatially random and clustered faults, using different random seeds to evaluate the models’ effectiveness in detecting these faults. For the CMM method, we consider up to five closest neighbors within $100$ km of each target sensor/station S$_i$; for all other methods, we include all available sensors as neighbors when diagnosing S$_i$. 
	
	The length-scale parameters for the kernel regression models are empirically estimated using the training set to minimize the mean squared residuals. A length scale of $60$ achieved the lowest error. The learning rate for $\lambda$ in PCSD-JIKR and PCSD-JIGM was set to $\lambda=0.1$. The IBM ILOG CPLEX optimizer was invoked to solve the integer linear programming problems in PCSD-SAT \cite{cplex2009v12}. Lastly, we used values of $\alpha=0.3$, $\alpha=0.4$ and $\alpha=0.4$ for $\alpha$-consistency in PCSD-SAT, PCSD-JIKR, and PCSD-JIGM respectively. 
	
	We compare the proposed PCSD-SAT, PCSD-JIKR, and PCSD-JIGM methods against baseline approaches using average detection delay and false alarm rate (FAR) at $5\%$ missed alarm rate (MAR), which corresponds to a recall of $95\%$. Recall represents the proportion of actual faulty sensors correctly identified by a QC method out of all the faulty sensors. Both the PCSD-SAT method and the PWSQC (FZ-filter) baseline generate daily boolean fault indicators for each station. To evaluate average detection delay and FAR at $5\%$ MAR, we slide a window of varying sizes across each station's predictions, and declare an alarm when a sequence of flags matches the window size. We compute the average detection delay, MAR, and FAR for each window size (from 1 to the maximum number of consecutive days flagged by the QC method), and we report the average detection delay ($\bar{\Delta}$) and FAR at $5\%$ MAR. Other baseline QC methods output continuous scores, whereas PCSD-JIKR and PCSD-JIGM generate $\theta_i$ as their belief about the correct functioning state of sensor S$_i$. We use $1-\theta_i$ as the fault score for PCSD-JIKR and PCSD-JIGM. 
	
	Inspired by the PWSQC (Fz-filter) algorithm \cite{de2019quality}, which flags daily rainfall as anomalous when zero rainfall is reported while neighboring values are non-zero, we assess quality control (QC) methods based on varying levels of flag aggregation. Specifically, we evaluate these methods by selecting a fault threshold $\tau$ such that sensors with a fault score above $\tau$ on a given day are flagged. We then use a sliding window approach, akin to the PCSD-SAT and PWSQC (FZ-filter) techniques, to aggregate these flags. An alarm is triggered when the number of flags within the window equals the window size. For each window size, we report $\bar{\Delta}$ and FAR metrics that achieve a $5\%$ MAR. We consider window sizes ranging from 1 to 7 days, allowing up to a week of evidence to inform the diagnosis.
	
	High MAR or FAR can be costly in sensor networks: missed alarms may lead to the publication of inaccurate observations, while false alarms incur economic costs due to unnecessary technician visits. Our goal is to ensure data quality by minimizing missed alarms and subsequently reducing false alarms and detection delays.

	\subsection{Results and Discussion} \label{subsec:ch2_results}
	We assess the methods under three distinct fault spatial distributions:
	\begin{itemize}
		\item Spatially Random Faults (SRF): Faulty sensors are distributed randomly and independently across the network, each with different fault onsets.
		\item Spatially Clustered Faults Inside (SCF-IN): Faulty sensors are grouped as neighbors within the network, all sharing the same fault onset.
		\item Spatially Clustered Faults Outside (SCF-OUT): Faulty sensors are located at the periphery of the network, also sharing the same fault onset.
	\end{itemize}

	\subsubsection{Detection of Spatially Random Faults (SRF)}
\begin{figure}
		\setlength\tabcolsep{2pt}
		\centering
		\begin{tabular}{ccc}
			\includegraphics[width=0.49\textwidth]{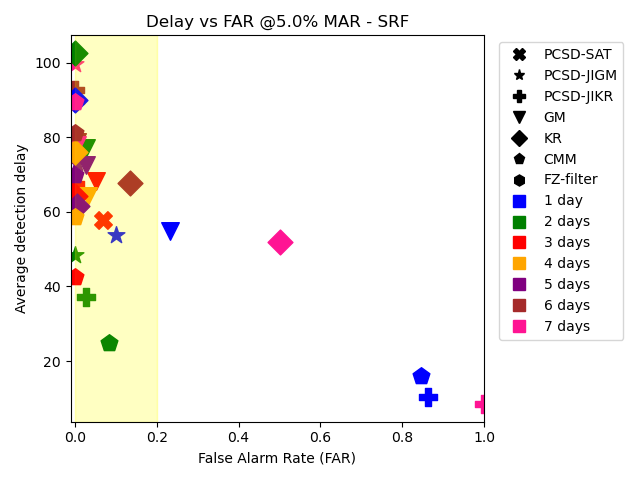} &
			\includegraphics[width=0.49\textwidth]{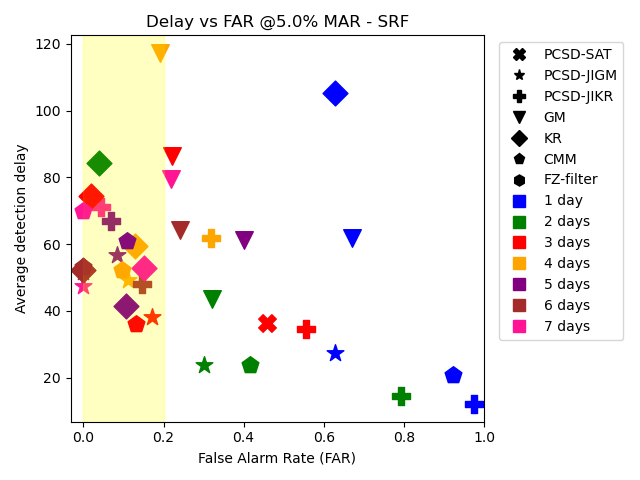}
		\end{tabular}
		\caption{Detection performance for spatially random faults (SRF) in both the Oklahoma Mesonet (left) and the TAHMO Kenya (right) networks, with $15\%$ faulty sensors. The shapes represent the methods and the colors represent the number of days of evidence need to mark a sensor as faulty.}
		\label{fig:ch2_performance_SRF_0.15_faulty}
	\end{figure}

\begin{figure}\centering
		\subfloat[Rainfall observed on day $86$ in the Oklahoma Mesonet test dataset. The sensor locations with red labels are the ground truth faulty sensors, and the color of the marker represents the amount of rainfall observed (see color-bar).]{\label{a1}\includegraphics[width=.99\linewidth]{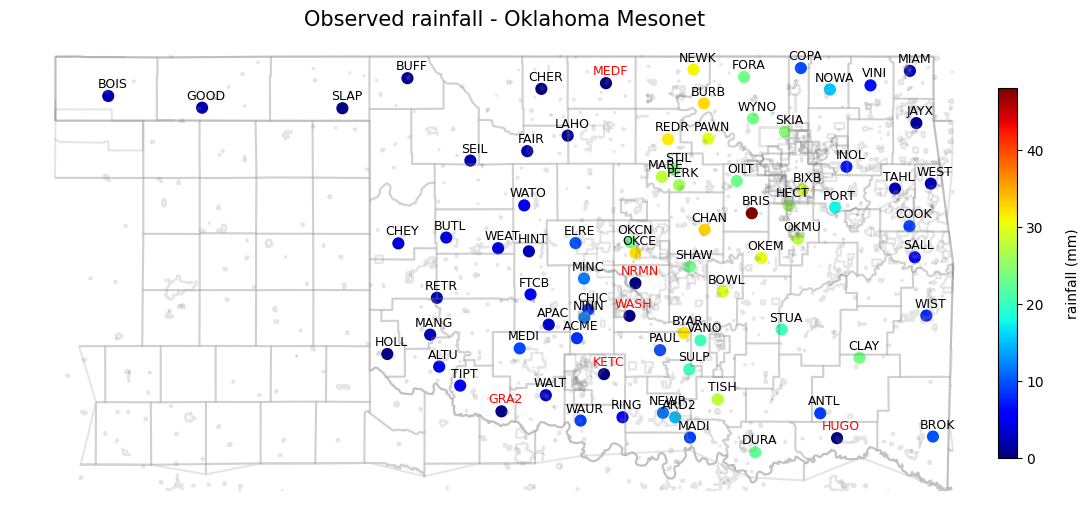}}\par
		\subfloat[Belief update steps by PCSD-JIGM.]{\label{b1}\includegraphics[width=.49\linewidth]{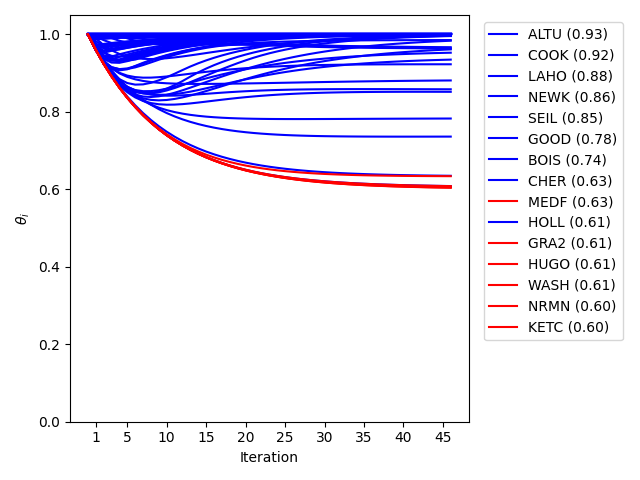}}\hfill
		\subfloat[Belief update steps by PCSD-JIKR.]{\label{c1}\includegraphics[width=.49\linewidth]{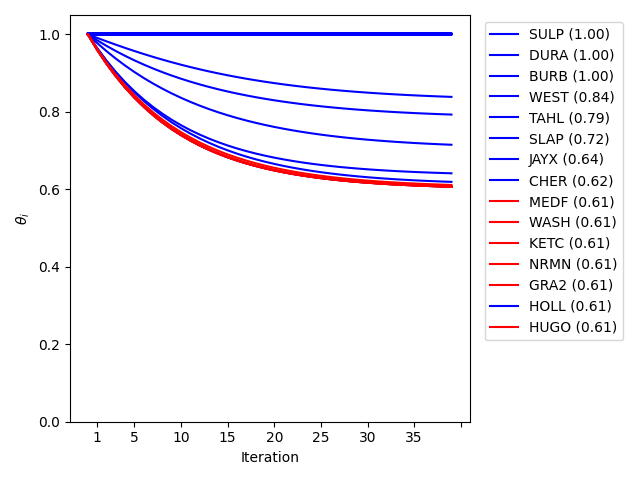}}
		\caption{Example of joint inference learning in Oklahoma Mesonet test dataset with spatially random faults (SRF) on day $86$. The labels show the final value of $\theta_i$s for the 15 sensors with the smallest values. The smaller the $\theta_i$, the stronger the belief that S$_i$ is faulty. The sensors with red colors are the ground truth faulty sensors.}
		\label{fig:joint_inference_example_ok_srf}
	\end{figure}

\begin{figure}\centering
		\subfloat[Rainfall observed on day $161$ in the Oklahoma Mesonet test dataset. The sensor locations with red labels are the ground truth faulty sensors, and the color of the marker represents the amount of rainfall observed (see color-bar).]{\label{a2}\includegraphics[width=.99\linewidth]{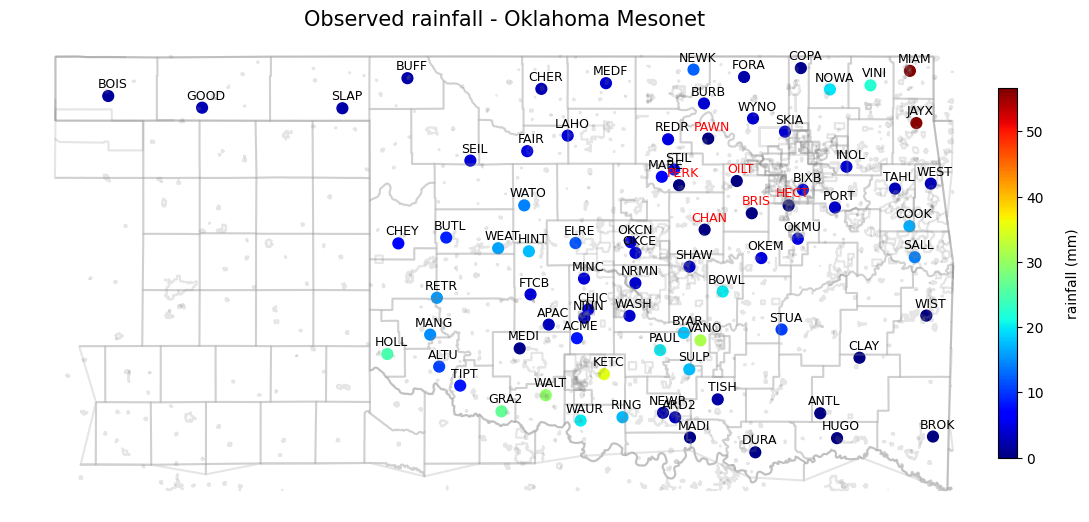}}\par
		\subfloat[Belief update steps by PCSD-JIGM.]{\label{b2}\includegraphics[width=.49\linewidth]{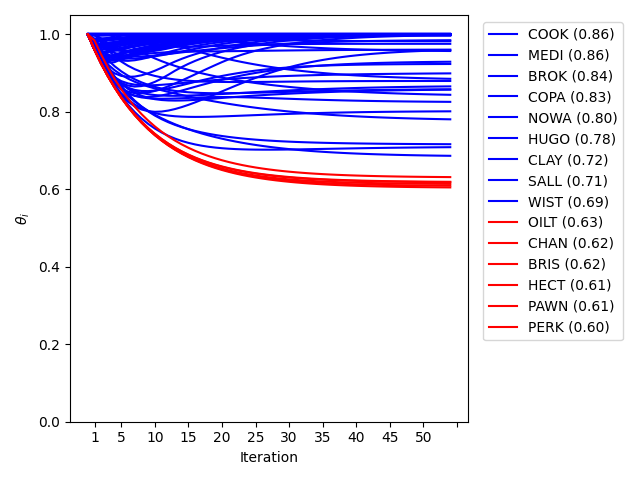}}\hfill
		\subfloat[Belief update steps by PCSD-JIKR.]{\label{c2}\includegraphics[width=.49\linewidth]{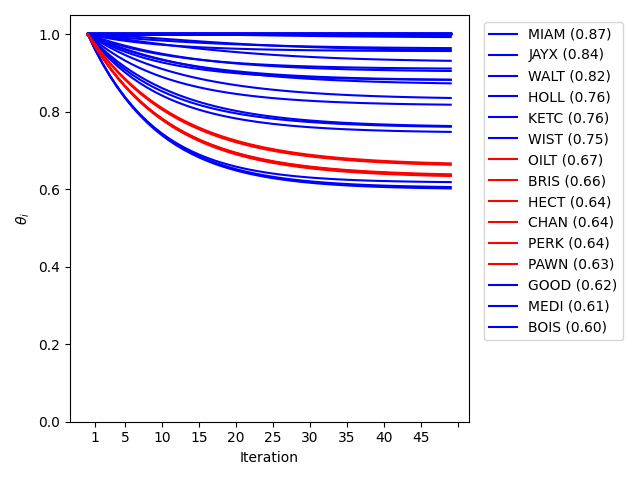}}
		\caption{Example of joint inference learning in Oklahoma Mesonet test dataset with spatially clustered faults (SCF) on day $161$. The labels show the final value of $\theta_i$s for the 15 sensors with the smallest values. The smaller the $\theta_i$, the stronger the belief that S$_i$ is faulty. The sensors with red colors are the ground truth faulty sensors.}
		\label{fig:joint_inference_example_ok_scf}
	\end{figure}

	In the SRF fault configuration, we evaluate the quality control (QC) methods using both the Oklahoma Mesonet and the TAHMO Kenya network, each with $15\%$ faulty sensors. Our goal is to minimize detection delay, missed alarms, and false alarms. We compare the methods based on their average detection delay at a fixed missed alarm rate (MAR) of $5\%$ (equivalent to $95\%$ recall) and various maximum false alarm rates (FAR). Specifically, we consider FARs of $5\%$, $10\%$, $15\%$, and $20\%$. We aim to determine how long each method takes to detect at least $95\%$ of faulty sensors while maintaining FARs below these thresholds. Figure \ref{fig:ch2_performance_SRF_0.15_faulty} illustrates the trade-off between average detection delay ($\bar{\Delta}$) and FAR for both networks. Tables \ref{tab:ch2_ke-add_vs_far-srf} and \ref{tab:ch2_ok-add_vs_far-srf} summarize performance within the FAR ranges of $0\%$ to $5\%$, $10\%$, $15\%$, and $20\%$. \\
	\\
	In the Oklahoma Mesonet network, the CMM method demonstrates the best detection performance with an average detection delay of 23 days at FARs of $10\%$, $15\%$, and $20\%$. PCSD-JIKR follows with a delay of 36 days. Other methods, such as PCSD-JIGM, PCSD-SAT, KR, and GM, have delays of 44, 58, 63, and 66 days, respectively. In the more constrained FAR range of $5\%$, PCSD-JIKR remains the best with a delay of 36 days. Figures \ref{fig:joint_inference_example_ok_srf} and \ref{fig:joint_inference_example_ok_scf} provide examples of rainfall data from the Oklahoma Mesonet, highlighting faulty sensors and showing the evolution of $\bm{\theta}$ for PCSD-JIKR and PCSD-JIGM during days 86 and 161. \\
	\\
	In the TAHMO Kenya network, CMM also achieves the best performance at FARs of $15\%$ and $20\%$, with an average detection delay of 38 days. At FAR $15\%$, KR, PCSD-JIGM, and PCSD-JIKR show delays of 41, 48, and 48 days, respectively. At FAR $20\%$, KR, PCSD-JIGM, and GM have delays of 41, 39, and 118 days, respectively. For the more stringent FAR ranges of $5\%$ and $10\%$, PCSD-JIGM is the best performer with a delay of 48 days. \\
	\\
	For spatially random faults, when a FAR greater than $5\%$ is unacceptable, the consistency-based methods PCSD-JIGM and PCSD-JIKR are preferable. If a slightly higher FAR is tolerable, CMM provides a significantly lower detection delay. Additionally, the Oklahoma Mesonet network shows shorter detection delays compared to TAHMO Kenya, likely due to its denser and more uniformly distributed stations, as well as more frequent heavy rainfall (see Figure \ref{fig:ch2_rainfall_distribution}). The PWSQC (Fz-filter) method is absent from the because its detection delays predominantly exceed the stringent MAR requirement of $5\%$.

	\begin{table}
		\centering    
		\begin{tabular}{c|c|c|c|c}        
			Method & FAR $\leq 5\%$ & FAR $\leq 10\%$ & FAR $\leq 15\%$ & FAR $\leq 20\%$ \\
			\hline                  
			PCSD-SAT & - & 58 & 58 & 58  \\
			PCSD-JIGM & 44 & 44 & 44 & 44 \\
			PCSD-JIKR & \bf36 & 36 & 36 & 36 \\
			GM & 66 & 66 & 66 & 66 \\
			KR & 63 & 63 & 63 & 63 \\
			CMM & 42 & \bf23 & \bf23 & \bf23 \\
			Fz-filter & - & - & - & - \\
		\end{tabular}    
		\captionsetup{justification=centering}
		\caption{Average detection delay in the Oklahoma Mesonet network at a maximum MAR of 5\%, and a maximum FAR of 5\%, 10\%, 15\%, and 20\%. The best performing QC method in boldface.}
		\label{tab:ch2_ok-add_vs_far-srf}
	\end{table}
	
	\begin{table}
		\centering    
		\begin{tabular}{c|c|c|c|c}        
			Method & FAR $\leq 5\%$ & FAR $\leq 10\%$ & FAR $\leq 15\%$ & FAR $\leq 20\%$ \\
			\hline                  
			PCSD-SAT & - & - & - & -  \\
			PCSD-JIGM & \bf48 & \bf48 & 48& 39 \\
			PCSD-JIKR & 71 & 67 & 48 &  \\
			GM & - & - & - & 118 \\
			KR & 51 & 51 & 41 & 41 \\
			CMM & 70 & 51 & \bf38 & \bf38 \\
			Fz-filter & - & - & - & - \\
		\end{tabular}    
		\captionsetup{justification=centering}
		\caption{Average detection delay in the TAHMO Kenya network at a maximum MAR of 5\%, and a maximum FAR of 5\%, 10\%, 15\%, and 20\%. The best performing QC method in boldface.}
		\label{tab:ch2_ke-add_vs_far-srf}
	\end{table}

	\subsubsection{Detection of Clustered Faults}
	We compare the QC methods in the two clustered faults configurations SCF-IN and SCF-OUT. The two settings offer different challenges for the QC methods. When a cluster of faulty sensors is surrounded by nominal sensors, there's more opportunity for detection than when the cluster borders the edge of the network. In principle, on heavily rainy days, the sensors at the edge of the cluster will pull signal from nominal ones that are outside the cluster, increasing the chances of the QC methods to flag those faulty sensors. When the clusters are at the periphery of the network, however, the majority of the neighbors of those faulty sensors will also be faulty, which makes harder to detect them due to a lower signal-to-noise ratio.
	
	We evaluate the average detection delay ($\bar{\Delta}$) of various QC methods at $5\%$ MAR for clusters of sizes 2-8 in the Oklahoma Mesonet network and sizes 2-5 in the TAHMO Kenya network. The TAHMO Kenya network is too small to permit larger cluster sizes. We examine performance in two FAR ranges: $\leq 10\%$ and $\leq 20\%$. Figure \ref{fig:ch2_performance_ok_scf} and \ref{fig:ch2_performance_ke_scf} present the trade-off between average detection delay and FAR in both the Oklahoma Mesonet and the TAHMO Kenya network, while Table \ref{tab:ch2_ok-add_vs_far-scf} and \ref{tab:ch2_ke-add_vs_far-scf} summarize each QC method's best performance within the $5\%$ MAR and the two FAR ranges $\leq 10\%$ and $\leq 20\%$, in both networks. 
	
	In the Oklahoma Mesonet network:
	\begin{itemize}
		\item At MAR = $5\%$ and FAR $\leq 20\%$, CMM achieves the lowest detection delays for clusters of sizes 2-5, with average delays of 50, 37, 37, and 38 days. PCSD-JIKR and PCSD-JIGM are next, performing best for clusters of sizes 2-4 and 4-5, respectively. For clusters larger than 5, CMM’s performance degrades, while PCSD-JIGM and PCSD-JIKR outperform other methods.
		\item In the stricter performance space (MAR = $5\%$ and FAR $\leq 10\%$), CMM is best for clusters of size 2-4. PCSD-JIGM excels overall, with detection delays of 44, 47, 48, and 42 days for sizes 5-8. PCSD-JIKR has the lowest delay for size 7 at 33 days. GM performs well for clusters of sizes 6 and 8, with delays of 60 and 54 days.
	\end{itemize} 
	
	For SCF-OUT in the Oklahoma Mesonet network:
	\begin{itemize}
		\item At MAR = $5\%$ and FAR $\leq 20\%$, CMM leads for cluster sizes 2 and 4, with delays of 40 and 36 days. PCSD-JIGM has the lowest delays for sizes 3 (39 days) and 5-8 (46, 48, and 30 days).
		\item In the stricter MAR = $5\%$ and FAR $\leq 10\%$ space, PCSD-JIGM achieves the lowest delays for sizes 3 and 5-8.
	\end{itemize} 
	
	In the TAHMO Kenya network:
	\begin{itemize}
		\item At MAR = $5\%$ and FAR $\leq 20\%$, different methods excel in different cluster sizes. For size 2, CMM and PCSD-JIGM achieve the best delays of 25 and 47 days, respectively. For size 3, PCSD-JIKR and CMM perform best with delays of 62 and 63 days. For clusters larger than 3, CMM's performance drops, while GM and PCSD-JIKR perform well.
		\item In the MAR = $5\%$ and FAR $\leq 10\%$ space, PCSD-JIGM and KR perform best for sizes 2 and 3. PCSD-JIKR and KR are best for size 4. In the stricter MAR = $5\%$ and FAR $\leq 5\%$ space, PCSD-JIGM leads for sizes 2-4, with delays of 47, 87, and 83 days. For SCF-OUT, PCSD-JIGM and CMM achieve detection delays of 44 and 66 days for size 2, with PCSD-JIGM being the only method to perform within the space for size 3 (74 days).
	\end{itemize}
	
	In summary, the detection delays of SCF are higher than those of SRF. This is expected, since a cluster of simultaneously faulty sensors provides a weaker signal when diagnosing faults within the cluster and therefore reduces the likelihood of timely detection. For spatially clustered faults, in the Oklahoma Mesonet network, CMM excels when the cluster size does not exceed 4. Beyond size 4, CMM breaks down if more than $10\%$ FAR is unacceptable, while consistency-based methods PCSD-JIKR and PCSD-JIGM outperform all baselines. PCSD-JIKR and PCSD-JIGM also offer the second lowest detection delay in cluster sizes less than or equal to 4. When the faulty sensors are at the periphery, PCSD-JIGM is generally the best model at low FAR of $10\%$ or less, with GM generally being the second best method. CMM's leading performance in small size clusters can be explained by its reliance on local neighbors solely for fault detection. This gives CMM a high predictive power over the rest of the methods that use the contribution of all available sensors for diagnosis. As the cluster size gets bigger, the signal from the neighbors becomes weaker, and the predictive power of CMM decreases. In that bigger cluster setting, consistency-based methods PCSD-JIKR and PCSD-JIGM, through their joint reasoning process, increasingly discount the contribution of those faulty sensors that are at the periphery of the cluster and pull information from nominal sensors outside of the cluster. Throughout the iterations, the signal penetrates into the cluster as the contribution of those sensors at the periphery of the cluster is increasingly discounted. GM and KR also outperform CMM, albeit with longer detection delays than PCSD-JIKR and PCSD-JIGM. This is because despite not doing joint reasoning, GM and KR, use all available neighbors for detection, and are still able to extract some signals from distant sensors in SCF. 
	
	For TAHMO Kenya, only PCSD-JIGM is achieves the lowest detection delay in more than one cluster size. With different methods scoring the best in different cluster sizes. This highlights the dominance of dry days in the TAHMO Kenya dataset, coupled with the sparse geographical distribution of the sensors. 
	
	The PCSD-SAT and FZ-filter methods, by their more rigid binary flagging system, result in them not being able to achieve $5\%$ MAR (or $95\%$ recall) without an exploding FAR.

	\begin{figure}
		\setlength\tabcolsep{2pt}
		\centering
		\begin{tabular}{ccc}
			\includegraphics[width=0.49\textwidth]{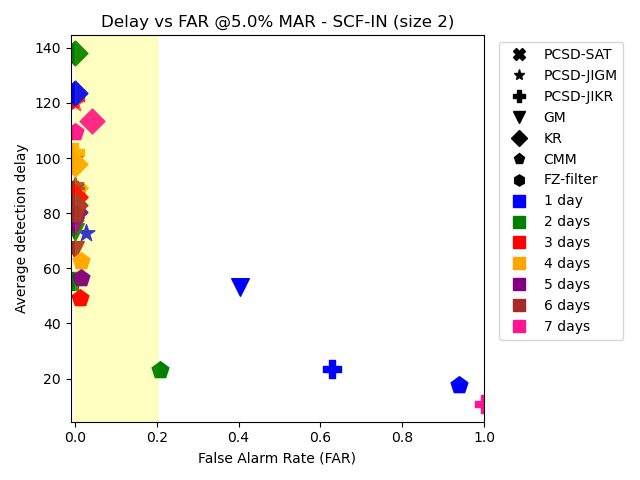} &
			\includegraphics[width=0.49\textwidth]{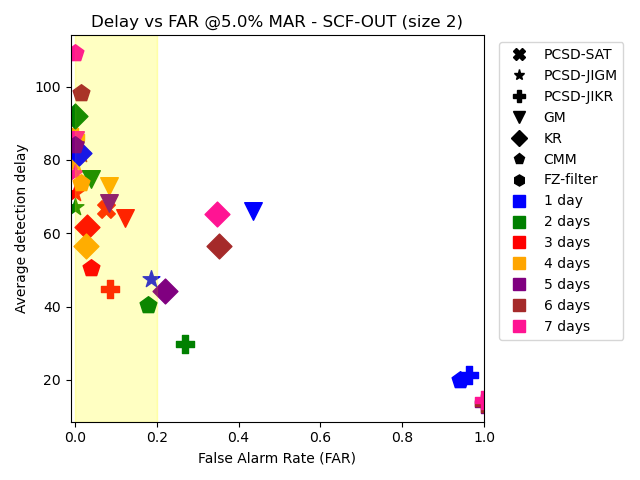} \\
			\includegraphics[width=0.49\textwidth]{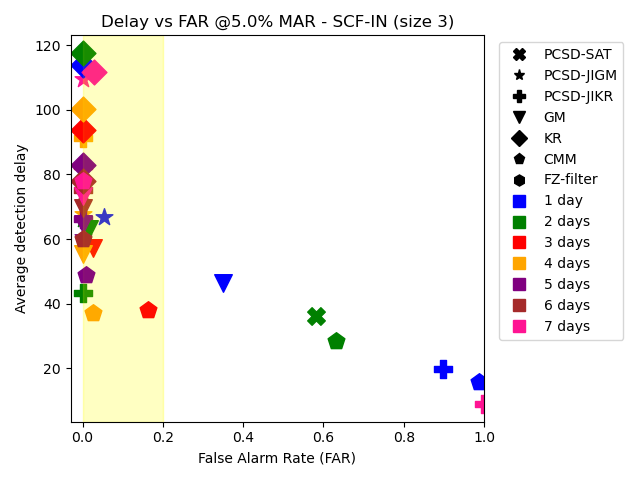} &
			\includegraphics[width=0.49\textwidth]{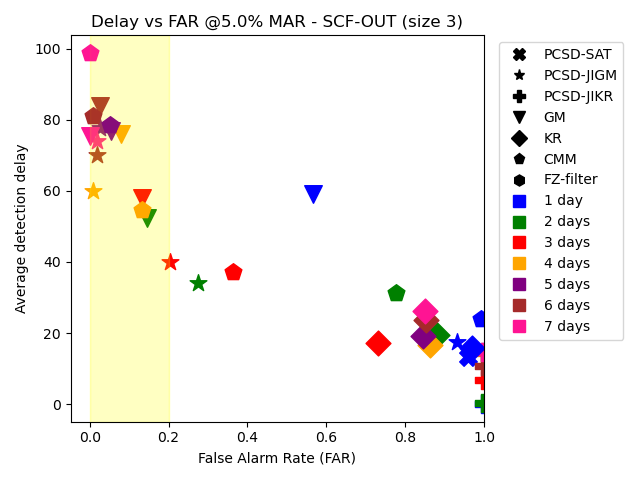} \\
			\includegraphics[width=0.49\textwidth]{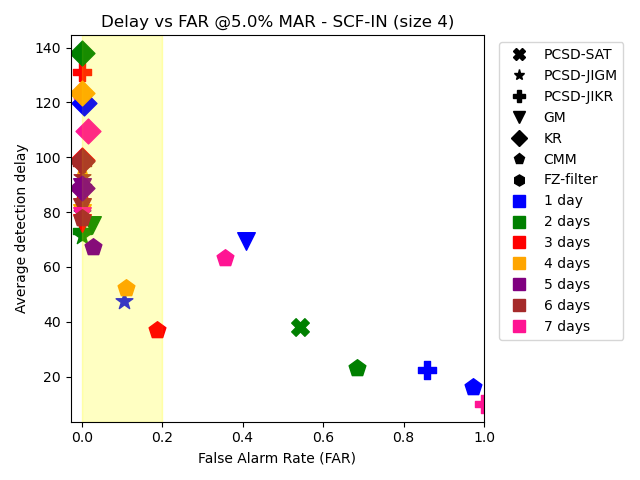} &
			\includegraphics[width=0.49\textwidth]{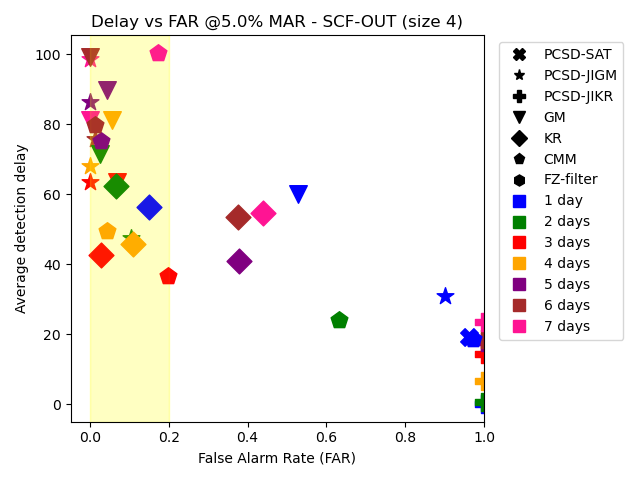} 
		\end{tabular}
		\caption{Detection performance for spatially cluster faults SCF-IN and SCF-OUT on the Oklahoma Mesonet dataset with cluster size 2, 3, and 4. The shapes represent the methods and the colors represent the number of days of evidence need to mark a sensor as faulty.}
	\end{figure}
	
	\begin{figure}\ContinuedFloat
		\setlength\tabcolsep{2pt}
		\centering
		\begin{tabular}{ccc}
			\includegraphics[width=0.49\textwidth]{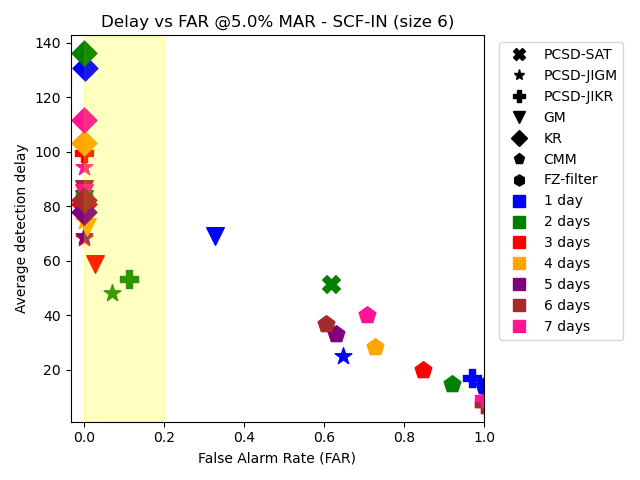} &
			\includegraphics[width=0.49\textwidth]{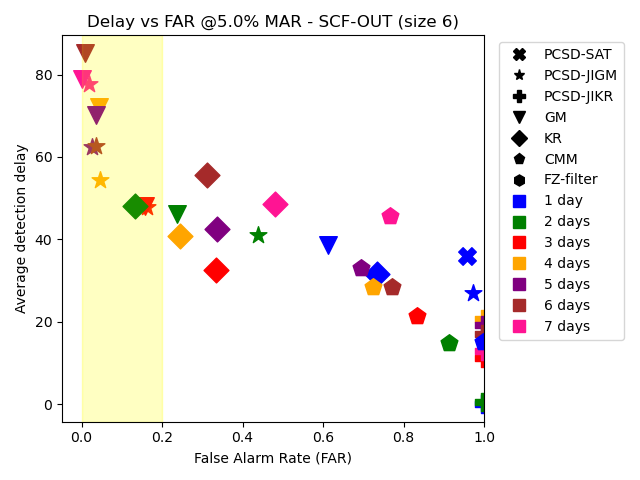} \\
			\includegraphics[width=0.49\textwidth]{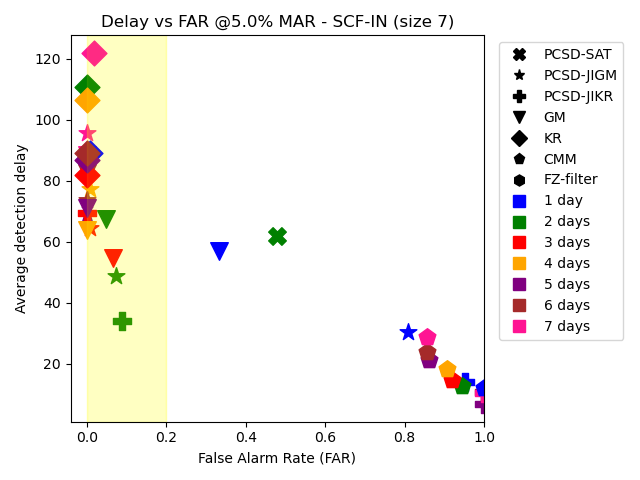} &
			\includegraphics[width=0.49\textwidth]{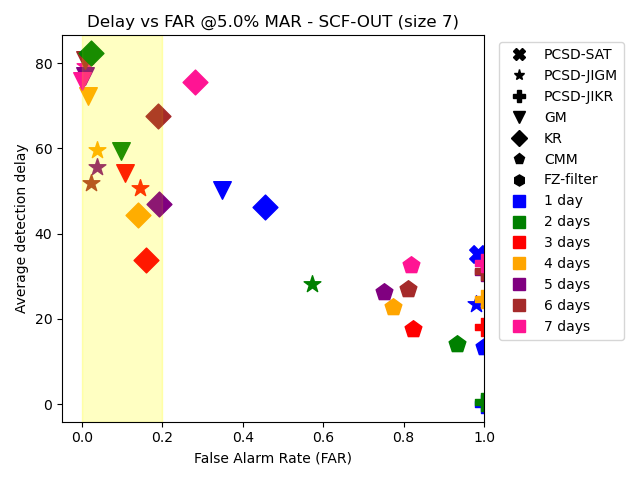} \\
			\includegraphics[width=0.49\textwidth]{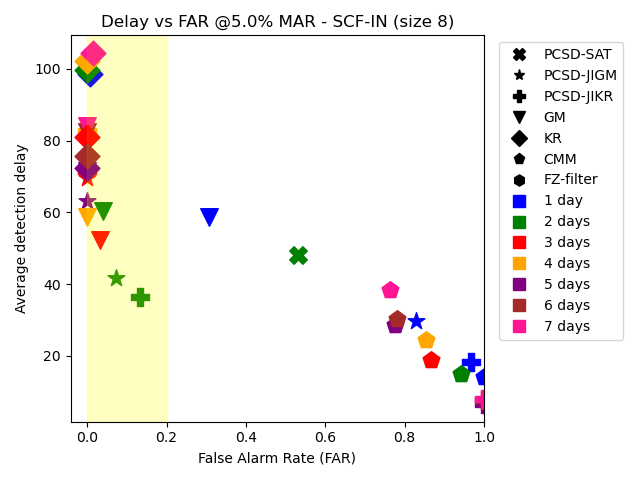} &
			\includegraphics[width=0.49\textwidth]{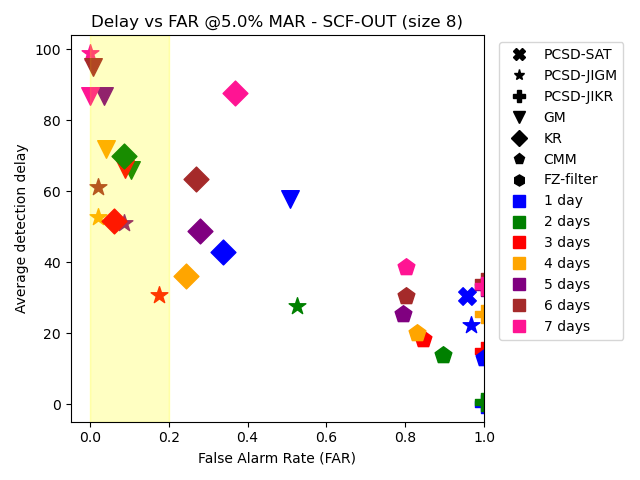} 
		\end{tabular}
		\caption{Detection performance for spatially cluster faults SCF-IN and SCF-OUT on the Oklahoma Mesonet dataset with cluster size 6, 7 and 8. The shapes represent the methods and the colors represent the number of days of evidence need to mark a sensor as faulty.}
		\label{fig:ch2_performance_ok_scf}
	\end{figure}

	\begin{figure}
		\setlength\tabcolsep{2pt}
		\centering
		\begin{tabular}{ccc}
			\includegraphics[width=0.49\textwidth]{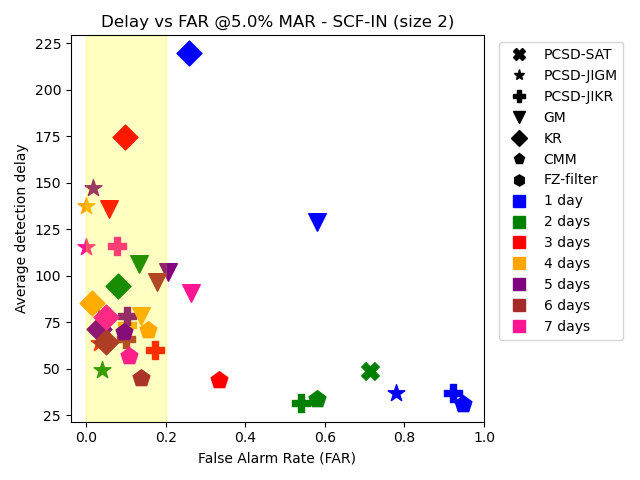} &
			\includegraphics[width=0.49\textwidth]{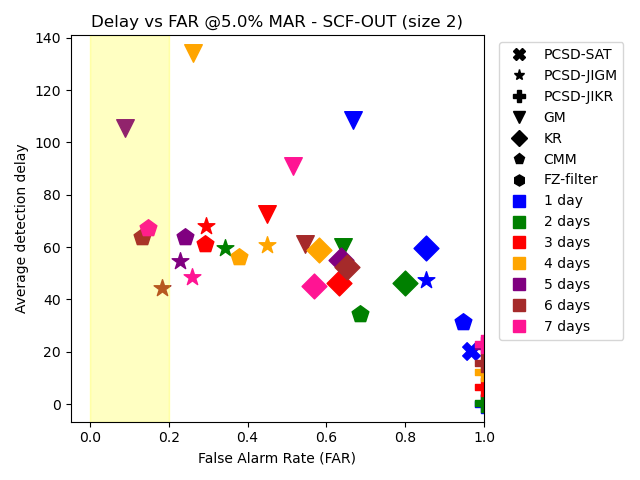} \\
			\includegraphics[width=0.49\textwidth]{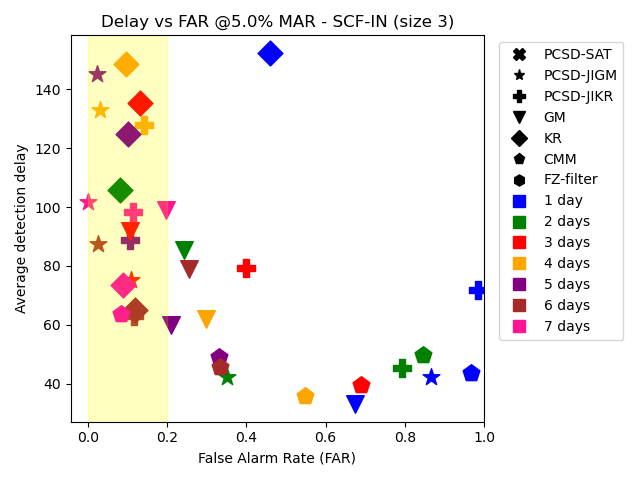} &
			\includegraphics[width=0.49\textwidth]{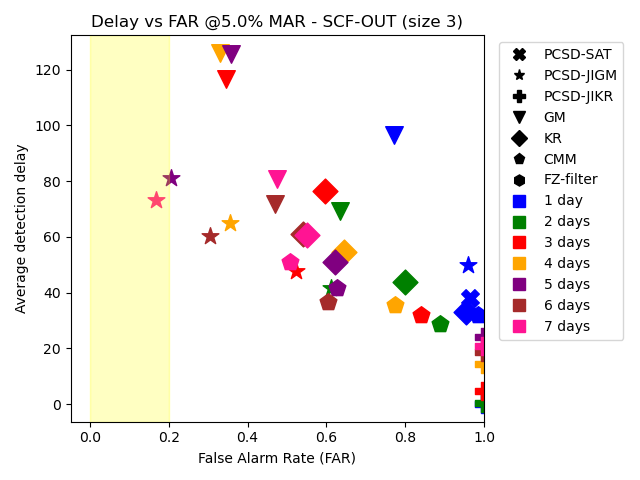} \\
			\includegraphics[width=0.49\textwidth]{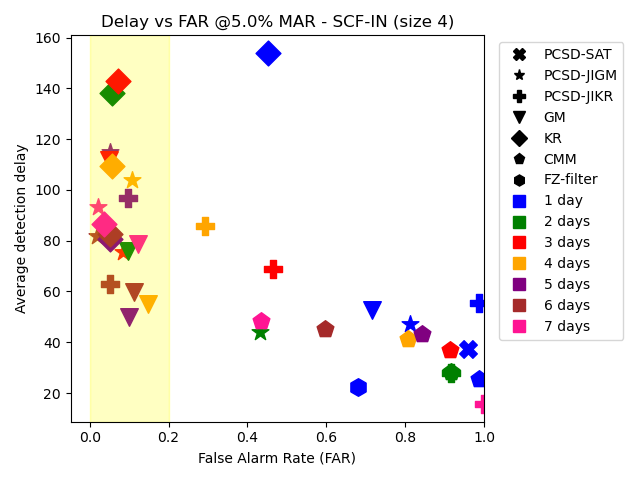} &
			\includegraphics[width=0.49\textwidth]{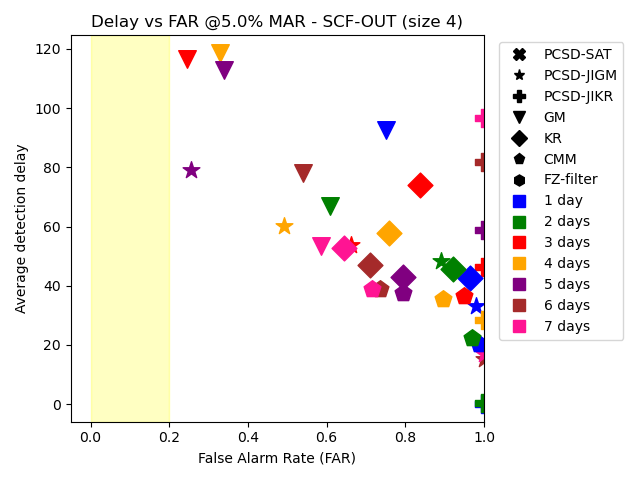} 
		\end{tabular}
		\caption{Detection performance for spatially cluster faults SCF-IN and SCF-OUT on the TAHMO Kenya dataset with cluster size 2, 3 and 4. The shapes represent the methods and the colors represent the number of days of evidence need to mark a sensor as faulty.}
	\end{figure}
	
	\begin{figure}\ContinuedFloat
		\setlength\tabcolsep{2pt}
		\centering
		\begin{tabular}{ccc}
			\includegraphics[width=0.49\textwidth]{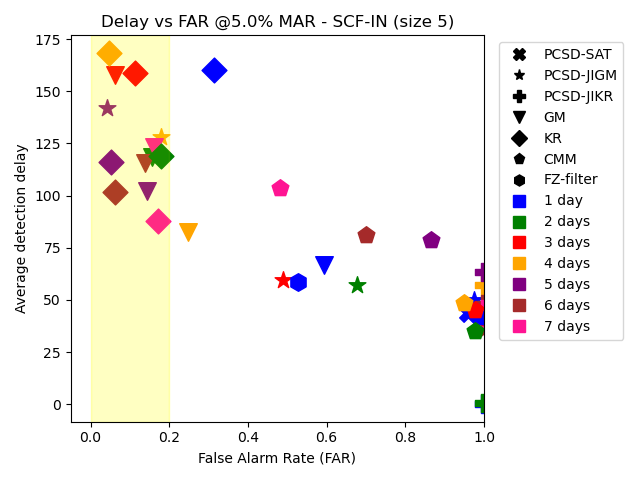} &
			\includegraphics[width=0.49\textwidth]{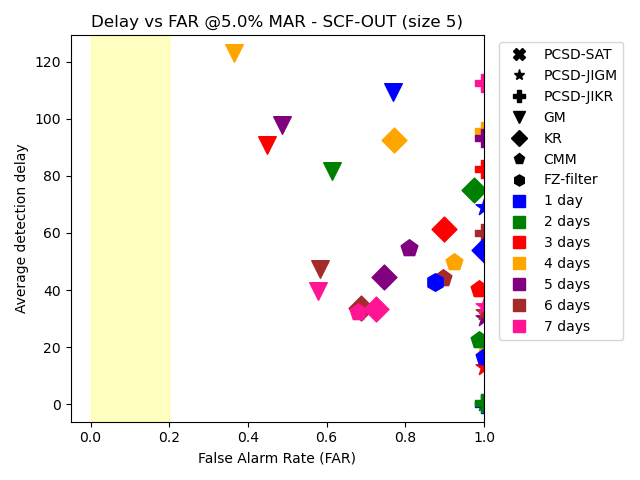}
		\end{tabular}
		\caption{Detection performance for spatially cluster faults SCF-IN and SCF-OUT on the TAHMO Kenya dataset with cluster size 5. The shapes represent the methods and the colors represent the number of days of evidence need to mark a sensor as faulty.}
		\label{fig:ch2_performance_ke_scf}
	\end{figure}

	\begin{table}
		\centering    
		\begin{tabular}{c|ccccccc|ccccccc}  
			& \multicolumn{7}{c|}{FAR $\leq 20\%$} & \multicolumn{7}{c}{FAR $\leq 10\%$} \\ 
			Method & 2 & 3 & 4 & 5 & 6 & 7 & 8 & 2 & 3 & 4 & 5 & 6 & 7 & 8 \\
			\hline                
			PCSD-SAT$_{IN}$ & - & - & - & - & - & - & -  & - & - & - & - & - & - & -  \\
			PCSD-JIGM$_{IN}$ & 71 & 67 & \underline{48} & \underline{44} & \bf47 & \underline{48} & \underline{42} & 71 & 67 & \underline{71} & \bf44 & \bf47 & \underline{48} & \bf42 \\
			PCSD-JIKR$_{IN}$ & \underline{56} & \underline{43} & 73 & 45 & \underline{53} & \bf33 & \bf35 & \underline{56} & \underline{43} & 73 & \underline{45} & 99 & \bf33 & 70 \\
			GM$_{IN}$ & 68 & 57 & 76 & 52 & 60 & 55 & 54 & 68 & 57 & 76 & 52 & \underline{60} & 55 & \underline{54} \\
			KR$_{IN}$ & 84 & 76 & 88 & 70 & 77 & 83 & 74 & 84 & 76 & 88 & 70 & 77 & 83 & 74 \\
			CMM$_{IN}$ & \bf50 & \bf37 & \bf37 & \bf38 & - & - & - & \bf50 & \bf37 & \bf67 & 46 & - & - & - \\
			Fz-filter$_{IN}$ & - & - & - & - & - & - & - & - & - & - & - & - & - & - \\
			\hline 
			PCSD-SAT$_{OUT}$ & 68 & - & - & - & - & - & -  & - & - & - & - & - & - & - \\
			PCSD-JIGM$_{OUT}$ & 47 & \bf39 & 47 & \bf46 & 48 & \underline{51} & \bf30 & 67 & \bf60 & \underline{47} & \bf46 & \bf55 & \bf51 & \bf50 \\
			PCSD-JIKR$_{OUT}$ & \underline{45} & - & - & 53 & - & - & -  & \bf45 & - & - & \underline{53} & - & - & - \\
			GM$_{OUT}$ & 65 &\ \underline{53} & 65 & 55 & 48 & 55 & 67 & 70 & \underline{75} & 65 & 62 & \underline{69} & \underline{72} & \underline{67} \\
			KR$_{OUT}$ & 57 & - & \underline{43} & 55 & 48 & \bf33 & \underline{52} & 57 & - & \bf43 & 55 & - & 83 & - \\
			CMM$_{OUT}$ & \bf40 & 55 & \bf36 & \underline{48} & - & - & - & \underline{50} & 82 & 50 & 60 & - & - & - \\
			Fz-filter$_{OUT}$ & - & - & - & - & - & - & - & - & - & - & - & - & - & - \\
		\end{tabular}    
		\captionsetup{justification=centering}
		\caption{Average detection delay at a maximum MAR of 5\%, and a maximum FAR of 20\% for spatially clustered faults SCF-IN and SCF-OUT in the Oklahoma network. The columns show the cluster size and the rows are the QC methods. The boldface and underline highlight the best and the second best performing QC methods.}
		\label{tab:ch2_ok-add_vs_far-scf}
	\end{table}
	
	\begin{table}
		\centering    
		\begin{tabular}{c|cccc|cccc|cccc}   
			& \multicolumn{4}{c|}{FAR $\leq 20\%$} & \multicolumn{4}{c|}{FAR $\leq 10\%$} & \multicolumn{4}{c}{FAR $\leq 5\%$} \\
			Method & 2 & 3 & 4 & 5 & 2 & 3 & 4 & 5 & 2 & 3 & 4 & 5 \\
			\hline                  
			PCSD-SAT$_{IN}$ & - & - & - & -  & - & - & - & - & - & - & - & - \\
			PCSD-JIGM$_{IN}$ & \underline{47} & 75 & 75 & 128 & \bf47 & 93 & 75 & - & \bf 47 & \bf 87 & 83 & 140 \\
			PCSD-JIKR$_{IN}$ & 61 & \bf62 & \underline{63} & - & 67 & 88 & \bf63 & 140  & - & - & - & - \\
			GM$_{IN}$ & 81 & 93 & \bf49 & 104 & 137 & 93 & 113 & 161 & - & - & - &  \\
			KR$_{IN}$ & 65 & 65 & 64 & 87 & \underline{65} & \underline{73} & \underline{64} & 102 & 71 & - & 83 & \bf 102 \\
			CMM$_{IN}$ & \bf45 & \underline{63} & - & - & 70 & \bf63 & - & - & - & - & - & - \\
			Fz-filter$_{IN}$ & - & - & - & - & - & - & - & - & - & - &  &  \\
			\hline 
			PCSD-SAT$_{OUT}$ & - & - & - & - & - & - & - & -  & - & - & - & - \\
			PCSD-JIGM$_{OUT}$ & \bf44 & \bf74 & - & - & - & - & - & - & - & - & - & - \\
			PCSD-JIKR$_{OUT}$ & - & - & - & - & - & - & - & - & - & - & - & - \\
			GM$_{OUT}$ & 108 & - & - & - & 108 & - & - & - & - & - & - & - \\
			KR$_{OUT}$ & - & - & - & - & - & - & - & - & - & - & - & - \\
			CMM$_{OUT}$ & \underline{66} & - & - & - & - & - & - & - & - & - & - & - \\
			Fz-filter$_{OUT}$ & - & - & - & - & - & - & - & - & - & - & - & - \\
		\end{tabular}    
		\captionsetup{justification=centering}
		\caption{Average detection delay at a maximum MAR of 5\%, and a maximum FAR of 20\% for spatially clustered faults SCF-IN and SCF-OUT in the TAHMO Kenya network. The columns show the cluster size and the rows are the QC methods. The boldface and underline highlight the best and the second best performing QC methods.}
		\label{tab:ch2_ke-add_vs_far-scf}
	\end{table}

	\section{Conclusions} \label{sec:ch2_conclusion}
	In this paper, we have introduced three joint reasoning quality control (QC) methods for fault detection. We empirically evaluated these methods on two real-world rainfall datasets from the Oklahoma Mesonet and the TAHMO Kenya weather networks and compared them to four baseline methods for detecting blocked rain gauges. We measured the fault detection delay, false alarm rate, and missed alarm rate. Our results indicate that QC methods utilizing joint reasoning---especially, PCSD-JIKR and PCSD-JIGM---are particularly effective at detecting large clusters of simultaneously faulty sensors. Conversely, the CMM method, currently deployed in the TAHMO network, excelled in scenarios involving spatially random faults or small clustered faults. We recommend that network managers integrate CMM, PCSD-JIKR, and PCSD-JIGM into a unified method as follows. If CMM raises an alarm, the unified method should also raise an alarm. In addition, if either of the joint methods detects a cluster of faults, an alarm should be raised for each station in the cluster. 
	
	Finally, we note that the PCSD methods are not specialized to detecting only blocked rain gauges. We hypothesize that they will work well on other geospatial QC problems provided that the data can be transformed so that linear regression, kernel regression, or joint Gaussian models can be applied.

	\phantom{}\clearpage
	\phantom{}\vfill
	\begin{center}
		\heading
		Novel-SS: A Benchmark for Integrated Novelty-Aware Computer Vision Systems
	\end{center}
	\vfill
	Iren\'{e} Tematelewo, Amin Ullah, Alexander Guyer, Li Fuxin, Stefan Lee, and Thomas G. Dietterich
	\vfill\noindent

	\vfill
	\chapter{Novel-SS: A Benchmark for Integrated Novelty-Aware Computer Vision Systems} \label{ch3:novel_ss}

	\section{Abstract}
	Most computer vision systems encounter a wide variety of novel situations after deployment. Researchers have studied many kinds of novelty separately including changes in object appearance and style, changes in class distribution, novel object classes, novel activities, and novel imaging conditions. To detect and respond to these novelties, various techniques have been developed, each with separate benchmarks. This paper argues that it is time for the field to build integrated systems that can \textit{detect} and \textit{accommodate} many types of novelty, not just one type. To promote this effort, we introduce Novel-Snapshot Serengeti (Novel-SS), a benchmark derived from the Snapshot Serengeti camera trap dataset collected by a network of 225 camera traps deployed in Serengeti National Park (Tanzania). In this application, many forms of novelty arise. In our Novel-SS protocol, the vision system has the option to request annotations for individual images, which allows it to obtain a small amount of supervised data through active learning to support novelty accommodation. The overall goal is to maximize task performance while minimizing the number of requested annotations. This paper describes the training data, performance task, types of novelty, evaluation protocol, and evaluation metrics. To demonstrate the feasibility of the task, we present a novelty-aware system, DCA-SS, and compare it to a baseline system that is completely unaware of novelty.

	\section{Introduction}
	\label{sec:ch3_intro}
	The fundamental assumption of machine learning is that the training data is representative of the post-deployment data. However, this is rarely the case in practice. Instead, deployed systems encounter many different types of novelty including changes in object appearance and style, changes in class distribution, novel object classes, novel activities, and novel imaging conditions. Researchers have studied the problem of detecting and, in some cases, adapting to these kinds of novelty, but each type has been studied separately. 
	Many techniques have been developed including novelty detection \cite{perera2019deep, sabokrou2018adversarially, pimentel2014review}, meta learning \cite{hospedales2021meta, finn2017model}, transfer learning  \cite{zhuang2020comprehensive, tan2018survey, yang2020transfer}, active learning \cite{settles2009active, ren2021survey, sener2017active}, domain adaptation \cite{wang2018deep, csurka2017domain}, and continual learning \cite{hadsell2020embracing, parisi2019continual, li2017learning, Kirkpatrick2017}. Benchmarks corresponding to each of these problems have been developed and well-studied \cite{verwimp2023clad, peng2017visda, koh2021wilds, yang2022openood, song2023comprehensive}. 
	
	However, deployed systems do not encounter just one type of novelty that can be detected and handled by one technique. Instead, these systems must be capable of detecting and responding to all of these different types of novelty. This introduces the need to not only detect novelty but to \textbf{characterize} the type of the novelty so that the appropriate repair strategy can be applied. This paper proposes a new benchmark task that challenges computer vision systems to \textbf{detect} multiple kinds of novelty, \textbf{characterize} the novelties, and apply appropriate strategies to \textbf{accommodate} the novelties (i.e., adapt to them). Integrating these three processes in a single system is a step toward overcoming the well-known brittleness of existing computer vision systems. The benchmark---which we call Novel-SS---is derived from the Snapshot Serengeti camera trap dataset \cite{swanson2015snapshot}, and it exhibits five types of real-world novelty: novel animal species, novel animal behaviors, novel combinations of animal species, novel combinations of animal behaviors, and novel imaging conditions (lighting, weather). These different kinds of novelty pose different detection and accommodation challenges. Detecting novel species and behaviors are instances of open set recognition, and accommodating them requires online continual learning of new classes. Detecting combinations of species and behaviors requires the classification strategy to change from multi-class classification to multi-label classification. Detecting novel imaging conditions involves detecting changes in lower-level image features. Accommodating those novel conditions could require continual learning with additional augmentations as well as continual learning of new styles. In Novel-SS, the system can request image-level annotations (within a budget). Selecting which images to query adds an active learning aspect to the benchmark.

\begin{figure*}
		\centering
		\includegraphics[width=\textwidth]{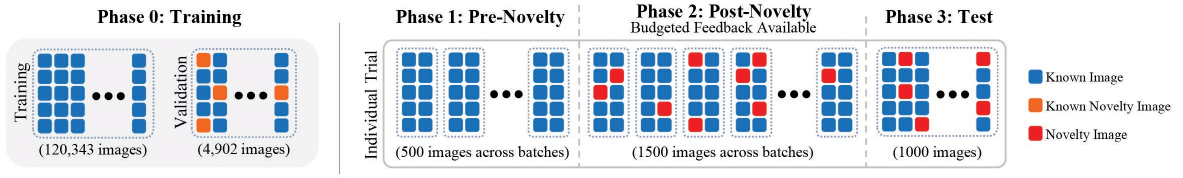}    
		\caption{Schematic diagram of the training and execution phases of the Novel-SS task.}
		\label{fig:phases}
	\end{figure*}

	The protocol for evaluating a novelty-aware system operates in four phases (see Fig.~\ref{fig:phases}). In phase 0, a vision system is trained in the usual way on a large training set that contains no novelty. A small validation dataset that contains some ``known novelties'' is made available as well. (However, these types of known novelties will not appear in phase 3.)
	In phase 1 (``pre-novelty''), the system is presented with a sequence of images. For each image, it must perform various recognition and counting tasks. Like the training images, the phase 1 images do not contain novelty. The purpose of this phase is to measure the performance of the system in the absence of novelty. The learning system is not told where the pre-novelty phase ends and the post-novelty phase begins.
	In phase 2 (``post-novelty''), the system is presented with a sequence of images where an unknown fraction $\alpha$ of the images exhibit novelty. The images are presented in batches of 10. The system computes and outputs its predictions for the recognition and counting tasks. It
	can then choose a subset of $\lfloor \beta \times 10 \rfloor$ images and request the annotations of those images, where $\beta$ is the query budget. This supervised feedback can be applied to update the vision system. Note that $\alpha$ can be zero, which is to say that phase 2 may contain no novelty.
	Finally, in phase 3 (``test''), the system is evaluated, without further feedback or training, on a test set containing a mix of images of which fraction $\alpha$ exhibit novelty.
	
	The evaluation is divided into a set of trials. Each trial consists of new phases 1, 2, and 3, and only one type of novelty occurs. In the NOVEL-SS evaluation protocol, each trial consists of 500 Phase 1 images, 1500 Phase 2 images, and 1000 Phase 3 images. Each trial is independent of the others, so that they can all be run concurrently starting from the same Phase 0 trained system.

	We make the following contributions. The primary contribution is the Novel-SS dataset, task structure, and evaluation metrics (section \ref{sec:ch3_data_task_and_metrics}). A secondary contribution is a baseline system, DCA-SS (Detect, Characterize, and Accommodate), that demonstrates both that existing methods can achieve non-trivial performance on the Novel-SS benchmark and that there is substantial room for improvement (section \ref{sec:ch3_dca_baseline}). Experiment results(section \ref{sec:ch3_experimental_assessment})  show that DCA-SS outperforms a novelty-unaware baseline (NU) for many kinds of novelty. However, the performance improvements achieved by the state-of-the-art continual learning method, EWC, are somewhat disappointing and far too computationally expensive. This shows that developing effective and computationally efficient online continual learning remains a significant open problem.
	
	Instructions for downloading the dataset, as well as code for the evaluation harness and DCA-SS are available at \url{https://github.com/guyera/NOVEL-SS-and-DCA-repos}.

	\section{Novel-SS Dataset, Task Structure, and Evaluation Metrics}
	\label{sec:ch3_data_task_and_metrics}
	
	To design a dataset for integrated novelty-aware computer vision, we sought a real-world application rather than working with collections of arbitrary images such as ImageNet \cite{deng2009imagenet} or CIFAR-100 \cite{cifar}. Camera traps are revolutionizing biological conservation and field ecology. A camera trap is triggered by a motion detector and then collects one or more images. Over time, the traps collect millions of images, which must then be analyzed to estimate the presence of each species, the presence of various activities, and, ideally, the number of animals of each species in the image. Species presence/absence is important for modeling the habitat requirements of different species \cite{Dorazio2005} and their migration dynamics \cite{SheldonEK07,SheldonD11}. Species co-occurrence helps ecologists understand species-species interactions \cite{Groom2021,Strydom2021}. Activities, and their co-occurrence, are important for understanding the degree to which a species is under stress (e.g., from food scarcity or predation). Animal counts provide a valuable input to estimating species abundance \cite{Iijima2020}, which is critical for monitoring endangered species. Camera traps also provide an ideal computer vision challenge, because they encounter many natural types of novelty including novel species, activities, and imaging conditions.
	
	One of the largest datasets of labeled camera trap images is Snapshot Serengeti (SS).  SS contains over 7 million camera trap images from 225 sites in Serengeti National Park, Tanzania \cite{swanson2015snapshot}. Each time the motion detector is triggered, the camera captures an \textbf{image sequence} of  three images, each separated by one second.  Each image sequence is annotated by several volunteers. For each species present in the image, a \textbf{count annotation} gives the minimum, maximum, and median counts assigned by the volunteers. The counts take values in the set $\{1, 2, 3, 4, 5, 6, 7, 8, 9, 10, 11\mbox{–}50, 51+\}$. The \textbf{activity annotation} is given as the fraction of volunteers who judged that at least one individual from the given species is engaged in the activity. The activities are \textit{standing, moving, interacting, eating} and \textit{resting}. 
	
	\begin{figure*}[t]
		\centering
		\includegraphics[width=\textwidth]{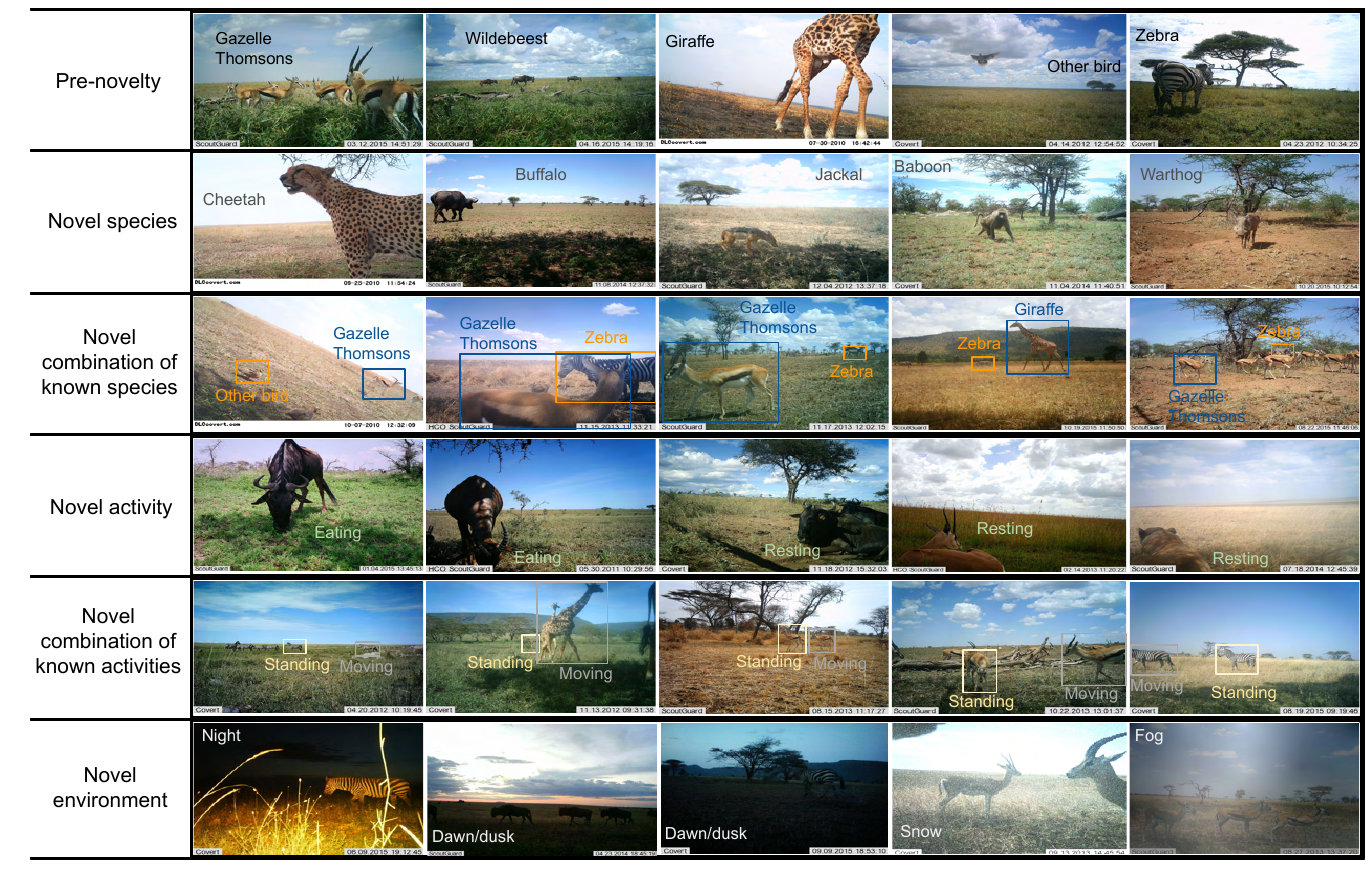}
		
		\caption{Example no-novelty and novelty images from the Novel-SS dataset.}
		\label{fig:sample_images_from_novel_ss}
	\end{figure*}
	
	\subsection{The Novel-SS Dataset} \label{sec:ch3_dataset_Structure}
	Our Novel-SS dataset is a collection of 163,560 images extracted from the Snapshot Serengeti (SS) camera trap dataset. It is composed of two groups: non-novel images and novel images. All non-novel images contain one or more animals from a single known species engaged in a single known activity, and they are captured only during daylight hours (defined as the period from one hour after sunrise to one hour before sunset).  We refer to daytime as the ``known environment'', all non-novel species as the ``known species'', and  all non-novel activities as the ``known activities''.  
	
	The properties of the non-novel images imply that the image-level annotations can be ``pushed down'' to the individual bounding boxes for each detected animal, and classifiers for species and activities can be trained on these bounding boxes using standard multiclass supervised learning. To make image-level predictions, the predictions of the individual bounding boxes can be voted to determine the most likely single species and activity present in the image. 
	
	\textbf{Novelty types}. The Novel-SS novelties are divided into five types: (a) \textbf{\textit{Novel species}}: images captured under the same conditions as the no-novelty data, but containing at least one novel species and with all animals engaged in a single known activity. These images contain at most one known species. (b) \textbf{\textit{Novel combination of known species}}: images captured under the same lighting conditions as the no-novelty data, but containing two or more known species and all animals engaged in the same known activity. This kind of novelty is of great interest to ecologists, because it generally reveals which species can comfortably co-occur. (c) \textbf{\textit{Novel activity}}: images captured under the same conditions as the no-novelty data, but containing a single known species with at least one animal engaged in a novel activity. All animals not engaged in a novel activity are engaged in the same known activity. More than one novel activity may be present. (d) \textbf{\textit{Novel combination of activities}}: images captured under the same conditions as the no-novelty data, but containing animals from a single known species engaged in the two or more known activities. (e) \textbf{\textit{Novel environment}}: images captured under a different imaging condition (or environment) than no-novelty images and containing a single known species with all animals engaged in the same known activity. Note that the first four novelty types require a change in the prediction strategy, because it is no longer possible to vote the species and activity predictions of the individual bounding boxes. 
	
	\begin{table}
		\centering
		\def\arraystretch{1.5}
		\caption{Known and novel species, activities and environments.}
		\label{tab:species_n_activities}
		\begin{tabularx}{\columnwidth}{|l|X|}
			\hline
			{\centering} \textbf{Novelty types} & {\centering} \textbf{Categories} \\
			\hline

			{\centering} Known species & \textit{Elephant, Gazelle grants, Gazelle Thomsons, Giraffe, Hyena Spotted, Lion female, Other bird, Topi, Wildebeest, Zebra} \\ 
			\hline
			{\centering} Novel species & \textit{Baboon, Buffalo, Cheetah, Dikdik, Eland, Guineafowl, Hartebeest, Impala, Jackal, Koribustard, Mongoose, Ostrich, Reedbuck, Warthog, InsectSpider} \\
			\hline
			{\centering} Known activities & \textit{moving, standing} \\ 
			\hline
			{\centering} Novel activities & \textit{eating, resting} \\
			\hline
			{\centering} Known envs & \textit{daytime} \\
			\hline
			{\centering} Novel envs & \textit{dawn/dusk, night time, fog, snow} \\ 
			\hline
		\end{tabularx}
	\end{table}
	
	\noindent
	\textbf{Image selection}. To build Novel-SS, we partition the species, activities and environments as shown in Table \ref{tab:species_n_activities}. The species were split into known and novel subsets such that each set contains species that are grazers, predators, and scavengers. Each subset exhibits class imbalance, because grazers are more common than scavengers, and predators are very rare. After this split, all SS image sequences satisfying the criteria of either the no-novelty data or any of the five novelty types are selected.  We define four novel environments. Dawn/dusk images are captured during the period from 30 minutes before to 30 minutes after sunrise or sunset. Night images are captured during the period from 1 hour after sunset to 1 hour before sunrise. The snow and fog images are constructed by randomly selecting some no-novelty images, removing them from the no-novelty set, and transforming them using image corruption techniques \cite{hendrycks2019benchmarking}. 
	
	\noindent\textbf{Image filtering}. The crowd-sourced annotations are provided only at the level of each sequence of 3 images. As our goal is to treat each image separately and ensure that its annotations are sensible, we apply a series of filters and annotation adjustments, as described in Appendix~\ref{app:ch3_dataset}. 
	
	The resulting Novel-SS dataset contains 120,343 training, 4,902 validation, and 38,315 testing images. All training images are from the no-novelty set. To construct a validation set, a few of the potential novelties are declared to be ``known novelties'' (see Table~\ref{tab:summary_train_val_test}). Images containing these novelties are removed from the test set and placed in the validation set. The test set contains 9,420 no-novelty images and 28,895 novel images. Because images are captured in sequences, eligible images from the same sequence are all put in the same set (i.e., training, validation, or test). Appendix~\ref{app:ch3_dataset} provides detail on the distributions of species, activities, and co-occurrence in Novel-SS.
	
	\subsection{Task Structure}
	For each image in Novel-SS we provide a list of bounding boxes specifying the location in the image of each detected animal. These are computed using the excellent Megadetector \cite{beery2019efficient,microsoft2020}. Let $K_s$ and $K_a$ be the number of known species and known activities in the no-novelty set, and $M_s = 20$ and $M_a = 5$ be upper bounds on the number of novel species and novel activities in the novelty set. We convert the crowd-sourced annotations into three vectors: (i) a $K_s+M_s$-dimensional vector $N_i=[n_{i1},\ldots,n_{iK_s},n_{iK_s+1},\ldots,n_{iK_s+M_s}]$ of counts per species; (ii) a $K_s+M_s$-dimensional Boolean vector $U_i$ of species presence, and (iii) a $K_a+M_a$-dimensional Boolean vector $V_i$ of activity presence. The task is to predict these three vectors for each image. We denote the predictions by $\hat{N}_i$, $\hat{U}_i$, and $\hat{V}_i$. The learning system can request feedback on image $i$, which is provided as ($N_i$, $U_i$,  $V_i$). 
	
	To evaluate performance on each novelty type, Novel-SS provides 9 trials. One trial contains no novelty, and there is one trial each for novel species, novel activities, novel species combinations, and novel activity combinations. Finally, there are four trials for novel environments (night, dawn/dusk, snow, fog). The vision system does not know which type of trial it is processing, so it must be able to handle all novelty types.
	
	\subsection{Evaluation Metrics} \label{ch3:evaluation_metrics}
	We measure several metrics on each trial during the post-novelty and test phases. 
	
	\textbf{Species and activity presence.} For each species and activity, we compute the $F_1$ score and then average those scores across all of the species (or activities, respectively). We adopted $F_1$ rather than the area under the ROC curve (AUROC) because the average AUROC does not correspond to a coherent threshold setting across all classes and, hence, exaggerates performance. We also discovered that the AUROC value was influenced by the values of $M_s$ and $M_a$. The mean F1 for species and activity presence is computed as 
	\begin{equation}
		\label{eq:f1_score}
		\bar{F_1} = \frac{1}{|Q|} \sum_{j=1}^{|Q|} \Big( 2 \frac{\text{precision}_j \cdot \text{recall}_j}{\text{precision}_j + \text{recall}_j} \Big),
	\end{equation}
	where $\text{precision}_j$ and $\text{recall}_j$ are the precision and recall of species (or activity) $j$, $Q$ is the set of known and novel species (or activities) present in the trial, and $|Q|$ is the cardinality of $Q$.

	\textbf{Species counts.} For scoring species counts on a set of $N$ images, we compute the average across the species of the mean absolute error between the predicted number and the true number of animals as follows. Let $i$ index the images, $k$ index the species, $I_k = \sum_i \mathbb{I}[n_{ik}>0]$ be the number of images containing species $k$, $N_k=\sum_{i=1}^N n_{ik}$ be the number of ground truth animals in species $k$ summed across all images, and $\hat{N}_k =\sum_{i=1}^N \hat{n}_{ik}$ be the number of predicted animals in species $k$ summed across all images. The mean absolute error is defined as
	\begin{equation}
		\label{eq:mean_absolute_error}
		MAE = \frac{1}{|Q|} \sum_{k=1}^{|Q|} \Big( \frac{1}{I_k} \sum_{i=1}^{N} |n_{ik} - \hat{n}_{ik}| \Big).
	\end{equation}

	\textbf{Feedback queries.} We measure the number of feedback queries made by the classifier. We also report the area under the ROC curve for the novelty detectors for species, activities, and environments to separately assess novelty detection performance.
	
	In order to perform well on the evaluation metrics, a system needs to detect images containing novel species and novel activities, request annotations for those images, and learn to recognize them in new images. To achieve high average F1 scores, it must do well on both rare and common species---hence, it must handle class imbalance well. Finally, the system must learn to handle the novel environments while recognizing the known species and activities.

	\section{Baseline Solution: DCA-SS (Detect, Characterize, and Accommodate)} \label{sec:ch3_dca_baseline}
	
	\begin{figure}
		\centering
		\includegraphics[width=\textwidth]{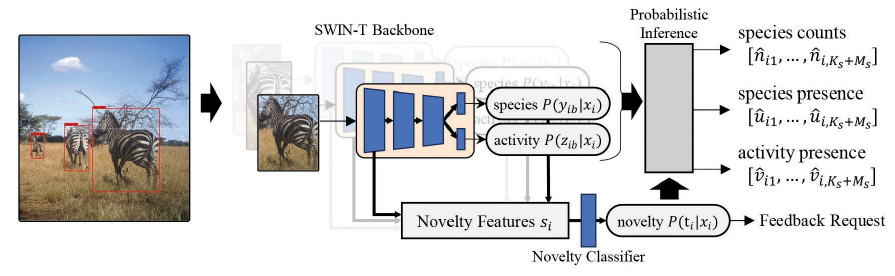}
		\caption{Architecture of the DCA-SS system. Initial training is not shown. The network iterates over each bounding box in the image. The resulting predictions for species and activity are aggregated by probabilistic inference to produce predictions for the species counts, species presence, and activity presence tasks.}
		\label{fig:architecture}
		\vskip -0.15in
	\end{figure}
	
	We developed a baseline solution, DCA-SS, for Novel-SS. Our goal is to show that the task is feasible. We describe our solution in terms of three subcomponents: detect, characterize, and accommodate. The architecture of DCA-SS is depicted in Fig.~\ref{fig:architecture}. The heart of the system is a Swin transformer backbone with two classification heads, one for species and one for activities. It is trained on the training images and then calibrated with temperature scaling \cite{Guo2017} on the validation images. Training and calibration requires 2.5 days on a 4-GPU NVidia dgx2. This network is applied to each bounding box $b=1,\ldots,B_i$ in the image $x_i$. The resulting probabilities $\{P(y_{ib} \mid x_i) \text{ and } P(z_{i}b \mid x_i) : b=1,\ldots,B_i\}$, along with the predicted novelty type, are combined through probabilistic reasoning to compute the predicted outputs $N_i$, $U_i$, and $V_i$. We now discuss the novelty detection, characterization, and accommodation methods.
	
	\subsection{Detect and Characterize}

	The first step of our solution is to predict what type of novelty, if any, each image contains. This formulation provides novelty detection and characterization simultaneously. Our strategy is to extract a set of novelty signals (Mean of Max Logit, Max of Mean Logit, and Image Density Estimation) from the two-headed network trained on the no-novelty data (details below). A multinomial logistic regression model (the ``novelty classifier'' in Fig.~\ref{fig:architecture}) is then trained to predict the novelty type using the known novelties from the validation set.

	\noindent\textbf{Mean of Max Logit.}
	Let $\ell_k^c(i,b)$ be the logit score for class $k$ of classifier $c$ on bounding box $b$ of image $i$. Vaze \etal~\cite{vaze2022openset}, among others, has shown that the (negative of the) maximum of the softmax probabilities is a state-of-the-art novelty score. Dietterich and Guyer~\cite{Dietterich2022} show that the logits can be viewed as measures of object familiarity, so unfamiliar objects produce small logit scores. We define $s_{sam}(i)$ to be the average of the species max logit scores, and $s_{aam}(i)$ to be the mean of the activity max logit scores for image $i$ with $B_i$ bounding boxes:
	\begin{equation*}
		s_{sam}(i) = \frac{1}{B_i} \sum_{b=1}^{B_i} \max_{k \in \{1,\ldots,K_s\}} \ell_k^s(i,b); \quad \quad
		s_{aam}(i) = \frac{1}{B_i} \sum_{b=1}^{B_i} \max_{k \in \{1,\ldots,K_a\}} \ell_k^a(i,b).
	\end{equation*}
	This signal is useful for images containing only a single species or a single activity. If either the species or the activity is novel, then every max logit will be small, so their mean will be small.
	
	\noindent\textbf{Max of Mean Logit.} A second novelty signal is the maximum of the mean logit scores. We define $s_{sma}$ and $s_{ama}$ analogously:
	\begin{equation*}
		s_{sma}(i) = \max_{k \in \{1,\ldots,K_s\}} \frac{1}{B_i} \sum_{b=1}^{B_i} \ell_k^s(i,b); \quad \quad
		s_{aam}(i) = \max_{k \in \{1,\ldots,K_a\}} \frac{1}{B_i} \sum_{b=1}^{B_i} \ell_k^a(i,b).
	\end{equation*}
	This is intended to capture cases where there are multiple combinations of known species or known activities. In such cases, the logits for known class $k$ will vary across the bounding boxes with some high and some low. Hence, the mean will be intermediate, and the max of these mean logit scores will not be large.

	\noindent\textbf{Image Density Estimator.} 
	To extract a useful signal for detecting novel environments, we estimate $-\log P(x_i)$, the probability density of image $x_i$ computed with respect to the distribution of the training images. We fit the density estimator by first applying PCA to the training set activations of an early layer in the network. We keep the top $K_{PCA}=64$ PCA dimensions capturing the highest variance and discard the rest. Then we fit a Gaussian kernel density estimator $P(x_i)$ with bandwidth 50 in this reduced PCA space to the activations of the non-novel validation set images. Let $s_{env}(i) = -\log P(x_i)$. The values of $K_{PCA}$ and bandwidth were selected using the validation data.
	
	\textbf{Training the Novelty Type Classifier.}
	In Novel-SS, each image belongs to exactly one of six novelty types: no-novelty, novel species, novel combination of known species, novel activity, novel combination of known activities, novel environment. Let $t = 1, \ldots, T$ index these, and let $t_i$ denote the index of the novelty type for image $i$. We train a multinomial logistic regression model to take the vector of novelty detection signals $s_i$ and predict the probability $P(t_i  \mid s_i)$. This logistic regression is fit to a held-out subset of the (no-novelty) training data and the validation set of ``known novel'' images. After training this classifier, we estimate the probability that an image is novel as $s_{nov}(i) = \sum_{t=2}^T P(t_i = t \mid s_i) = 1 - P(t_i=1)$. Note that feedback during the post-novelty phase can eliminate some novelty types. We incorporate this by removing those novelty types and renormalizing $P(t \mid x_i)$ over the remaining novelty types.
	
	\textbf{Probabilistic Inference to Compute Counts and Presence.} For each image $i$, once DCA-SS has predicted the probabilities for the species and activity of each bounding box, along with the probabilities of each possible novelty type for the image, it applies probabilistic inference to compute the image-level predictions for species counts ($\hat{N}_i$), species presence ($\hat{U}_i$), and activity presence ($\hat{V}_i$).  Because the bounding box predictions output by the classifier are computed separately from the novelty type prediction $P(t_i  \mid  s_i)$, they may be inconsistent. For example, the classifier may predict that a bounding box belongs to a novel species class with high probability even though the multinomial logistic regression predicts with high probability that the image contains no novelty. To reconcile these predictions, we assume that the multinomial logistic regression is well-calibrated, and we transform and aggregate the bounding box predictions in a manner that respects the outputs of the logistic regression. For image $i$, let $n_{ik}$ denote the count of species $k$, let $u_{ik}$ denote the binary presence label (0 for absent, 1 for present) of species $k$, and let $v_{ik}$ denote the binary presence label for activity $k$. The expected count of species $k$ in image $i$ is predicted as
	\begin{equation}
		\label{eq:count_prediction}
		\hat{n}_{ik} := \sum_{t=1}^T \mathbb{E}[n_{ik} \mid t_i = t] P(t_i = t\mid s_i).
	\end{equation}
	\noindent Similarly, the probabilities of species and activity presence are predicted as 
	\begin{equation}
		\label{eq:presence_prediction}
		\begin{aligned}
			P(\hat{u}_{ik}=1 \mid s_i) &= \sum_{t=1}^T P(u_{ik}=1 \mid t_i=t)P(t_i=t \mid s_i),\\
			P(\hat{v}_{ik}=1 \mid s_i) &= \sum_{t=1}^T P(v_{ik}=1 \mid t_i=t)P(t_i=t \mid s_i).
		\end{aligned}
	\end{equation}
	The quantities $\mathbb{E}[n_{ik} \mid t_i = t]$, $P(u_{ik}=1 \mid t_i=t)$, and $P(v_{ik}=1 \mid t_i=t)$ are predicted separately for each novelty type $t$ by conditioning the predictions of the bounding box labels on $t$ and aggregating them under a conditional independence assumption. The details are given in Appendix~\ref{app:DCA-details}. 
	
	Recall that the overall objective of a solution to Novel-SS is to maximize prediction accuracy while minimizing the number of feedback queries. To avoid making queries during the pre-novelty period, we apply a change point detection method to determine when to start making queries. Specifically, as DCA-SS processes the images in a trial, it conducts a Kolmogorov-Smirnov test (KS test, \cite{smirnov1939estimate}) to compare the distribution of the $s_{nov}$ scores of the first 300 images in the trial to the $s_{nov}$ scores of the most recent $N_W = 40$ images. If the $p$-value of this test drops below a specified threshold $\gamma$, DCA-SS raises the ``novelty alarm'' and begins requesting feedback and accommodating novelty. This strategy assumes that there is no novelty in the first 300 images. Similar assumptions are usually safe to make in real applications.  (Alternatively, one could use held out non-novel images.) To choose $\gamma$, we set $N_W=40$, simulate $1{,}000$ non-novel trials constructed from held-out training data, and set $\gamma$ to achieve a false alarm rate of 10\%. Note that a larger value of $N_W$ decreases the variance of the KS test's $p$-value but increases the system's novelty detection delay.

	\subsection{Accommodate}
	
	Once the novelty alarm is raised, DCA-SS begins requesting feedback. Recall that the system is permitted to request feedback on fraction $\beta$ of the 10 images in each batch. DCA-SS sorts the images in descending order of $s_{nov}$ and selects the top $\lfloor \beta \times 10 \rfloor$ sorted images. The goal is to request feedback on the images that are most likely to contain novelty. In all of our experiments, $\beta=0.5$.
	
	Ideally, we would update the species and activity classifiers after each batch of feedback is received. However, such online learning is very expensive, so we opted to collect all feedback information and apply the Elastic Weight Consolidation (EWC) continual learning method \cite{Kirkpatrick2017} at the end of the post-novelty phase. EWC constrains weight updates to stay close to an ellipsoidal region centered on the original trained weights. It has a hyperparameter that controls the trade-off between fitting the new (feedback) data versus retaining performance on the original training data. This was set to $10{,}000$ based on experiments with the validation data. We allow EWC to update all of the weights in the network.
	
	As we have seen, the feedback information is provided at the image level. If an image contains multiple species or multiple activities, this creates a multiple instance problem \cite{dllp-smipapr-97}. To bridge the gap between the bounding box classifiers and the image-level annotations, we compute the negative log likelihood that the image-level feedback labels are satisfied by the box-level predicted probabilities, and rely on auto-differentiation to implement backpropagation through this computation. A naive computation of the negative log likelihood requires considering all possible assignments of species and activity labels to the bounding boxes in the image. Fortunately, Shukla \etal~\cite{mi_dyn_programs} provide an efficient dynamic program for the negative log likelihood for two-class classifiers, and we extended their method to the multiclass case. (See Appendix~\ref{app:DCA-details} for details.)
	
	Updating the network with EWC causes the novelty detection features ($s_i$) to change. Hence, the novelty type classifier must be retrained each time EWC is run.

	\subsubsection{Formulating image-level predictions from box-level predictions}
	
	Given the box-level predictions for species $P(y_{ib} \mid x_i)$ and activities $P(z_{ib} \mid x_i)$ (for each bounding box $b = 1,\ldots, B_i$ in image $i$) and the probability distribution over the novelty type $t$ of image $x_i$, $P(t \mid x_i)$, from the multinomial logistic regression, we want to compute three image-level values:
	\begin{itemize}
		\item The expected number of animals belonging to each species $k$: 
		\[\hat{n}_{ik} = \mathbb{E}\left[\sum_{b=1}^{B_i} P(y_{ib}=k \mid x_i)\right].\]
		\item The probability that species $k$ is present in the image:
		\[1 - \prod_{b=1}^{B_i} P(y_{ib} \neq k \mid x_i).\]
		\item The probability that activity $a$ is present in the image:
		\[1 - \prod_{b=1}^{B_i} P(z_{ib} \neq k \mid x_i).\]
	\end{itemize}
	We do this by first conditioning the bounding box predictions on the novelty type $t$, $P(y_{ib} = k  \mid  x_i, t)$ and $P(z_{ib} = k  \mid  x_i, t)$, and then marginalizing out $t$ based on $P(t \mid x_i)$ as in equation (\ref{eq:count_prediction}).

	\section{Experimental Assessment of Feasibility}  \label{sec:ch3_experimental_assessment}
	
	A good dataset challenge should be scientifically interesting and feasible. We assess the feasibility of Novel-SS through the following experiment. 
	
	\subsection{Experiment Design}
	We evaluate the performance of DCA-SS on the 8 novelty trials of Novel-SS and compare it to a novelty-unaware baseline. Let $f_{pre}$ be the two-headed neural network trained on the training data. We employ a SWIN-T architecture \cite{swin} with one head for species and the other for activities. Each head consists of two fully-connected layers. The first layer contains 256 units, and the second layer contains $N_s + M_s$ units for the species head and $N_a + M_a$ units for the activity head. 
	Hence, given bounding box $x$, $f_{pre}(x)=[P(y \mid x),P(z \mid x)]$, where $y \in \{1,\ldots,K_s\}$ ranges over the species, and $z \in \{1,\ldots,K_a\}$ ranges over the activities. 
	As a basis for comparison, NU (Novelty Unaware) is a version of DCA-SS that performs no novelty detection, characterization, or accommodation. It applies $f_{pre}$ to the task without any changes during the trial. It makes no feedback queries.
	
	\begin{table}[t]
		\centering 
		\caption{Animal counts and species presence $F_1$ during the test phase. The best value for each trial/metric is in boldface.}
		\label{tab:species-test-results}
		{
			\begin{tabular}{@{}lcm{0.29in}m{0.29in}m{0.29in}m{0.29in}m{0.06in}m{0.29in}m{0.29in}m{0.29in}m{0.29in}@{}} \toprule
				&\multicolumn{1}{l}{\relax}&\multicolumn{4}{c}{Animal Counts\ \ }&&\multicolumn{4}{c}{Species Presence\ \ }\\
				&  DCA   &\multicolumn{4}{c}{(MAE $\downarrow$)}&&\multicolumn{4}{c}{($F_1$ $\uparrow$)}\\
				\cmidrule{3-6}\cmidrule{8-11}
				&Feedback&\multicolumn{2}{c}{Known}&\multicolumn{2}{c}{Novel}&&\multicolumn{2}{c}{Known}&\multicolumn{2}{c}{Novel}\\
				\cmidrule{3-4}\cmidrule{5-6}\cmidrule{8-9}\cmidrule{10-11}
				Trial Type       &Queries&{NU}&{DCA}&{NU}&DCA&&{NU}&{DCA}&{NU}&{DCA}\\
				\midrule
				No Novelty             & 570     & {\bf 0.28} &      0.34 &     \ \ --&     \ \ -- & &      0.83 & {\bf 0.84}&     \ \ --&  \ \ --   \\
				Novel Species          & 720     & {\bf 0.30} &      0.40 &      2.58 & {\bf 2.56} & &      0.84 & {\bf 0.88}&      0.05 & {\bf 0.07}\\
				Species Combination    & 735     & {\bf 0.28} &      0.44 &      7.64 & {\bf 2.45} & & {\bf 0.85}&      0.82 &      0.46 & {\bf 0.60}\\
				Dawn/Dusk              & 735     & {\bf 0.34} & {\bf 0.34}& {\bf 1.30}& {\bf 1.30} & & {\bf 0.84}& {\bf 0.84}&      0.47 & {\bf 0.49}\\
				Night                  & 740     & {\bf 0.34} &      0.40 &      1.52 & {\bf 1.42} & & {\bf 0.81}& {\bf 0.81}&      0.29 & {\bf 0.32}\\
				Snow                   & 735     & {\bf 0.30} &      0.34 & {\bf 1.12}&      1.16  & &      0.84 & {\bf 0.85}& {\bf 0.74}& {\bf 0.74}\\
				Fog                    & 745     & {\bf 0.32} &      0.38 & {\bf 0.86}&      0.90  & & {\bf 0.83}&      0.82 &      0.77 & {\bf 0.78}\\
				\bottomrule
			\end{tabular}
		}
		\vskip -0.1in
	\end{table}
	
	\subsection{Results}
	Table~\ref{tab:species-test-results} compares the performance of NU and DCA-SS on the species count and species presence metrics during the test phase of each trial, and Table~\ref{tab:activity-test-results} reports the results for activities. Each pair of columns shows the performance of NU and DCA-SS on the relevant metric. Note that when new classes are added to a classification problem, the problem becomes more challenging, and accuracy drops. Hence, even if a metric is unchanged in a novelty trial, this is a positive result. A second subtlety is that the test images for each novelty trial are different, because of the constraints imposed on the data. For example, the only trials that involve multiple species within a single image are Novel Species and Species Combination. The MAE values for those trials are much larger than in other trials. This shows that comparisons across rows in the tables are not meaningful.
	
	\begin{table}[t]
		\centering 
		\caption{Activity presence $F_1$ during the test phase. The best value for each trial/metric is in boldface.}
		\label{tab:activity-test-results}
		{
			\begin{tabular}{@{}lcm{0.29in}m{0.29in}m{0.29in}m{0.29in}m{0.06in}@{}} \toprule
				&DCA&\multicolumn{4}{c}{Activity Presence ($F_1$ $\uparrow$)}\\
				\cmidrule{3-6}
				&Feedback&\multicolumn{2}{c}{Known}&\multicolumn{2}{c}{Novel}\\
				\cmidrule{3-4}\cmidrule{5-6}
				Trial Type          & Queries   &{NU}  &{DCA}       &{NU}   &DCA        \\
				\midrule
				No Novelty          &  570   & {\bf 0.95} &      0.88  &     \ \ --& \ \ --     \\
				Novel Activities    &  735   & {\bf 0.87} &      0.86  &      0.17 & {\bf 0.19} \\
				Activity Combination&\ \ \ 0 & {\bf 0.86} & {\bf 0.86} &      0.68 & {\bf 0.70} \\
				Dawn/Dusk           &  735   & {\bf 0.85} & {\bf 0.85} &      0.67 & {\bf 0.69} \\
				Night               &  740   & {\bf 0.88} & {\bf 0.88} &      0.57 & {\bf 0.60} \\
				Snow                &  735   & {\bf 0.88} & {\bf 0.88} &      0.76 & {\bf 0.78} \\
				Fog                 &  745   & {\bf 0.88} & {\bf 0.88} & {\bf 0.82}&      0.81  \\
				\bottomrule
			\end{tabular}
		}
	\end{table}
	
	There are two things to immediately note. First, the NU system gives excellent scores when there is no novelty---it is a good closed-world classifier. Second, DCA-SS incorrectly detects novelty in the No Novelty trial (and makes 570 feedback queries as a result), and it fails to detect novelty in the Activity Combinations trial (and makes 0 feedback queries as a result). In all of the other trials, the KS-test novelty detection was excellent (the number of feedback queries is close to the best possible value of 750). To assess the impact of these novelty detection errors, we ran a version of DCA-SS with perfect novelty detection. The results on Activity Combination $F_1$ are 0.84 (known) and 0.74 (novel) based on 750 feedback queries. The AUROC for DCA-SS detecting novel images during the post-novelty period are as follows: 0.83 (novel species), 0.71 (novel activities), 0.96 (species combinations), 0.00 (activity combinations), 0.84 (dawn/dusk), 1.00 (night), 0.74 (snow), and 0.56 (fog). 
	
	The need for novelty-aware computer vision is clearly demonstrated by the drop in performance of NU on the novel images in all trials, where it makes errors on all novel images. However, even though DCA-SS is novelty-aware, it still struggles to beat the NU system. For Animal Counts, it beats NU on novel images in two trials, but in both of those trials, this comes at the cost of slight drops in performance on known images, and the DCA-SS performance is still very poor. For Snow and Fog, NU is better on both known and novel images. For Species Presence, the story is better: DCA-SS is able to match or exceed NU in 5 trials on known images and in all trials on novel images. Finally, for Activity Presence, DCA-SS is able to match or exceed NU on both known and novel images in five trials, and it is only slightly worse on one or the other in two trials. NU is never better on both known and novel images. 
	
	\subsection{Lessons Learned}
	The failures of the KS-test could be addressed by combining the test with a small number of feedback queries. The queries could confirm or contradict the findings of the KS-test and prevent these failures. The poor F1 performance of DCA-SS on recognizing novel species and activities is caused by a total failure to recognize rare novel species and activities. We employed a loss-reweighting strategy in EWC to give more weight to the losses of rare categories, but this was not sufficient. Better methods are needed for handling continual learning of rare categories.

	\section{Limitations}\label{app:ch3_limitations}
	
	The Novel-SS dataset and task includes many forms of novelty (novel species, novel activities, novel environments) that pose major challenges for existing computer vision and machine learning methods. However, Novel-SS does not evaluate camera traps at novel locations, especially ones that might observe a different distribution of species. Each trial in Novel-SS also only exercises a single novelty type. Because novelty is relatively rare in real applications (at least those where a substantial training dataset has been collected), this is not an unrealistic assumption. But it would be nice to create trials that simultaneously exhibited multiple types of novelty. We anticipate releasing a second version of Novel-SS that can address both of these novelty configurations. 
	
	A methodological weakness of the validation set is that it contains validation data for three of the four novel environments that will be encountered in the trials. This did not have any detectable effect on the performance of DCA-SS. In fact, DCA-SS gives improvements on the Night environment, which was \textit{not} included in the validation set. Nonetheless, in the second version of Novel-SS, we will replace the validation novel environments with other synthetic novel environments.
	
	Novel-SS also does not test the ability of object detectors to detect novel classes of objects. Addressing this would involve requiring any proposed solution to also learn to detect novel species. In our data release, we do provide the original images, so researchers can explore this task as well, but the evaluation scripts would need to be extended to score the detection process.
	
	The paper did not explore an additional benefit of Novel-SS, namely that it can be applied to evaluate and compare individual methods for novelty detection, domain adaptation, and continual learning. It would be worthwhile to write additional evaluation scripts for this purpose.

	\section{Conclusion}
	
	This paper has argued that research on novelty-aware computer vision would benefit from a benchmark that requires detecting, characterizing, and accommodating multiple forms of novelty. The benchmark should stress online, continual learning based on budgeted annotation feedback. The paper has presented the Novel-SS benchmark dataset, evaluation protocol, and metrics, which meets these requirements. Our baseline system, DCA-SS, demonstrates that current methods in computer vision are able to achieve some improvements in performance compared to a novelty-unaware system. But a strong solution to Novel-SS will require integrating novelty detection, active learning, and online continual learning for novel categories and novel imaging conditions. We encourage researchers to tackle these challenges and measure their progress using Novel-SS.

	\section{Acknowledgements} \label{sec:ch3_acknowledgements}
	We would like to acknowledge Alexander Guyer for his contribution to extending the dynamic program by Shukla \etal~\cite{mi_dyn_programs} to the multicalss setting for the proposed DCA baseline. We would also like to acknowledge Dr. Amin Ullah for his implementation of DCA's novelty accommodation strategy.

	\chapter{Conclusion}
	
	\clearpage
	\section{Conclusion}
	In this thesis we discussed the problem of anomaly detection in weather sensor networks and novelty detection computer vision. For anomaly detection in weather sensor network, given the crucial role of rainfall prediction for food security, disaster prevention and management, we focused our investigation on the problem of detecting blocked rain gauges, to improve data reliability and rainfall data products. For novelty detection in computer vision, we concentrated on the problem of multiple novelties. 
	
	In chapter \ref{ch1:comparison-of-methods}, we described four existing methods for detecting blocked rain gauges, the conditional mixture model (CMM), the PWSQC (Fz-filter), a kernel regression model and a joint Gaussian model. Using precipitation data from the Oklahoma Mesonet and the TAHMO Kenya weather networks, we compared the methods on simulated blocked rain gauges. We evaluated these methods under three fault conditions including (a) varying the proportions of faults, (b) varying the distance from each station to its nearest neighbors, and (c) varying the different spatial distribution of faults. We studied two types of spatial distribution of faults: when the faulty sensors are randomly distributed and become fail independently, and when the faulty sensors are spatially clustered and fail simultaneously. We further studied scenarios where the clustered faults are surrounded by working sensors versus scenarios where they are at the periphery of the network. We analyzed the detection delay, the missed alarm rate, and the false alarm rate of the methods. This analysis identified the conditions under which each method performed well and showed that the these methods are not robust to large clusters of faulty sensors.
	
	In chapter \ref{ch2:pcsd}, we designed three general purpose QC methods, PCSD-SAT, PCSD-JIKR, and PCSD-JIGM. These methods integrate probabilistic fault detection into a consistency-based diagnosis framework to perform joint reasoning. Using rainfall from the Oklahoma Mesonet and the TAHMO Kenya weather networks, we evaluated the proposed methods on the task of detecting blocked rain gauges. We specifically tested them on the detection of spatially random faults (SRF) and spatially clustered faults (SCF-IN and SCF-OUT). Using detection delay, false alarm rate and missed alarm rate, we compared PCSD-SAT, PCSD-JIKR, and PCSD-JIGM against the QC methods from Chapter \ref{ch1:comparison-of-methods} as baselines. We showed empirically that PCSD-JIKR and PCSD-JIGM outperform all the baselines at detecting large clusters of faults.
	
	In chapter \ref{ch3:novel_ss}, we introduced NOVEL-SS, a benchmark dataset for multiple novelty detection and accommodation in computer vision. Along with NOVEL-SS, we presented DCA (detect, accommodate and characterize), a baseline system for NOVEL-SS. We showed empirically that computer systems that can handle multiple types of movelty are more robust than systems trained in a closed world manner.

	\bibliographystyle{plainnat}
	\bibliography{thesis}
	
	\appendix

	\chapter{Integrating Probabilistic Consistency-Based Diagnosis and Joint Inference for Advanced Sensor Diagnosis}
	
	\section{Reduction of the boolean formula $f(S_a)$} \label{appx:boolean_formula_reduction}
	\begin{align*}
		f(S_a) &= (\bar{b}_a \land \bar{b}_b \land b_c) \lor (\bar{b}_a \land b_b \land b_c) \lor (b_a \land \bar{b}_b \land \bar{b}_c) \lor (b_a \land \bar{b}_b \land b_c) \lor (b_a \land b_b \land \bar{b}_c) \\
		&\quad \lor (b_a \land b_b \land b_c) \\
		&= \bar{b}_a \bar{b}_b b_c + \bar{b}_a b_b b_c + b_a \bar{b}_b \bar{b}_c + b_a \bar{b}_b b_c + b_a b_b \bar{b}_c + b_a b_b b_c \\
		&= (\bar{b}_a \bar{b}_b + \bar{b}_a b_b + b_a \bar{b}_b + b_a b_b) b_c + b_a \bar{b}_c (b_b + \bar{b}_b) \\
		&= \big(\bar{b}_a (\bar{b}_b + b_b) + b_a (\bar{b}_b + b_b) \big) b_c + b_a \bar{b}_c \\
		&= (\bar{b}_a + b_a) b_c + b_a \bar{b}_c \\
		&= b_c + b_a \bar{b}_c \\
		&= b_c + b_a
	\end{align*}

	\chapter{Novel-SS: A Benchmark for Integrated Novelty-Aware Computer Vision Systems}
	
	\section{Appendix / supplemental material}

	\section{Novel-SS Dataset Details}
	\label{app:ch3_dataset}
	
	\subsection{Filtering and Annotation Adjustment}
	
	The SS dataset annotations are given at the sequence level. In Novel-SS, we ``push'' these annotations down to the individual images. This can lead to a mismatch between the annotation and the actual number of animals visible in the individual image.  One case arises when an animal runs very close to the camera, gets captured in the first frame, but leaves the field of view by the time the second and third frames are captured.  Another case occurs when an animal at the edge of the field of view triggers the camera and then moves out of view (see the last 2 images of Fig. \ref{fig:exple_label_noise_n_occluded_animal}). To improve the per-image labels, we run all candidate images through the MegaDetector animal detector \cite{beery2019efficient, microsoft2020} and adjust the animal counts so that the total counts match the number of bounding boxes returned by the MegaDetector. For images annotated as containing only one species, we accept the MegaDetector's count as the ground truth. Images annotated as containing more than one species are processed as follows. If the MegaDetector's count matches both the sum of per-species maximum counts and the sum of per-species median counts, then we accept the annotation. If the annotation involves only two species and if the MegaDetector's count matches either the sum of per-species maximum counts or the sum of per-species median counts (but not both), then we adopt the corresponding annotation count (max or median) as ground truth for the minority species and label the majority species with the difference between the MegaDetector's count and the minority species count. All other cases where the annotation does not match the MegaDetector are discarded. Examples of detection are presented in Fig. \ref{fig:exple_animal_detection_megadetector}.
	
	To further improve the quality of the images in the Novel-SS, the images containing novel species that successfully passed through the filtering process above are manually curated to exclude highly-occluded images, as well as dawn/dusk images that are visually indistinguishable from no-novelty images.
	
	Visual inspection of single images suggested to us that it is not easy to distinguish between the ``standing'' and ``moving'' activities. To test whether a deep net can perform this task, we fine-tuned a ResNet18 model pre-trained on ImageNet on the binary task of predicting moving vs.~standing behavior \cite{deng2009imagenet, he2016deep}. The dataset was the union of the no-novelty and novelty images containing a single species either standing or moving. Somewhat to our surprise, the network was able to achieve {78.5\%} correct discrimination. Hence, we retained standing and moving as distinct activities.
	
	\subsection{SS-NOVEL Dataset Statistics}
	
\begin{figure}
		\centering
		\includegraphics[width=0.5\textwidth]{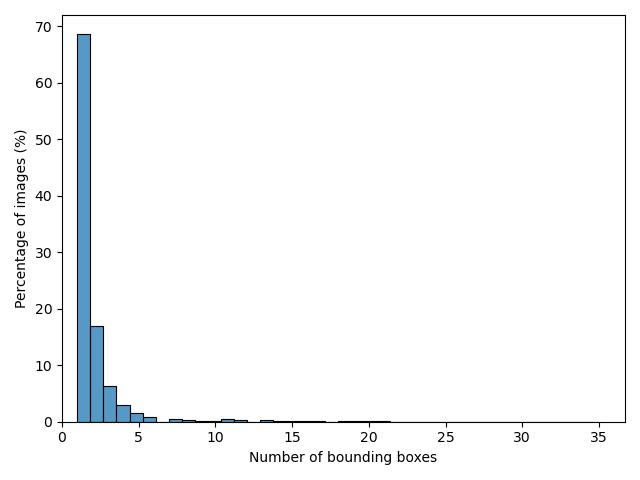}
		
		\caption{Distribution of bounding boxes counts in the Novel-SS.}
		\label{fig:distribution_bboxes_counts}
	\end{figure}

\begin{figure*}
		\centering		
		\begin{subfigure}{\textwidth}
			\centering
			\includegraphics[width=\textwidth]{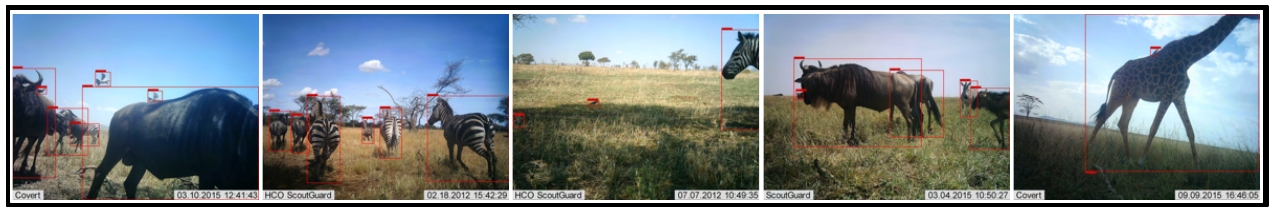}
			
			\caption{Examples of excellent detections by the Megadetector.}
			\label{fig:exple_animal_detection_megadetector}
		\end{subfigure} \hfill
		\begin{subfigure}{\textwidth}
			\centering
			\includegraphics[width=\textwidth]{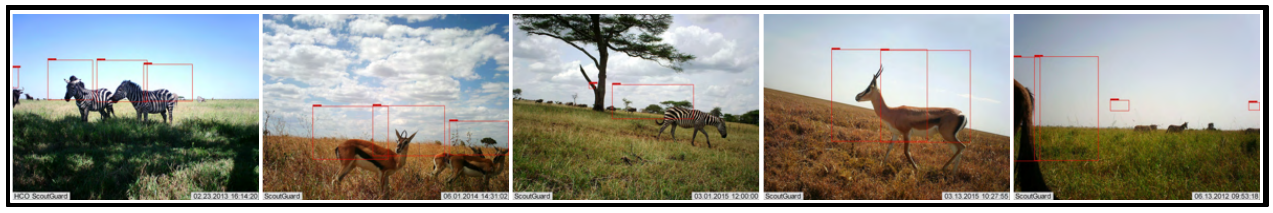}
			
			\caption{Example of miss-detection by the Megadetector including false detections, missed detections, and poorly-aligned bounding boxes.}
			\label{fig:exple_wrong_detection_megadetector}
		\end{subfigure} \hfill
		\begin{subfigure}{\textwidth}
			\centering
			\includegraphics[width=\textwidth]{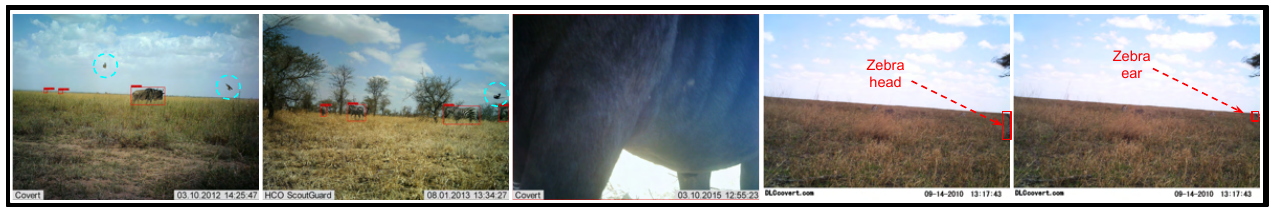}
			
			\caption{Additional missed detections (circled) and cases where the image boundary clips much or all of the animal.}
			\label{fig:exple_label_noise_n_occluded_animal}
		\end{subfigure}
		\caption{Images illustrating the behavior of the Megadetector.}
		\label{fig:megadetector_failures}
	\end{figure*}

	Table~\ref{tab:summary_train_val_test} summarizes species and activities that define the training, validation, and testing datasets. Figure~\ref{fig:distribution_bboxes_counts} summarizes the number of bounding boxes per image in Novel-SS. Figure~\ref{fig:megadetector_failures} provides examples of bounding boxes including failures of the Megadetector and cases where a single animal either fills the entire image (with extreme clipping/occlusion) or is barely visible in the image. Table~\ref{fig:statistics_images_in_novel_ss} provides statistics on the number of images in the validation and test set where pairs of species co-occur. Unsurprisingly, most animals are not observed with either hyenas or lions. Note that animal counts vary hugely across species. Table~\ref{tab:image_counts_nov_activities} provides similar information for the co-occurrence of activities. We can observe that lions spend virtually all of their time resting, and we only have one image of a lion eating. Figure~\ref{fig:statistics_images_in_novel_ss} provides statistics on the number of images per species in the training, validation, and test sets. 
	
	\begin{table*}
		\centering
		\def\arraystretch{1.2}
		\begin{tabularx}{\textwidth}{l|c|C|C|C|C|C|C|}
			\cline{3-8}
			\multicolumn{2}{c|}{} & \textbf{no-novelty} & \textbf{Novel species} & \textbf{Novel comb species} & \textbf{Novel activities} & \textbf{Novel comb activities} & \textbf{Novel env} \\ 
			\toprule
			\multirow{3}{*}{Train} & spe & all known & - & - & - & - & -   \\
			\cline{2-8} & act & \{\textit{St}\} or \{\textit{Mo}\} & - & - & - & - & -  \\
			\cline{2-8} & env & \{\textit{day}\} & - & - & - & - & -  \\
			\bottomrule
			\multirow{3}{*}{Valid} & spe & - & \{{\color{red} \textit{Ko}}\} or \{{\color{red} \textit{Ela}}\} & \{{\color{red}(}\textit{Ze, GT}{\color{red})}, {\color{red}(}\textit{Ze, Gi}{\color{red})}, {\color{red}(}\textit{OB, Wi}{\color{red})}, {\color{red}(}\textit{OB, GT}{\color{red})}\} & \{\textit{Wi}\} or \{\textit{To}\} & \{\textit{El}\}, \{\textit{Wi}\} or \{\textit{To}\} & \{\textit{Ze}\} or \{\textit{Wi}\} \\
			\cline{2-8} & act & - & \{\textit{St}\} or \{\textit{Mo}\}  & \{\textit{St}\} or \{\textit{Mo}\} & \{{\color{red} \textit{Ea}}\} & \{\textit{St}\} {\color{red}and} \{\textit{Mo}\} & \{\textit{St}\} or \{\textit{Mo}\} \\
			\cline{2-8} & env & - & \{\textit{day}\} & \{\textit{day}\} & \{\textit{day}\} & \{\textit{day}\} & \{{\color{red} \textit{fog, snow, dawn/dusk}}\} \\
			\bottomrule
			\multirow{3}{*}{Test} & spe & all known & {\color{red} novel spe} not in valid & other comb not in valid & known spe not in valid & known spe not in valid & known spe not in valid \\
			\cline{2-8} & act &  \{\textit{St}\} or \{\textit{Mo}\} &  \{\textit{St}\} or \{\textit{Mo}\} &  \{\textit{St}\} or \{\textit{Mo}\} & \{{\color{red} \textit{Ea}}\} and/or \{{\color{red} \textit{Re}}\} & \{\textit{St}\} {\color{red}and} \{\textit{Mo}\} & \{\textit{St}\} or \{\textit{Mo}\} \\
			\cline{2-8} & env & \{\textit{day}\} & \{\textit{day}\} & \{\textit{day}\} & \{\textit{day}\} & \{\textit{day}\} & \{{\color{red} \textit{fog, snow, dawn/dusk, night}}\} \\
			\bottomrule
		\end{tabularx}
		\caption{Summary of species (spe), activities (act) and environment (env) in the training, validation and test data sets of the Novel-SS. Notation \{(A, B)\} means that A and B are simultaneously present in the image while \{A, B\} means that at least one of A and B is present. The red color mean the species, activity or environment is novel. \textbf{Species abbreviation}: \textit{Ko}=Koribustard, \textit{El}=Elephant, \textit{Ela}=Eland, \textit{Ze}=Zebra, \textit{Gi}=Giraffe, \textit{Wi}=Wildebeest, \textit{To}=Topi, \textit{GT}=Gazelle Thomsons, \textit{OB}=Other bird. \textbf{Activities abbreviation}: \textit{St}=standing, \textit{Mo}=moving, \textit{Ea}=eating, \textit{Re}=resting.}
		\label{tab:summary_train_val_test}
	\end{table*}
	
	\begin{figure*}
		\begin{subfigure}{0.49\linewidth}
			\includegraphics[width=\linewidth]{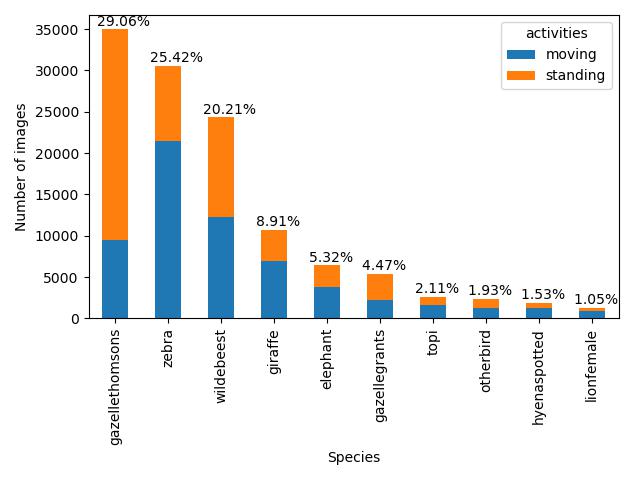}
			\caption{Number of images per species in the training set (no-novelty images).}
			\label{fig:train_set_image_counts}
		\end{subfigure}
		\hfill
		\begin{subfigure}{0.49\linewidth}
			\includegraphics[width=\linewidth]{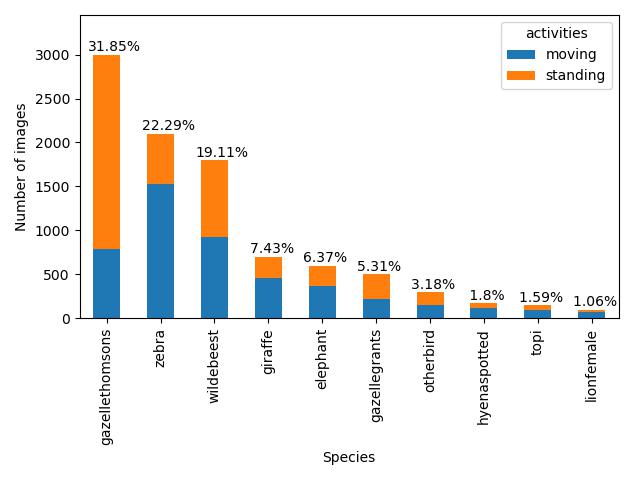}
			\caption{Number of images per species in the no-novelty set of the test set.}
			\label{fig:test_set_prenovelty_image_counts}
		\end{subfigure}
		\bigskip
		\begin{subfigure}{0.54\linewidth}
			\includegraphics[width=\linewidth]{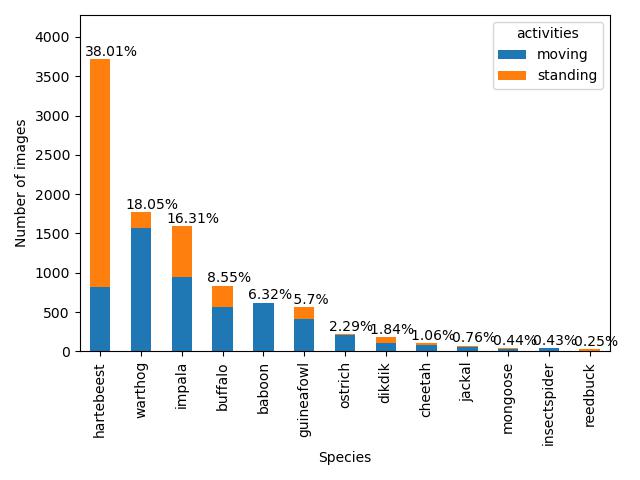}
			\caption{Number of images per species in the test set of the novel species novelty type..}
			\label{fig:test_set_nov_spe_image_counts}
		\end{subfigure}
		\hfill
		\begin{subfigure}{0.45\linewidth}
			\includegraphics[width=\linewidth]{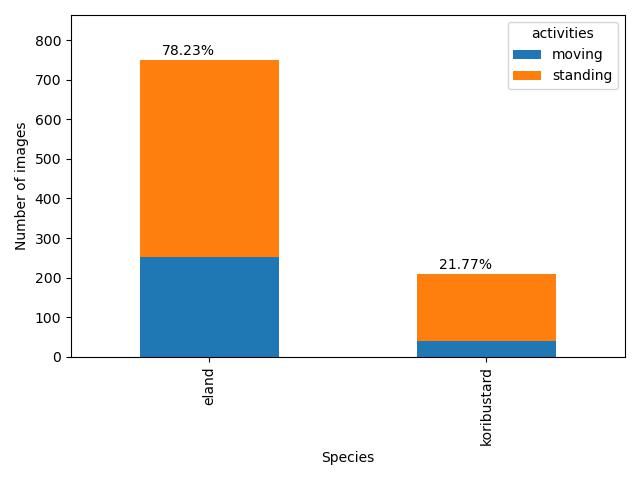}
			\caption{Number of images per species in the validation set of the novel species novelty type.}
			\label{fig:valid_set_nov_spe_image_counts}
		\end{subfigure}
		
		\bigskip
		\begin{subfigure}{0.49\linewidth}
			\includegraphics[width=\textwidth]{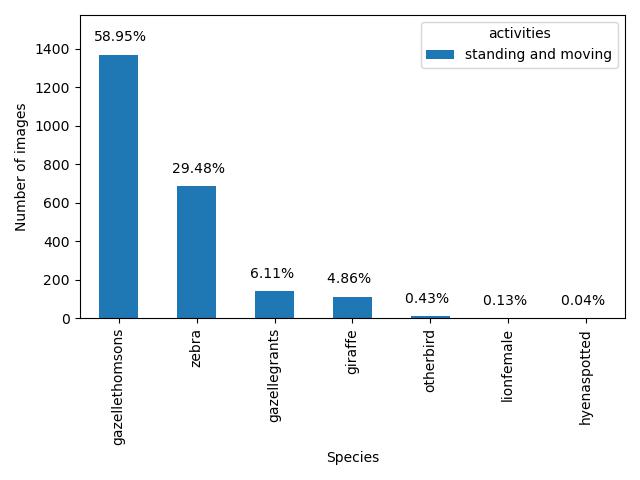}
			\caption{Number of images per species in the test set of the novel combination of known activities novelty type.}
			\label{fig:test_set_nov_comb_act_image_counts}
		\end{subfigure}
		\hfill
		\begin{subfigure}{0.49\linewidth}
			\includegraphics[width=\textwidth]{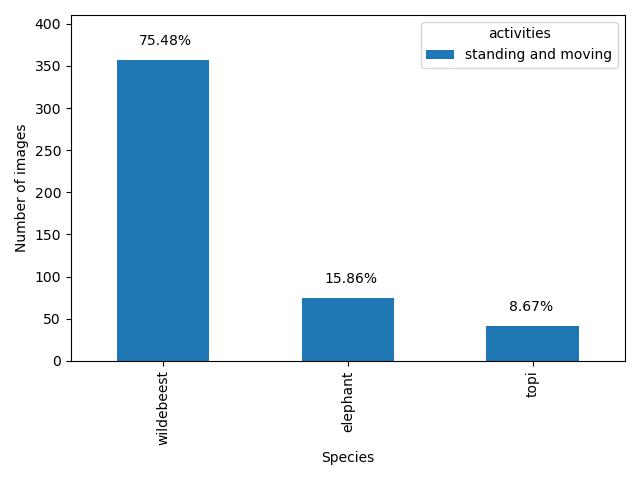}
			\caption{Number of images per species in the validation set of the novel combination of known activities novelty type.}
			\label{fig:valid_set_nov_comb_act_image_counts}
		\end{subfigure}
		\caption{Statistics of images in the Novel-SS.}
	\end{figure*}
	\begin{figure*}[!tbp]\ContinuedFloat
		\centering
		\begin{subfigure}{0.49\textwidth}
			\includegraphics[width=\textwidth]{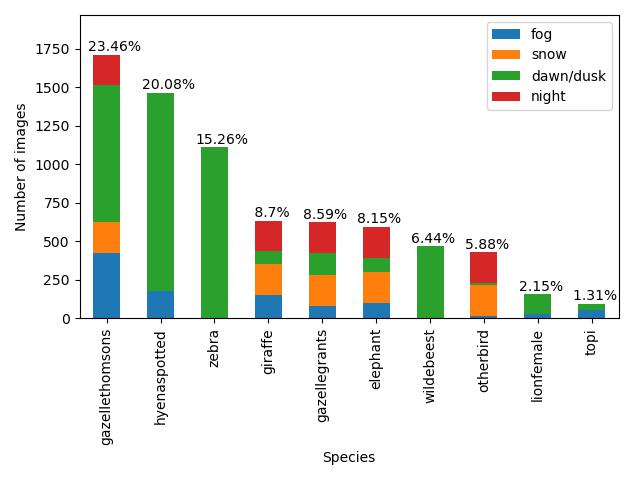}
			\caption{Number of images per species in the test set of the novel environment novelty type.}
			\label{fig:test_set_nov_env_image_counts}
		\end{subfigure}
		\hfill
		\begin{subfigure}{0.49\textwidth}
			\includegraphics[width=\textwidth]{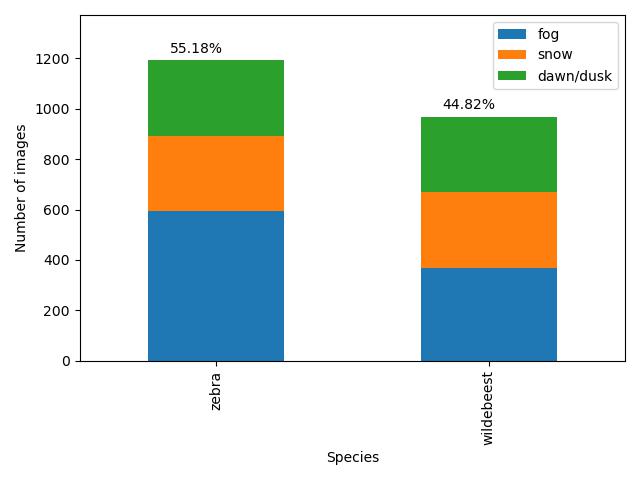}
			\caption{Number of images per species in the validation set of the novel environment novelty type.}
			\label{fig:valid_set_nov_env_image_counts}
		\end{subfigure}
		\caption{Statistics of images in the Novel-SS (cont.).}
		\label{fig:statistics_images_in_novel_ss}
	\end{figure*}

	\begin{table*}[ht]
		\centering
		\setlength\tabcolsep{1.5pt}
		\begin{tabular}{@{}l|c|c|c|c|c|c|c|c|c|c|@{}}
			\cline{2-11}
			\multicolumn{1}{c|}{ \rotatebox[origin=c]{45}{\textbf{Species}}} & \rotatebox[origin=c]{90}{Zebra} & \rotatebox[origin=c]{90}{Wildebeest} & \rotatebox[origin=c]{90}{ Gazelle Thom. } & \rotatebox[origin=c]{90}{Gazelle Grants} & \rotatebox[origin=c]{90}{Other bird} & \rotatebox[origin=c]{90}{Topi} & \rotatebox[origin=c]{90}{Giraffe} & \rotatebox[origin=c]{90}{Elephant} & \rotatebox[origin=c]{90}{Hyena spotted} & \rotatebox[origin=c]{90}{Lion female}  \\
			\hline
			Zebra & \cellcolor{black!50} & {\color{red} 547};{\color{blue} 727} & \cellcolor{yellow!75} {\color{red} 42} ; {\color{blue} 55} & {\color{red} 12} ; {\color{blue} 4} & {\color{red} 25} ; {\color{blue} 32} & {\color{red} 18} ; {\color{blue} 9} & \cellcolor{yellow!75}{\color{red} 21} ; {\color{blue} 19} & {\color{red} 8} ; {\color{blue} 6} & {\color{red} 0} ; {\color{blue} 4} & {\color{red} 0} ; {\color{blue} 2}  \\ \hline
			Wildebeest & {\color{red} 547};{\color{blue} 727} & \cellcolor{black!50} & {\color{red} 20} ; {\color{blue} 11} & {\color{red} 11} ; {\color{blue} 10} & \cellcolor{yellow!75} {\color{red} 19} ; {\color{blue} 25} & \cellcolor{black!50} & {\color{red} 4} ; {\color{blue} 2} & {\color{red} 4} ; {\color{blue} 0} & {\color{red} 0} ; {\color{blue} 1} & {\color{red} 0} ; {\color{blue} 3} \\ \hline
			Gazelle Thom. & \cellcolor{yellow!75} {\color{red} 42} ; {\color{blue} 55} & {\color{red} 20} ; {\color{blue} 11} & \cellcolor{black!50} & {\color{red} 50} ; {\color{blue} 12} & \cellcolor{yellow!75}{\color{red} 109};{\color{blue} 17} & {\color{red} 10} ; {\color{blue} 11} & {\color{red} 6} ; {\color{blue} 6} & {\color{red} 0} ; {\color{blue} 3} & {\color{red} 3} ; {\color{blue} 0} & \cellcolor{black!50}  \\ \hline
			Gazelle Grants & {\color{red} 12} ; {\color{blue} 4} & {\color{red} 11} ; {\color{blue} 10} & {\color{red} 50} ; {\color{blue} 12} & \cellcolor{black!50} & {\color{red} 9} ; {\color{blue} 3} & {\color{red} 2} ; {\color{blue} 0} & {\color{red} 0} ; {\color{blue} 3} & \cellcolor{black!50} & \cellcolor{black!50} & \cellcolor{black!50}  \\ \hline
			Other bird & {\color{red} 25} ; {\color{blue} 32} & \cellcolor{yellow!75} {\color{red} 19} ; {\color{blue} 25} & \cellcolor{yellow!75}{\color{red} 109};{\color{blue} 17} & {\color{red} 9} ; {\color{blue} 3} & \cellcolor{black!50} & {\color{red} 3} ; {\color{blue} 0} & {\color{red} 1} ; {\color{blue} 8} & {\color{red} 2} ; {\color{blue} 7} & {\color{red} 2} ; {\color{blue} 3} & \cellcolor{black!50}  \\ \hline
			Topi & {\color{red} 18} ; {\color{blue} 9} & \cellcolor{black!50} & {\color{red} 10} ; {\color{blue} 11} & {\color{red} 2} ; {\color{blue} 0} & {\color{red} 3} ; {\color{blue} 0} & \cellcolor{black!50} & \cellcolor{black!50} & \cellcolor{black!50} & \cellcolor{black!50} & \cellcolor{black!50} \\ \hline
			Giraffe & \cellcolor{yellow!75} {\color{red} 21} ; {\color{blue} 19} & {\color{red} 4} ; {\color{blue} 2} & {\color{red} 6} ; {\color{blue} 6} & {\color{red} 0} ; {\color{blue} 3} & {\color{red} 1} ; {\color{blue} 8} & \cellcolor{black!50} & \cellcolor{black!50} & \cellcolor{black!50} & \cellcolor{black!50} & \cellcolor{black!50} \\ \hline
			Elephant & {\color{red} 8} ; {\color{blue} 6} & {\color{red} 4} ; {\color{blue} 0} & {\color{red} 0} ; {\color{blue} 3} & \cellcolor{black!50} & {\color{red} 2} ; {\color{blue} 7} & \cellcolor{black!50} & \cellcolor{black!50} & \cellcolor{black!50} & \cellcolor{black!50} & \cellcolor{black!50} \\ \hline
			Hyena spotted & {\color{red} 0} ; {\color{blue} 4} & {\color{red} 0} ; {\color{blue} 1} & {\color{red} 3} ; {\color{blue} 0} & \cellcolor{black!50} & {\color{red} 2} ; {\color{blue} 3} & \cellcolor{black!50} & \cellcolor{black!50} & \cellcolor{black!50} & \cellcolor{black!50} & \cellcolor{black!50}  \\ \hline
			Lion female & {\color{red} 0} ; {\color{blue} 2} & {\color{red} 0} ; {\color{blue} 3} & \cellcolor{black!50} & \cellcolor{black!50} & \cellcolor{black!50} & \cellcolor{black!50} & \cellcolor{black!50} & \cellcolor{black!50} & \cellcolor{black!50} & \cellcolor{black!50} \\ \hline
		\end{tabular}
		\caption{Validation and test image counts for novel combination of known species novelty type. The cells with yellow background are the combination of species in the \hl{validation set} and the ones with white background are the test set. The red colored numbers are the counts of images with animals engaged in the activity "{\color{red}standing}" and the blue ones are those with animals engaged in the activity "{\color{blue}moving}". Four validation images (not included in the table) contain 3 species: 2 of them contain the triplet (giraffe, wildebeest, zebra) and 2 contain the triplet (gazelle Thomsons, gazelle Grants, zebra). The test set on the other hand contains only 1 image with 3 species (other bird, wildebeest and zebra).}
		\label{tab:image_counts_nov_comb_species}
	\end{table*}
	
	\begin{table*}[ht]
		\centering
		\setlength\tabcolsep{4.5pt}
		\begin{tabular}{@{}l|c|c|c|c|c|c|c|c|c|c|@{}}
			\cline{2-11}
			\diagbox{\\ \textbf{Activities} \\}{\\ \textbf{Species} \\} & \rotatebox[origin=c]{90}{Zebra} & \rotatebox[origin=c]{90}{Wildebeest} & \rotatebox[origin=c]{90}{ Gazelle Thom. } & \rotatebox[origin=c]{90}{Gazelle Grants} & \rotatebox[origin=c]{90}{Other bird} & \rotatebox[origin=c]{90}{Topi} & \rotatebox[origin=c]{90}{Giraffe} & \rotatebox[origin=c]{90}{Elephant} & \rotatebox[origin=c]{90}{Hyena spotted} & \rotatebox[origin=c]{90}{Lion female}  \\
			\hline
			\color{red} eating & 1,306 & \cellcolor{yellow!75} 892 & 1,895 & 346 & 9 & \cellcolor{yellow!75} 40 & 38 & 14 & \cellcolor{black!50} & 1 \\ \hline
			moving $+$ \color{red} eating & 24 & \cellcolor{yellow!75} 10 & 15 & 4 & \cellcolor{black!50} & \cellcolor{black!50} & \cellcolor{black!50} & \cellcolor{black!50} & \cellcolor{black!50} & \cellcolor{black!50} \\ \hline
			standing $+$ \color{red} eating & 96 & \cellcolor{yellow!75} 52 & 380 & 17 & \cellcolor{black!50} & \cellcolor{yellow!75} 2 & \cellcolor{black!50} & \cellcolor{black!50} & \cellcolor{black!50} & \cellcolor{black!50} \\ \hline
			\color{red} resting & 16 & 412 & 405 & 126 & 18 & 31 & \cellcolor{black!50} & \cellcolor{black!50} & 249 & 1,626 \\ \hline
			moving $+$ \color{red} resting & \cellcolor{black!50} & 7 & 1 & \cellcolor{black!50} & \cellcolor{black!50} & \cellcolor{black!50} & \cellcolor{black!50} & \cellcolor{black!50} & \cellcolor{black!50} & 6 \\ \hline
			standing $+$ \color{red} resting & 23 & 393 & 341 & 10 & \cellcolor{black!50} & 7 & \cellcolor{black!50} & \cellcolor{black!50} & \cellcolor{black!50} & 1 \\ \hline
			\color{red} eating \color{black} $+$ \color{red} resting & \cellcolor{black!50} & \cellcolor{black!50} & 10 & \cellcolor{black!50} & \cellcolor{black!50} & 2 & \cellcolor{black!50} & \cellcolor{black!50} & \cellcolor{black!50} & \cellcolor{black!50} \\ \hline
			standing $+$ \color{red} eating \color{black} $+$ \color{red} resting & \cellcolor{black!50} & 7 & 12 & \cellcolor{black!50} & \cellcolor{black!50} & \cellcolor{black!50} & \cellcolor{black!50} & \cellcolor{black!50} & \cellcolor{black!50} & \cellcolor{black!50} \\ \hline
		\end{tabular}
		\caption{Validation and test image counts for novel activities novelty type. The cells with yellow background are the species and activities in the \hl{validation set} and the ones with white background are the test set. The red colored activities are the {\color{red}novel activities} and the lack ones are the known activities.}
		\label{tab:image_counts_nov_activities}
	\end{table*}

	\clearpage
	\section{Further details on DCA-SS}
	\label{app:DCA-details}

	\subsection{Conditioning species probabilities for each bounding box on the novelty type}
	
	We begin by describing how we handle species predictions. Let $known$ denote the set of all known species class labels, and let $novel$ denote the set of all novel species class labels. We will break the task into three cases based on the novelty type. 
	
	\subsubsection{Species probabilities when only one species can be present}
	For some novelty types, conditioning the bounding box predictions on the novelty type $t$ is relatively easy. We first consider $t \in $ \{no-novelty, novel environment, novel activity, novel combination of known activities\}. Under these novelty types, every bounding box in the image must belong to the same known species. Let $E_0(i,k)$ denote the event that all bounding boxes in image $i$ belong to species $k$. Then there are only $K_s$ possible events: $E_0(i,1), \ldots, E_0(i,K_s)$. Hence, if we consider bounding box $b$, the $P(y_{ib}=k \mid t_i = t,x_i) = P(E_0(i,k) \mid x_i).$ We can compute this as
	
	\begin{equation}
		\label{eq:trivial_species_count}
		\begin{aligned}
			& P(y_{ib} = k \mid t_i = t, x_i)\\
			& \hspace{10pt} = P(E_0(i,k) \mid x_i)\\
			& \hspace{10pt} = \frac{\prod_{b'} P(y_{ib'} = k \mid x_i)}{\sum_{k' \in known} \prod_{b'} P(y_{ib'} = k' \mid x_i)}.
		\end{aligned}
	\end{equation}
	For every novel species $k$, we set $P(y_{ib}=k \mid t_i = t) = 0$  to respect the property that images with these novelty types cannot contain novel species. In the remaining derivations, we will omit the explicit conditioning on $x_i$ for clarity. 
	
	\subsubsection{Species probabilities for the novel species type}
	We now consider the case where the novelty type $t =$ ``novel species''. In this case, we know that at least one novel species must be present in the image and there can be at most one known species present in the image. We will name these two conditions. Let $E_1(i)$ denote the event that there is at most one known species present in $x_i$, and let $E_2(i)$ denote the event that there is at least one novel species present in $x_i$. Our goal can be expressed as $P(y_{ib}=k \mid E_1(i), E_2(i))$ for each bounding box $b =1,\ldots,B_i$. Conditioning on $E_1(i)$ and $E_2(i)$ is equivalent to conditioning on $t_i=t$. We will condition on one event at a time, starting with $E_1(i)$.
	
	Let $E_3(i,b,k)$ denote the event $\forall b' \neq b: y_{ib'} \in \{k\} \cup novel$ (``all other bounding boxes $b'$ in image $i$ either belong to class $k$ or belong to a novel class''). Making an independence assumption across bounding boxes, for each known species class $k$, we compute
	
	\begin{equation}
		\label{eq:novel_count_1}
		\begin{aligned}
			& \hspace{13pt} P(y_{ib}=k, E_1(i))\\
			& = P(y_{ib}=k, E_3(i,j,k))\\
			& = P(y_{ib}=k) P(E_3(i,j,k))\\
			& = P(y_{ib}=k ) \prod_{b' \neq b} P(y_{ib'} \in \{k\} \cup novel)\\
			& = P(y_{ib}=k) \prod_{b' \neq b} \hspace{5pt} \sum_{k' \in \{k\} \cup novel} P(y_{ib'} = k')\\
		\end{aligned}
	\end{equation}
	
	Next, we compute the same probabilities for each novel species class $k$:
	
	\begin{equation}
		\label{eq:novel_count_2}
		\begin{aligned}
			& \hspace{13pt} P(y_{ib}=k, E_1(i))\\
			& = P(y_{ib}=k, \bigvee_{k' \in known} (\forall b' \neq b: S_{ib'} \in \{k'\} \cup novel))\\
			& = P(y_{ib}=k, \bigvee_{k' \in known} E_3(i,b,k'))\\
			& = P(y_{ib}=k) P(\bigvee_{k' \in known} E_3(i,b,k'))\\
			& = P(y_{ib}=k) \Big [\\
			& \hspace{8pt} \sum_{k' \in known} P(E_3(i,b,k')))\\
			& \hspace{8pt} - \sum_{k'_1 \in known} \hspace{5pt} \sum_{k'_2 \in known ; k'_2 > k'_1} P(E_3(i, b, k'_1), E_3(i, b, k'_2))\\
			& \hspace{8pt} + \sum_{k'_1 \in known} \hspace{5pt} \sum_{k'_2 \in known ; k'_2 > k'_1} \hspace{5pt} \sum_{k'_3 \in known, k'_3 > k'_2} ...\\
			& \Big].\\
		\end{aligned}
	\end{equation}
	
	Let $E_4(i,b)$ denote the event $\forall b' \neq b: y_{ib'} \in novel$. We can simplify (\ref{eq:novel_count_2}) as follows:
	
	\begin{equation}
		\begin{aligned}
			& ... = P(y_{ib}=k) \Big [\\
			& \hspace{8pt} \sum_{k' \in known} P(E_3(i,b,k')))\\
			& \hspace{8pt} + P(E_4(i,b)) \Big(-{K\choose 2} + {K \choose 3} - ... {K \choose K}\Big)\\
			& \Big]\\
			& = P(y_{ib}=k) \Big [\\
			& \hspace{8pt} \sum_{k' \in known} P(E_3(i,b,k')))\\
			& \hspace{8pt} + P(E_4(i,b)) (1 - K)\\
			& \Big].\\
		\end{aligned}
	\end{equation}
	\noindent $P(E_3(i,b,k'))$ can be computed as in (\ref{eq:novel_count_1}), and $P(E_4(i,b))$ can be computed similarly.
	
	Next, we compute the following conditional probabilities for each class $k$:
	
	\begin{equation}
		\label{eq:p_sij_given_e1}
		\begin{aligned}
			& \hspace{13pt} P(y_{ib}=k \mid E_1(i))\\
			& = \frac{P(y_{ib}=k, E_1(i))}{P(E_1(i))}\\
			& = \frac{P(y_{ib}=k, E_1(i))}{\sum_{k'} (P(y_{ib}=k', E_1(i)))}.
		\end{aligned}
	\end{equation}
	
	We must now further condition on $E_2(i)$. To this end, we compute the following for each known class $k$ and bounding box $b$:
	
	\begin{equation}
		\begin{aligned}
			& \hspace{13pt} P(y_{ib}=k, E_2(i) \mid E_1(i))\\
			& = P(y_{ib}=k \mid E_1(i)) P(E_2(i) \mid y_{ib}=k, E_1(i))\\
			& = P(y_{ib}=k \mid E_1(i))\\
			& \hspace{20pt} (1 - P(\forall b' \neq b: S_{ib'} \in known \mid y_{ib}=k, E_1(i)))\\
			& = P(y_{ib}=k \mid E_1(i))\\
			& \hspace{20pt} (1 - P(\forall b' \neq b: S_{ib'} = k \mid S_{ib'} \in \{k\} \cup novel))\\
			& = P(y_{ib}=k \mid E_1(i))\\
			& \hspace{20pt} \Big(1 - \prod_{b' \neq b}P(S_{ib'} = k \mid S_{ib'} \in \{k\} \cup novel)\Big)\\
			& = P(y_{ib}=k \mid E_1(i))\\
			& \hspace{20pt} \Big(1 - \prod_{b' \neq b}\frac{P(S_{ib'} = k)}{\sum_{k' \in \{k\} \cup novel} P(S_{ib'} = k')}\Big).\\
		\end{aligned}
	\end{equation}

	We then compute the same probabilities for each novel class $k$ and bounding box $b$, which is trivial:
	
	\begin{equation}
		\begin{aligned}
			& \hspace{13pt} P(y_{ib}=k, E_2(i) \mid E_1(i)) = P(y_{ib}=k \mid E_1(i)).\\
		\end{aligned}
	\end{equation}
	\noindent Notice that this has already been computed in (\ref{eq:p_sij_given_e1}).
	
	Finally, we condition on $E_2(i)$ like so:
	
	\begin{equation}
		\begin{aligned}
			& \hspace{13pt} P(y_{ib}=k \mid t_i = \text{ ``novel species''})\\
			& = P(y_{ib}=k \mid E_1(i), E_2(i))\\
			& = \frac{P(y_{ib}=k, E_2(i) \mid E_1(i))}{P(E_2(i) \mid E_1(i))}\\
			& = \frac{P(y_{ib}=k, E_2(i) \mid E_1(i))}{\sum_{k'}P(y_{ib}=k', E_2(i) \mid E_1(i))}.
		\end{aligned}
	\end{equation}
	
	\subsubsection{Species counts for novel combinations of existing species.} This case is simpler and can be handled with techniques very similar to those used in the previous section. The general pattern is the same. We break down each novelty type $t$ into sub-events, and we condition our predictions on the sub-events one at a time. To condition on a sub-event $E$, we simply compute:
	\begin{equation}
		\begin{aligned}
			P(y_{ib}=k \mid E) = \frac{P(y_{ib}=k, E)}{P(E)}.
		\end{aligned}
	\end{equation}
	\noindent Once the numerator has been computed, the denominator can be computed by further marginalizing out the species class (i.e., by normalization).
	
	\subsection{Conditioning activity presence for each bounding box on the novelty type}
	
	Activities can be handled in exactly the same way as species. For images whose novelty type is no-novelty, novel species, novel combination of species, or novel environment, all bounding boxes must have the same activity. For the novel activity type, at least one novel activity must appear and at most one known activity can appear. Finally, in the novel combination of known activities, at least two known activities (and no novel activities) must be present.  
	
	\subsection{Dynamic programs for weak image-level labels}
	
	The feedback received in the accommodation phase only contains weak count-based species labels and presence-based activity labels at the image level rather than class labels at the box level. Accommodating this feedback requires solving two separate multiple instance problems. Shukla \etal~\cite{mi_dyn_programs} provides a formulation of a dynamic program that computes the predicted likelihood of satisfying a weak count-based label in the binary classification setting. We extend this method to the multiclass setting, computing the predicted likelihood of a combination of box-label assignments that jointly satisfies the weak count-based labels for all species classes during the accommodation phase. We further adapt it for the presence-based multiple instance problem for activities as well.
	
	\subsubsection{Dynamic program for count-based multiple instance problem}
	
	This section explains our multiclass extension of the dynamic program provided by Shukla \etal~\cite{mi_dyn_programs} to accommodate feedback for images containing multiple unique species.
	
	Consider image $i$ wherein there are $B_i$ bounding boxes each belonging to one of $K$ classes. Let $N_i = [n_{i1}, \ldots n_{iK}]$ denote the vector of ground truth counts for each class across the $B_i$ bounding boxes, such that $n_{ik}$ is the number of the $B_i$ bounding boxes that belong to class $k$.
	
	Let us impose an ordering on the $B_i$ bounding boxes and let $\hat{p}_{bk} = P(y_{ib}=k \mid x_{ib})$ denote the probability predicted by the classifier that bounding box $b$ belongs to class $k$. We now introduce the dynamic programming table. Let $DP(c_1, c_2, ..., c_K, b)$ denote the predicted probability that, of the first $b$ boxes, $c_1$ of them belong to class $1$, $c_2$ of them belong to class $2$, and so on. Our goal is to compute the probability for the ground truth (and then take gradient ascent step to increase it).  $DP(n_{i1}, n_{i2}, \ldots, n_{iK}, N_i)$; this is the predicted likelihood that the weak count-based ground truth label $N_i$ is satisfied.

	$DP(c_1, c_2, ..., c_K, b)$ can be computed as the sum of at most $K$ disjoint sub-events about the first $b-1$ bounding boxes: in order for the first $b$ boxes to contain class counts $c_1, ..., c_K$, one possibility is that the first $b-1$ boxes contain class counts $c_1 - 1, c_2, ..., c_K$, and the $b^{th}$ box belongs to class $1$; another possibility is that the first $b-1$ boxes contain class counts $c_1, c_2 - 1, c_3, ..., c_K$, and the $b^{th}$ box belongs to class $2$; and so on. Since these events are disjoint, $DP(c_1, c_2, ..., c_K, b)$ can be computed as the sum of the probabilities of each of these events. Some of these events are inherently impossible, such as when any of $c_1, ..., c_k$ are negative; such events' probabilities are trivially zero.
	
	Each of these events' respective probabilities can be computed recursively. For example, the predicted probability that the first $b-1$ boxes contains class counts $c_1 - 1, c_2, ..., c_K$ is exactly $DP(c_1 - 1, c_2, ..., c_K, b - 1)$, and the predicted probability that the $b^{th}$ box belongs to class $1$ is simply $\hat{p}_{b1}$. To compute the joint probability of these two events, we make an independence assumption and naively multiply them together. Assuming that all probabilities are computed by conditioning on the respective boxes, this can be viewed as assuming that the labels of the boxes are conditionally independent given the boxes themselves.
	
	$DP(c_1, ..., c_K, b)$ can only be nonzero when $\sum_{k=1}^{K} c_k = b$ because each image belongs to exactly one class. The base cases are the nonzero values of the form $DP(c_1, ..., c_K, 1)$ (i.e., an event signifying that the first image belongs to some particular class). In these cases, exactly one of $c_1, ..., c_K$ must be $1$ while the rest must be $0$. Let $c_k$ be the nonzero value among $c_1, ..., c_K$. Then $DP(c_1, ..., c_K, 1)$ represents the predicted probability that the first image belongs to class $k$, which is simply $\hat{p}_{1k}$.
	
	$DP(c_1, ..., c_K, b)$ can be computed via a dynamic program. The recursive definition of $DP(c_1, ..., c_K, b)$ can depend on $DP(c'_1, ..., c'_k, b')$ only if $c'_1 \leq c_1, ..., c'_K \leq c_K,$ and $b' < b$. The goal is to compute $DP(y_1, ..., y_K, b)$, so the space of the dynamic program can be represented with a tensor with $K+1$ dimensions with respective sizes $n_{i1} + 1, n_{i2} + 1, ..., n_{iK} + 1, N_i$.
	
	Equivalently, a slice of dummy values can be prepended to the last dimension to reflect probabilities of events about the first $0$ out of $N_i$ images, and these dummy values can serve as the base cases instead. In such a case, $DP(0, 0, ..., 0) = 1$ since it reflects the probability that $0$ classes are observed among the first $0$ images, and all other $DP(c_1, ..., c_K, 0) = 0$ since they reflect the impossibility that a nonzero number of classes are observed among the first $0$ images. This is similar to the formulation depicted by Shukla \etal~\cite{mi_dyn_programs}.
	
	\subsubsection{Dynamic program for presence-based multiple instance problem}
	
	The above dynamic program can easily be adapted to the presence-based multiple instance problem for activities as well. In this problem, there are still $N_i$ bounding boxes each belonging to one of $K$ classes, but the ground truth $U_i=[u_{i1},\ldots,u_{iK}]$ is a vector of $0$s and $1$s corresponding to absent and present classes across the $N$ boxes. That is, $u_{ik}=1$ if the $k^{th}$ class is present among the $N_i$ boxes, and $u_{ik}=0$ otherwise. The value of $\hat{p}_{bk}$ is the predicted probability that bounding box $b$ belongs to class $k$. 
	
	The solution formulation is very similar. Let $DP(p_1, p_2, ..., p_K, n)$ denote the probability that, of the first $b$ images, class $1$ is absent if $p_1 = 0$ and present if $p_1 = 1$; class $2$ is absent if $p_2 = 0$ and present if $p_2 = 1$; and so on. The goal, then, is to compute $DP(u_{i1}, u_{i2}, ..., u_{iK}, N_i)$; this is the predicted likelihood that the weak presence-based ground truth label $U_i$ is satisfied.
	
	In this case, each $DP(p_1, p_2, ..., p_K, b)$ can be computed as the sum of at most $2K$ disjoint sub-events about the first $b-1$ images. For example, suppose the goal is to compute the probability that the first $b$ images only contain classes $1$ and $2$ (i.e., the goal is to compute $DP(1, 1, 0, ..., b)$). In this case, there are four possible sub-events:
	
	\begin{enumerate}
		\item The first $b-1$ images only contain classes $1$ and $2$, and the $b^{th}$ image belongs to class $1$.
		\item The first $b-1$ images only contain classes $1$ and $2$, and the $b^{th}$ image belongs to class $2$.
		\item The first $b-1$ images only contain class $1$, and the $b^{th}$ image belongs to class $2$.
		\item The first $b-1$ images only contain class $2$, and the $b^{th}$ image belongs to class $1$.
	\end{enumerate}
	
	As in the count-based problem, each of these events' respective probabilities can be computed recursively. For example, the predicted probability that the first $b-1$ images only contain classes $1$ and $2$ is exactly $DP(1, 1, 0, ..., 0, b-1)$, and the predicted probability that the $b^{th}$ image belongs to class $1$ is simply $\hat{y}_{n1}$. Once again, we compute the joint probability of these two events by making an independence assumption.
	
	Again, the base cases are values of the form $DP(p_1, ..., p_K, 1)$ where exactly one of $p_1, ..., p_K$ is $1$ and the rest are $0$ (i.e., an event signifying that the first image belongs to some particular class). Let $p_j$ be the nonzero value among $p_1, ..., p_K$. Then $DP(p_1, ..., p_K, 1)$ represents the predicted probability that the first image belongs to class $k$, which is simply $\hat{p}_{1k}$.
	
	Again, $DP(p_1, ..., p_K, b)$ can be computed via a dynamic program. The recursive dependency pattern is the same as in the count-based problem: the recursive definition of $DP(p_1, ..., p_K, b)$ can depend on $DP(p'_1, ..., p'_K, b')$ only if $p'_1 \leq p_1, ..., p'_K \leq p_K,$ and $b' < b$. The goal is to compute $DP(u_{i1}, ..., b_{iK}, N_i)$, so the space of the dynamic program can be represented with a tensor with at most $K+1$ dimensions with respective sizes $2, 2, ..., 2, N$. For classes that don't appear in any of the images, their respective dimensions are omitted from the dynamic program entirely (or, equivalently, they can be represented with dimensions of size $1$).
	
	\section{Additional DCA-SS Results}
	\label{app:results}
	
	\subsection{Tuning Continual Learning}
	
\begin{figure}
		\centering
		\includegraphics[width=\textwidth]{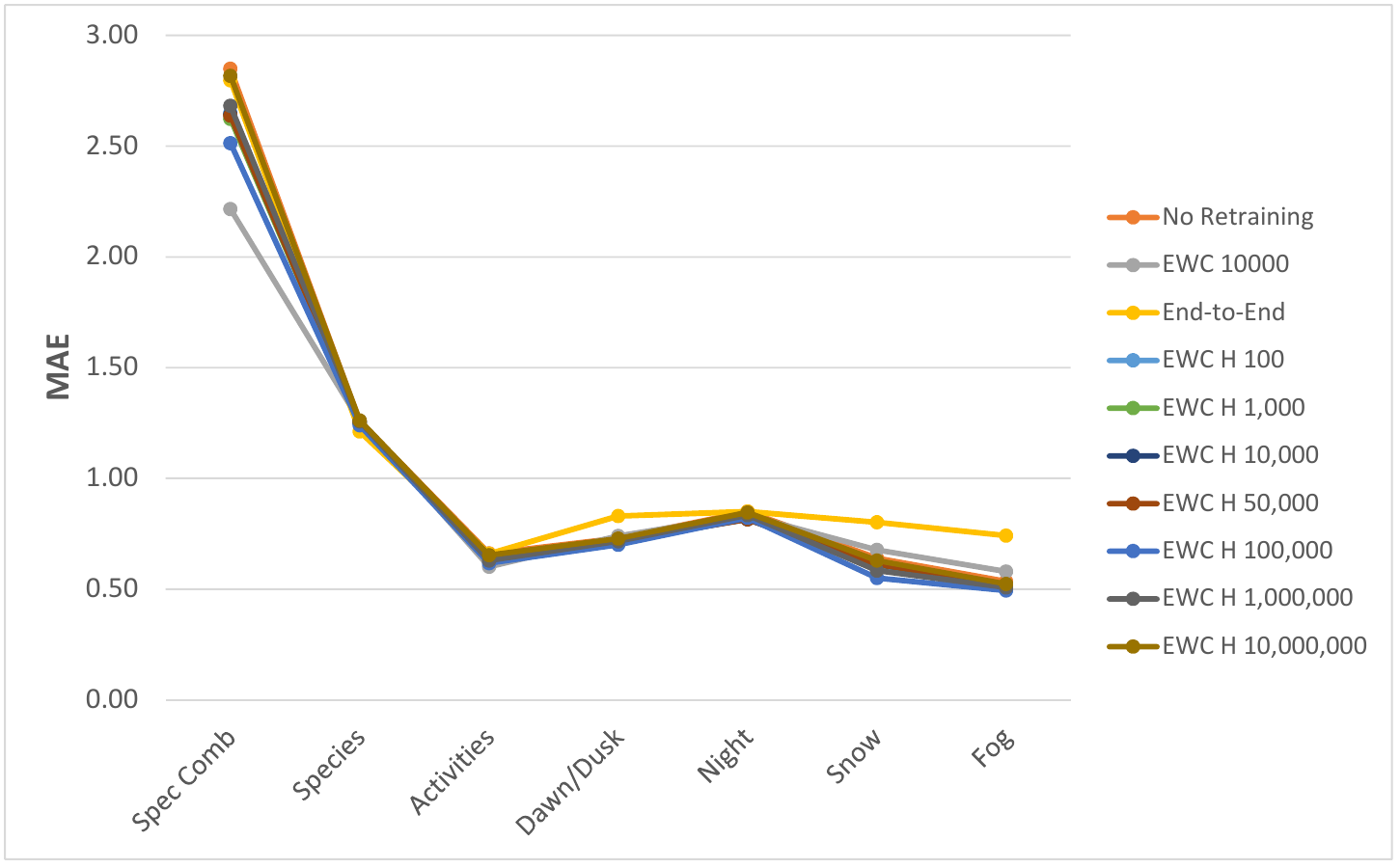}
		\caption{Comparison of Mean Average Error for Selected Trials of Various Continual Learning Configurations}
		\label{fig:mae-tuning}
	\end{figure}

\begin{figure}
		\centering
		\includegraphics[width=\textwidth]{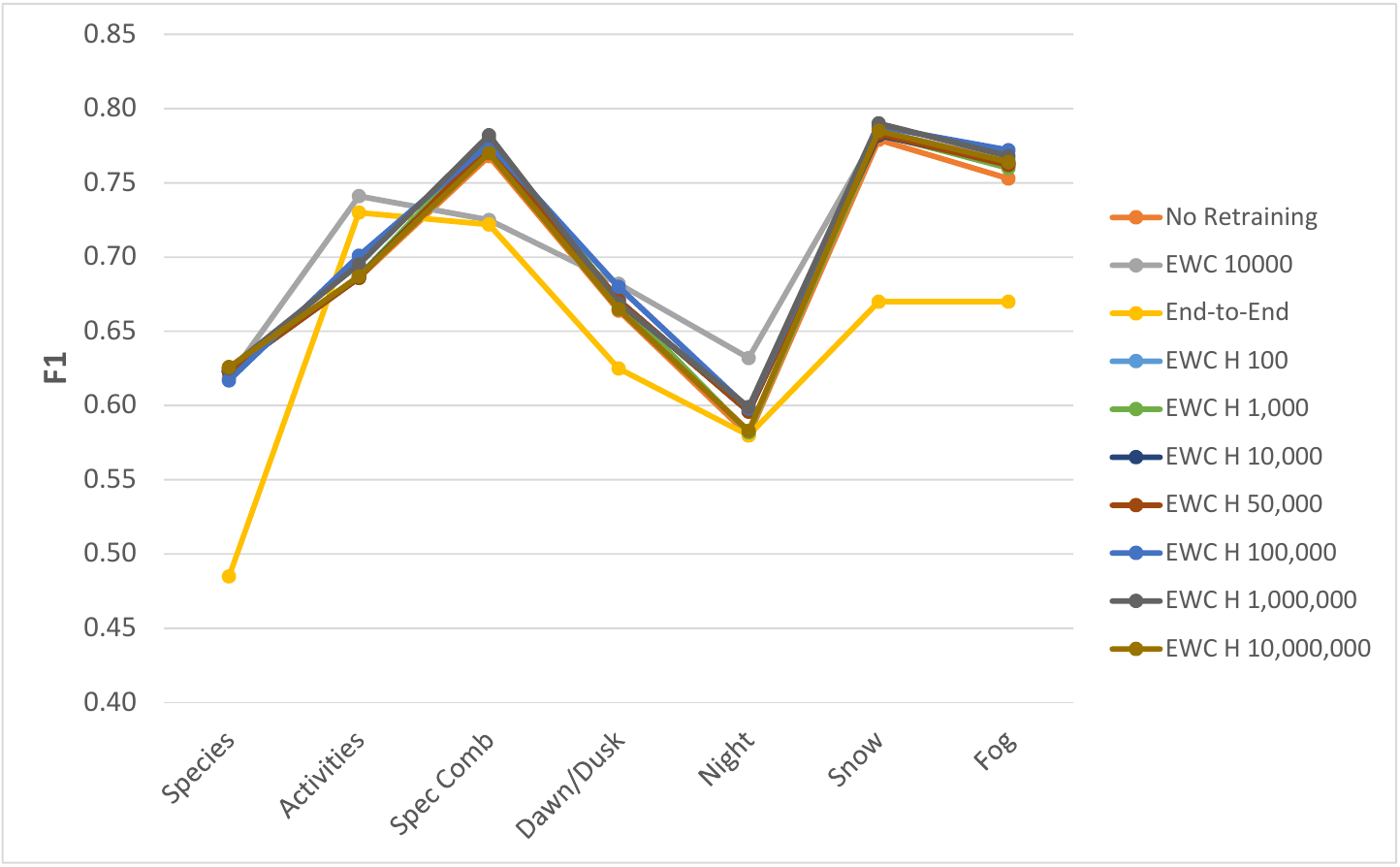}
		\caption{Comparison of Species Presence $F_1$ for Selected Trials of Various Continual Learning Configurations}
		\label{fig:species-presence-tuning}
	\end{figure}

\begin{figure}
		\centering
		\includegraphics[width=\textwidth]{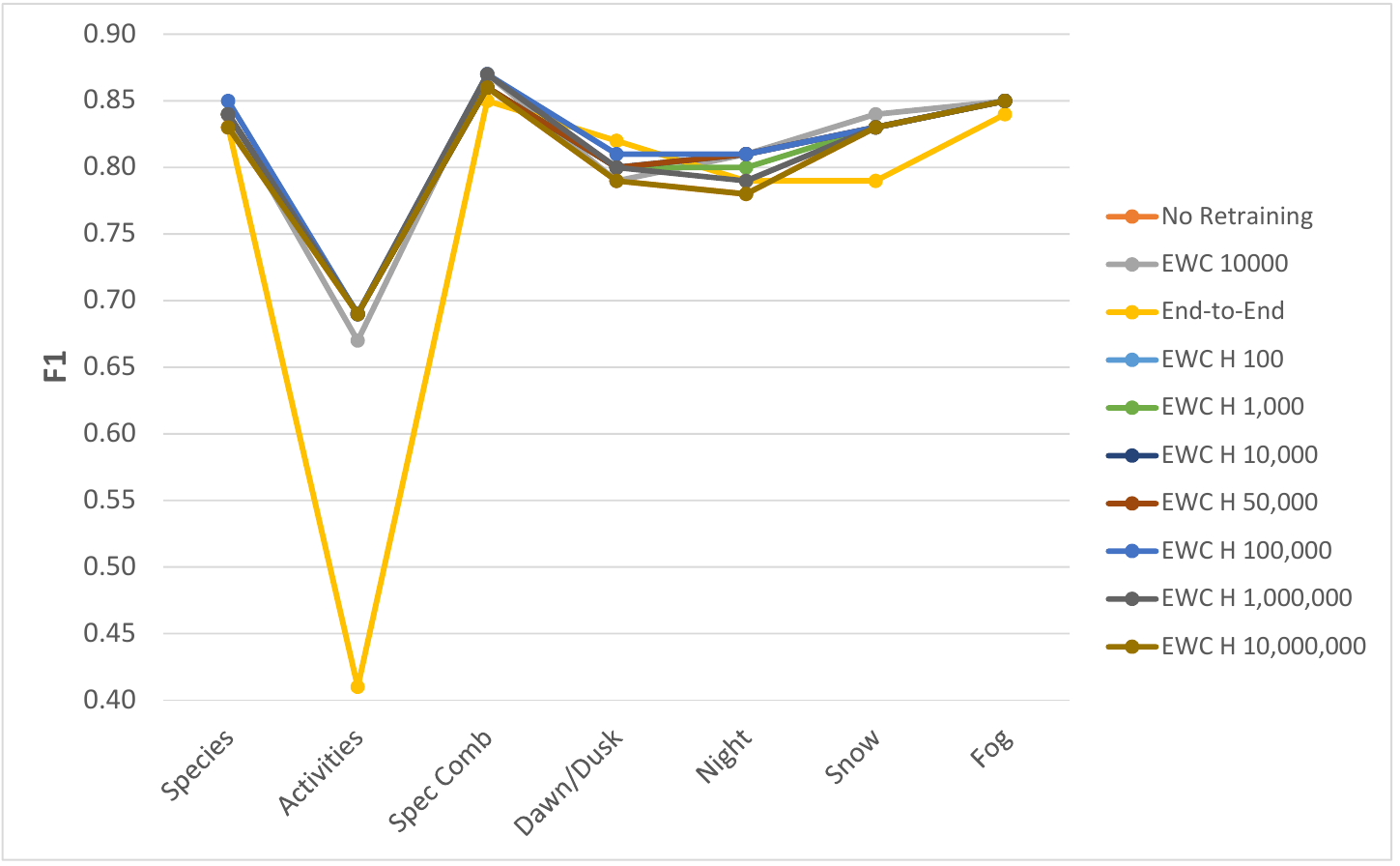}
		\caption{Comparison of Activity Presence $F_1$ for Selected Trials of Various Continual Learning Configurations}
		\label{fig:activity-presence-tuning}
	\end{figure}

	We ran several other configurations of EWC to see whether its performance could be improved. We varied two aspects. First, we varied the strength of the regularization parameter. The larger its value, the more EWC is encouraged to remain close to the pre-training weights. The second aspect that we varied was whether to have EWC modify all of the weights in the network or only the weights in the classifier heads (each a 2-layer fully-connected network). We also ran one experiment (``End-to-End'') in which we combined all of labeled data obtained from post-novelty feedback queries with all of the pre-training data and retrained the entire network from scratch. Figure~\ref{fig:mae-tuning}, \ref{fig:species-presence-tuning}, and \ref{fig:activity-presence-tuning} plot MAE and $F_1$ values for various configurations. The values plotted are the combined scores during the Test Phase (40\% novel and 60\% known images). EWC 10000 is the configuration of DCA-SS based on the validation set. The ``EWC H'' configurations apply EWC only to the classifier Heads and vary the EWC regularization parameter. Several trends are apparent. First, the End-to-End configuration generally performs poorly, despite the fact that it includes a class balancing mechanism. Second, EWC 10000 performs quite well. Third, the best overall performer is ``EWC H 100,000'', but the setting of the EWC regularization parameter has relatively little effect.
	
	
\end{document}